\documentclass{article}

\usepackage{microtype}

\usepackage{hyperref}

\usepackage{amsmath,amsfonts,bm}

\def\eqref#1{equation~\ref{#1}}

\def\1{\bm{1}}

\DeclareMathAlphabet{\mathsfit}{\encodingdefault}{\sfdefault}{m}{sl}
\SetMathAlphabet{\mathsfit}{bold}{\encodingdefault}{\sfdefault}{bx}{n}

\usepackage{booktabs}
\usepackage{graphicx}
\usepackage{multirow}
\usepackage{color}
\usepackage{colortbl}
\usepackage{hyperref}
\usepackage{url}
\usepackage{comment}
\usepackage{enumitem}
\usepackage{hyperref}       
\usepackage{makecell}
\usepackage{float}
\usepackage{subcaption}
\usepackage[dvipsnames]{xcolor}
\usepackage{wrapfig}
\usepackage{listings}
\usepackage{ragged2e}
\usepackage{pifont}
\usepackage{minitoc}
\usepackage{tocloft}

\definecolor{customgray}{HTML}{85888d}

\usepackage[noabbrev,capitalize]{cleveref}

\crefname{equation}{Eq.}{equations}
\crefname{line}{line}{lines}
\crefname{section}{\S}{\S\S}
\Crefformat{section}{#2\S#1#3}
\crefrangeformat{section}{\S\S#3#1#4--#5#2#6}

\addtocontents{toc}{\protect\setcounter{tocdepth}{0}} 
\setitemize[1]{leftmargin=12pt,itemsep=1.5pt,partopsep=0pt,parsep=0.5pt,topsep=0pt}

\usepackage[accepted]{icml2025}

\usepackage{amsmath}
\usepackage{amssymb}
\usepackage{mathtools}
\usepackage{amsthm}

\usepackage[capitalize,noabbrev]{cleveref}

\theoremstyle{plain}

\theoremstyle{definition}

\theoremstyle{remark}

\usepackage[textsize=tiny]{todonotes}

\icmltitlerunning{\textsc{Sigma}: Differential Rescaling of Query, Key and Value for Efficient Language Models}

\begin{document}

\twocolumn[
\icmltitle{\textsc{Sigma}: Differential Rescaling of Query, Key and Value \\ for Efficient Language Models}



\icmlsetsymbol{equal}{$\heartsuit$}
\icmlsetsymbol{core}{$\diamondsuit$}
\icmlsetsymbol{cont}{$\clubsuit$}
\icmlsetsymbol{lead}{$\spadesuit$}

\begin{icmlauthorlist}
\icmlauthor{Zhenghao Lin}{equal,sigma}
\icmlauthor{Zihao Tang}{equal,sigma}
\icmlauthor{Xiao Liu}{equal,sigma}
\icmlauthor{Yeyun Gong}{equal,sigma}
\\
\icmlauthor{Yi Cheng}{core,sigma}
\icmlauthor{Qi Chen}{core,sigma}
\icmlauthor{Hang Li}{core,sigma}
\icmlauthor{Ying Xin}{core,sigma}
\icmlauthor{Ziyue Yang}{core,sigma}
\\
\icmlauthor{Kailai Yang}{cont,sigma}
\icmlauthor{Yu Yan}{cont,sigma}
\icmlauthor{Xiao Liang}{cont,sigma}
\icmlauthor{Shuai Lu}{cont,sigma}
\icmlauthor{Yiming Huang}{cont,sigma}
\icmlauthor{Zheheng Luo}{cont,sigma}
\icmlauthor{Lei Qu}{cont,sigma}
\icmlauthor{Xuan Feng}{cont,sigma}
\icmlauthor{Yaoxiang Wang}{cont,sigma}
\icmlauthor{Yuqing Xia}{cont,sigma}
\icmlauthor{Feiyang Chen}{cont,sigma}
\icmlauthor{Yuting Jiang}{cont,sigma}
\icmlauthor{Yasen Hu}{cont,sigma}
\icmlauthor{Hao Ni}{cont,sigma}
\icmlauthor{Binyang Li}{cont,sigma}
\icmlauthor{Guoshuai Zhao}{cont,sigma}
\icmlauthor{Jui-Hao Chiang}{cont,sigma}
\icmlauthor{Zhongxin Guo}{cont,sigma}
\icmlauthor{Chen Lin}{cont,sigma}
\icmlauthor{Kun Kuang}{cont,sigma}
\icmlauthor{Wenjie Li}{cont,sigma}
\icmlauthor{Yelong Shen}{cont,sigma}
\icmlauthor{Jian Jiao}{cont,sigma}
\\
\icmlauthor{Peng Cheng}{lead,sigma}
\icmlauthor{Mao Yang}{lead,sigma}
\end{icmlauthorlist}

\icmlaffiliation{sigma}{Microsoft \textsc{Sigma} Team}


\icmlkeywords{Machine Learning, ICML}

\vskip 0.3in
]



\printAffiliationsAndNotice{\icmlEqualContribution} 

\begin{abstract}
We introduce \textsc{Sigma}, an efficient large language model specialized for the system domain, empowered \emph{DiffQKV attention}, 
and pre-trained on self-collected system domain data. Given the varing impacts on the model performance and efficienct indicators of Query (Q), Key (K), and Value (V), \textsc{Sigma} use DiffQKV attention to optimize them differentially and significantly enhance inference efficiency.
Specifically, we (1) differentially compress K and V, leveraging the model’s varying sensitivity to KV compression, as demonstrated in our extensive experiments, and (2) propose augmented Q to expand the dimension of the Q head, which improves the representation capacity of the model with minimal impacts on the inference speed.
Rigorous theoretical and empirical analyses reveal that DiffQKV attention significantly enhances inference efficiency, achieving up to a 33.36\% improvement over the conventional grouped-query attention (GQA) in long-context scenarios.
We pre-train \textsc{Sigma} on 6T tokens from various sources, including 19.5B system domain data that we carefully collect and 1T tokens of synthesized and rewritten data.
In general domains, \textsc{Sigma} performs on par with state-of-the-art models. In the system domain, it excels on \textsc{AIMicius}, the first comprehensive system-domain benchmark we propose, surpassing GPT-4 by up to 52.5\% across all tasks.
\end{abstract}

\section{Introduction}
In recent years, significant progress has been made in the development of large language models (LLMs), which have demonstrated remarkable performance across a wide range of domains \citep{bubeck2023sparks, jiang2023mistral, glm2024chatglm, dubey2024llama, yang2024qwen2}.
Meanwhile, a novel research direction, known as the ``system domain,'' has emerged with promising potential to further accelerate AI development through automated optimization of AI infrastructure \citep{xiong2024superbench,shi2024enhancing,202412.2294}.
This domain focuses on leveraging AI models to autonomously validate, evaluate, diagnose, and optimize the key components of AI infrastructure (e.g., hardware, configurations, cloud services, databases, and workloads).
Despite its promise, however, this important area of research has yet to receive commensurate attention.

To bridge this gap, this paper introduces \textsc{Sigma}, an efficient large language model specialized for the system domain, empowered by a novel architecture including \emph{DiffQKV} attention and our carefully collected system domain data. 

The DiffQKV attention substantially improves \textsc{Sigma}'s inference efficiency by mitigating the bottleneck associated with \emph{KV cache} \citep{pope2023efficiently}.\footnote{KV cache is
a common technique in the decoder-only Transformer architecture \citep{radford2019language}, which stores Key and Value vectors for future reuse in decoding. It can consume considerable GPU memory \citep{pope2023efficiently} and place substantial demands on memory bandwidth \citep{shazeer2019fast,ribar2024sparq}.}
Despite significant efforts to address the KV cache issue \citep{ainslie2023gqa, kwon2023efficient,luohe2024keep,zhang2024h2o}, previous studies tend to treat the compression of K and V vectors uniformly and rarely take into account the optimization of Q. 
In contrast, DiffQKV attention differentially optimizes the Query (Q), Key (K), and Value (V) components in the attention mechanism with tailored strategies, based on their varying impacts on the model performance and efficiency indicators.
Specifically, it involves two critical techniques: \emph{differentially compressed KV} and \emph{augmented Q}. 
\textbf{Differentially compressed KV} aggressively compresses K while lightly compressing V, grounded on our experimental findings that model performance is significantly more sensitive to compression in V vectors than in K, both in terms of dimension and the number of heads.
\textbf{Augmented Q} adopts a higher dimension for the Q head compared to KV heads, as we discover that introducing extra parameters to Q can effectively boost the model performance. 
With minimal impacts on the inference speed
, augmented Q can to some degree counteract the performance decline that inevitably results from KV compression. Rigorous theoretical and empirical analyses reveal that \textsc{Sigma}'s DiffQKV significantly enhances efficiency, achieving up to a 33.36\% improvement in inference speed over the conventional grouped-query attention (GQA) in long-context scenarios. 

To equip \textsc{Sigma} with the capability of addressing system domain tasks, we carefully identify 15 primary source categories from over 120 system-related websites, collecting a total of 19.5 billion data for system domain pre-training and fine-tuning. Moreover, we construct the \textsc{AIMicius} benchmark to facilitate the evaluation of system domain task performance. It includes four major tasks - CMDGen, Infrawise, Optiflow, and NL2KQL - based on Azure services, which assess the critical capabilities in the system domain, such as command generation, benchmark retrieval, topology optimization, and infrastructure issue analysis.

Building on the above commitments, we pre-train \textsc{Sigma} on 6T tokens from various sources, with around 1T tokens from synthesized and rewritten data and 19.5B system domain data, all of which have undergone extensive quality screening.
In general domains, \textsc{Sigma} achieves comparable performance to other state-of-arts models.
Besides, \textsc{Sigma} demonstrates remarkable performance across all tasks in \textsc{AIMicius}, significantly outperforming GPT-4 with an absolute improvement up to 52.5\%.

\section{DiffQKV Attention}
\label{sec:diffQKV}
In the standard Multi-Head Attention (MHA) introduced by \citet{vaswani2017attention}, Q, K, and V components consistently employ an equal number of heads, with each head maintaining the same dimension. 
Our research explores a more generalized form of attention mechanism, termed \textbf{DiffQKV attention}, where QKV can possess distinct numbers of heads as well as different dimensions per head.
In this section, we will first formally describe the DiffQKV attention process and then discuss the optimal configurations for it to effectively balance efficiency and model performance.

\noindent\textbf{Formulation of DiffQKV Attention.} 
Let the numbers of QKV heads as $n_q^h, n_k^h, n_v^h$, respectively, and the dimensions of QKV heads as $d_q^h, d_k^h, d_v^h$. Formally, given the attention input $h_t$, we begin by transforming it to: 
\vspace{-.1cm}
$$q_t = h_t\cdot W_Q, \quad k_t = h_t\cdot W_K,  \quad v_t = h_t\cdot W_V, $$
\vspace{-.1cm}
where the dimensions of $q_t$, $k_t$, $v_t$ are $[n_q^h, d_q^h]$, $[n_k^h, d_k^h]$, $[n_v^h, d_v^h]$ respectively.
Then, we load the previous K vectors from the cache and concatenate it with $k_t$, thereby obtaining $K_t$, which has a dimension of $[t, n_k^h, d_k^h]$. 
To prepare for the attention score computation, $K_t$ and $q_t$ are sliced into a series of heads. Since the number of K heads and Q heads can be inconsistent, we employ a \emph{GroupSharing} operation on $K_t$, which allows multiple Q heads to be jointly associated with the same K head (see Appendix, \cref{fig:overview}): 
\vspace{-.1cm}
\begin{align}
    [K_{(t,1)}; K_{(t,2)}; ..; K_{(t,n^h_q)}] &= \text{\emph{GroupSharing}}(K_t, n^h_q, n^h_k),\nonumber\\
    [{q}_{(t,1)}; {q}_{(t,2)}; ..; {q}_{(t,n^h_q)}] &= q_t, \nonumber
\end{align}
\vspace{-.1cm}
where $q_{(t,i)}\in \mathbb{R}^{d_k^h}$ represents the $i$-th Q head. $K_{(t,i)}$ is in the shape of $[t, d_k^h]$. The attention scores are computed as:
\vspace{-.1cm}
$$
    \alpha_{(t,i)} = \text{Softmax}_i [\frac{\text{\emph{Attend}}({q}^{\;T}_{(t,i)}, K_{(t,i)})}{\sqrt{d_k}}],  
    \label{eq:diffqkv_att_score}
$$
\vspace{-.1cm}
where $\alpha_{(t,i)} \in \mathbb{R}^t$. 
In standard MHA, the \emph{Attend} function computes the inner products of the Q heads and K heads. Nonetheless, as DiffQKV accommodates varying dimensions for the Q and K heads, an alternative implementation of the \emph{Attend} function is necessitated. In our experiments, we implement this part by transforming the dimension of K to the same as V through another feed-forward layer. 
Next, we load previous V cache and compute the outputs as: 
\vspace{-.1cm}
\begin{align}
    V_t = [V_{\text{t-1}}; v_t], \quad V_{\text{t-1}} &= \text{\emph{LoadCacheV}}(\&V_{\text{cache}}, \alpha_{(t,i)}),\nonumber\\
    [V_{(t,1)}; V_{(t,2)}; ..; V_{(t,n^h_q)}] & = \text{\emph{GroupSharing}}(V_t, n^h_q, n^h_v),\nonumber\\
    o_{(t,i)} &= \sum\nolimits_{j=1}^{t}\alpha_{(j,i)}V_{(j,i)},\nonumber\\
    x_t &=  W_O [o_{(t,1)}; o_{(t,2)}; ...; o_{(t,n_v^h)}].\nonumber
\end{align}
\vspace{-.1cm}
Here, the dimension of $V_t$ is $[t, n_v^h, v^h]$,  that of $V_{(t,i)}$ is $[t, d_v^h]$,  $o_{(t,i)}\in\mathbb{R}^{d^h_v}$ and $x_t\in\mathbb{R}^{d^m}$. 
Notably, the \emph{LoadCacheV} function, used for retrieving previous V vectors, can differ from the strategy for loading K cache. 

\label{sec:obs}
In order to find the optimal configurations of DiffQKV attention, we train the model with different settings of DiffQKV from scratch and assess their performance.
In the following experiments, we adopt 100B tokens from FineWeb-Edu \citep{penedo2024fineweb} as the pre-training data and the model are scaled to approximately 1B parameters. 
We employ the following benchmarks to evaluate the model: HellaSwag ~\citep{zellers2019hellaswag}, OpenBookQA ~\citep{OpenBookQA2018}, WinoGrande ~\citep{sakaguchi2021winogrande}, ARC Challenge ~\citep{clark2018think}, PIQA~\citep{bisk2020piqa}, SciQ~\citep{welbl2017crowdsourcing}, BoolQ ~\citep{clark2019boolq}, LogiQA ~\citep{liu2020logiqa}, and LAMBADA~\citep{lambada2016}. Our major observations are as below. 

\begin{table*}[!th]
\centering%
\begin{minipage}[t]{0.35\textwidth}
\centering%
\caption{Comparisons of model performance when reducing an equal number of K or V heads. The number of Q heads is 32 for all models ($n_q^h=32$). Note that the reported performance scores in Table \ref{tab:kv_head_compression}-\ref{tab:large_q_vs_mlp} are the average results across nine benchmarks. See \cref{sec:observation_detailed} for detailed results on each benchmark.}
\vspace{0.5em}
\label{tab:kv_head_compression}
\resizebox{0.85\linewidth}{!}{
\begin{tabular}{l l}
\toprule
\textbf{Model} & {\textbf{Performance}} \\
\midrule
\textbf{MHA} ($n_k^h$=$n_v^h$=32)& \textbf{52.40} \\ 
\rowcolor{customgray!10}
-50\% V Heads ($n_v^h$=16)
& \textbf{51.74} \tiny{($\downarrow$0.66)} \\ 
\rowcolor{customgray!20}
-50\% K Heads ($n_k^h$=16)& \textbf{52.83} \tiny{(\underline{$\uparrow$0.43})}\\ 
\midrule
\textbf{GQA} ($n_k^h$=$n_v^h$=16)& \textbf{52.14}  \\ 
\rowcolor{customgray!10}
-75\% V Heads ($n_v^h$=4) & \textbf{51.76} \tiny{($\downarrow$0.38)} \\ 
\rowcolor{customgray!20}
-75\% K Heads ($n_k^h$=4) & \textbf{51.97} \tiny{(\underline{$\downarrow$0.17})}\\ 
\midrule
\textbf{GQA} ($n_k^h$=$n_v^h$=4)
& \textbf{51.66}\\ 
\rowcolor{customgray!10}
-75\% V Heads ($n_v^h$=1) & \textbf{51.03}  \tiny{($\downarrow$0.63)} \\ 
\rowcolor{customgray!20}
-75\% K Heads ($n_k^h$=1)
& \textbf{51.67}  \tiny{(\underline{$\uparrow$0.01})}\\ 
\hline
\end{tabular}}
\end{minipage}%
\hfill
\begin{minipage}[t]{0.3\textwidth}
\centering%
\caption{The ablation studies of halving the K head dimension. The results indicate that this adjustment, while largely improving the inference efficiency by reducing the size of KV cache,  does not significantly compromise performance. The number of Q heads is 32 for all models ($n_q^h=32$). }
\vspace{1em}
\label{tab:k_headdim_compression}
\resizebox{0.98\linewidth}{!}{
\begin{tabular}{l l}
\toprule
{\textbf{Model}} & {\textbf{Performance}} \\
\midrule
\textbf{MHA} ($n_k^h$=$n_v^h$=32) & \textbf{52.40}  \\ 
\rowcolor{customgray!15}
\emph{w/} Half K Dim. & \textbf{52.56} \tiny{($\uparrow$0.16)}\\ 
\midrule

\textbf{GQA} ($n_k^h$=$n_v^h$=16)
& \textbf{52.14} \\ 
\rowcolor{customgray!15}
\emph{w/} Half K Dim.  & \textbf{52.06} \tiny{($\downarrow$0.08)} \\ 
\midrule
\textbf{GQA} ($n_k^h$=$n_v^h$=4)
& \textbf{51.66} \\ 
\rowcolor{customgray!15}
\emph{w/} Half K Dim.& \textbf{51.92} \tiny{($\uparrow$0.26)}  \\ 
\bottomrule
\end{tabular}
}
\end{minipage}
\hfill
\begin{minipage}[t]{0.3\textwidth}
\centering%
\caption{Compare baseline model architectures with those having augmented Q. $d_q^h$ refers to the intermediate Q head dimension. The number of Q heads is 32 for all models in the table ($n_q^h=32$). For the baseline without AugQ, the intermediate dimension of Q head is $d_q^h=2048$. }
\vspace{0.75em}
\label{tab:large_q}
\resizebox{0.9\linewidth}{!}{
\begin{tabular}{l l }
\toprule
{\textbf{Model}} & {\textbf{Performance}} \\
\midrule
\textbf{MHA} 
& \textbf{52.40}\\ 
\rowcolor{customgray!20}
+ AugQ ($d_q^h$=5632)
& \textbf{53.03} \tiny{($\uparrow$0.63)}\\ 
\midrule
\textbf{GQA} ($n_k^h$=$n_v^h$=16)
& \textbf{52.14}\\ 
\rowcolor{customgray!5}
+ AugQ ($d_q^h$=3072)
& \textbf{53.38} \tiny{(\underline{$\uparrow$1.24})}\\ 
\rowcolor{customgray!10}
+ AugQ ($d_q^h$=4096)
& \textbf{52.93} \tiny{($\uparrow$0.79)} \\ 
\rowcolor{customgray!20}
+ AugQ ($d_q^h$=5632)
& \textbf{53.07} \tiny{($\uparrow$0.93)}\\ 
\midrule
\textbf{GQA} ($n_k^h$=$n_v^h$=4)
& \textbf{51.66}\\ 
\rowcolor{customgray!20}
+ AugQ ($d_q^h$=5632)
& \textbf{53.13} \tiny{($\uparrow$1.47)}\\ 
\bottomrule
\end{tabular}}
\end{minipage}
\vspace{-4mm}
\end{table*}

\noindent\textbf{Observation 1: The model performance is more sensitive to the decrease in the number of V heads ($n^h_v$) than in the number of K heads ($n^h_k$).} 
Specifically, we reduce the number of heads of K and V in the DiffQKV attention, respectively, and assess their impacts on model performance by comparing with the baseline of equal KV heads. 
As shown in \cref{tab:kv_head_compression}, reducing K heads has a minor impact compared to reducing V heads in most cases. 
This differential impact is reasonable, considering the distinct roles of K and V within the attention mechanisms. 
The K heads primarily serve to compute attention matrices, which prove to have a remarkable sparsity exceeding 95\% during inference \citep{zhang2024h2o}, so a slight decrease in the parameters used for its calculation can still yield a precise approximation. 
In contrast, the V vectors, which directly influence the final attention output, demands a more nuanced approach.

\noindent\textbf{Observation 2: Reducing K head dimension to half of V head dimension ($d^h_k = d^h_v/2$) can preserve the model performance compared to equal-dimension K and V heads ($d^h_k = d^h_v$).} 
We explore compressing K components by shrinking K head dimension. To compute the attention scores when the dimensions of K and V are different, we need to transform the dimension of $K$ to the same as $V$ through a trainable feed-forward layer.
In essence, the reduction of K head dimension reduces KV cache at the cost of additional computation from a feed-forward layer. This trade-off is cost-effective as the primary bottleneck for LLM inference speed lies in memory consumption rather than computation \citep{shazeer2019fast, ribar2024sparq}. 
To study the impact, we conduct compare several commonly-used attention settings, such as MHA and GQA, with their counterparts having only half of the K head dimension. 
Results shown in \cref{tab:k_headdim_compression} demonstrate that the performance decrease from reducing K head dimension is negligible; in some cases, the model with a smaller K head dimension even outperforms the baseline setting prior to compression.

\label{sec:observation:augmenting_q}
\noindent\textbf{Observation 3: Augmenting Q components can boost model performance.} 
Augmenting with extra parameters is a straightforward way to enhance a model's representational capacity and performance. 
Since Q vectors in self-attention do not need to be cached during inference, their impact on memory usage and data transfer duration is minimal. Thus, adding extra parameters to Q components is more cost-effective than doing so for KV components. To verify this, we experiment by varying Q's parameter count. As demonstrated in \cref{tab:large_q}, adding extra parameters to Q can consistently enhance the model's performance to different extents, especially when K and V have higher compression levels ($n^h_v$=4,$n^h_k$=4).
Overall, a scaling factor of $\times$1.5 (i.e. $d_q^h$=3072) on the Q component appears optimal.

\begin{table}[t]
\centering
\small
\vspace{-2mm}
\caption{Comparisons of the model performance when incorporating the augmented Q component (AugQ) with different sizes and enlarging the FFN module (AugF). The baseline method is GQA, with the FFN dimension being 5632 and $n_k^h$=$n_v^h$=16. $\Delta d_F$ denotes the enlarged dimension for the FFN module, while $d_q^h$ represents the intermediate Q head dimension ($\delta$=$3072$). }
\vspace{0.5em}
\label{tab:large_q_vs_mlp}
\resizebox{0.82\linewidth}{!}{
\begin{tabular}{l l ccccccccc}
\toprule
{\textbf{Model}} & {\textbf{Performance}}\\
\midrule
\textbf{GQA}
& \textbf{52.14}\\ 
\rowcolor{customgray!8}
+ AugF ($\Delta d_{\text{F}}$=$\delta$)
& \textbf{53.26} \tiny{($\uparrow$1.12)}\\ 
\rowcolor{customgray!20}
+ AugQ ($d_q^h$=$\delta$)
& \textbf{53.38} \tiny{(\underline{$\uparrow$1.24})} \\ 
\midrule
\rowcolor{customgray!8}
+ AugF ($\Delta d_{\text{F}}$=$2\delta$)
& \textbf{53.16} \tiny{($\uparrow$1.02)}\\ 
\rowcolor{customgray!20}
+ AugF ($\Delta d_{\text{F}}$=$\delta$) \& AugQ ($d_q^h$=$\delta$)
& \textbf{54.55} \tiny{(\underline{$\uparrow$2.41})}\\ 
\midrule
\rowcolor{customgray!8}
+ AugF ($\Delta d_{\text{F}}$=$3\delta$)
& \textbf{54.50} \tiny{($\uparrow$2.36)}\\ 
\rowcolor{customgray!20}
+ AugF ($\Delta d_{\text{F}}$=$2\delta$) \& AugQ ($d_q^h$=$\delta$)
& \textbf{54.67} \tiny{(\underline{$\uparrow$2.53})}\\ 
\midrule
\rowcolor{customgray!8}
+ AugF ($\Delta d_{\text{F}}$=$5\delta$)
& \textbf{55.08} \tiny{($\uparrow$2.94)}\\ 
\rowcolor{customgray!20}
+ AugF {($\Delta d_{\text{F}}$=$3\delta$)} \& AugQ {($d_q^h$=$\delta$)}  & \textbf{55.09} \tiny{(\underline{$\uparrow$2.95})} \\ 
\bottomrule
\vspace{-6mm}
\end{tabular}
}
\end{table}
\noindent\textbf{Observation 4: With the same number of extra parameters, augmenting Q yields a greater performance boost than expanding the FFN module. Moreover, the performance gains from these two actions are independent. } That is, with a fixed augmented Q, further enlarging the FFN module still improves model performance. 
By adding the same number of parameters to the FFN and Q modules, respectively, we examine and compare their impacts on the model performance. 
Results in \cref{tab:large_q_vs_mlp} show that augmenting Q consistently gives a more significant performance boost than adding the same number of parameters to the FFN module. Comparing the last two lines of \cref{tab:large_q_vs_mlp}, even when FFN's additional parameters (2$\delta$) are double those of Q ($d^h_q=\delta$), the model performance is still slightly worse with FFN. This indicates that adding parameters to the Q component in the self-attention layer more effectively improves model performance than to the FFN module.

\noindent\textbf{Summary: Optimal Configurations for DiffQKV Attention.}
\label{sec:obs_sum}
The observations detailed above give rise to two primary guidelines for the optimal configurations of DiffQKV attention: 1) \textbf{differentially compressed KV}: Aggressive compression can be more pronouncedly applied to the K components as opposed to the V components, both in terms of the head number and head dimension; 
2) \textbf{augmented Q}: introducing extra parameters to the Q components can enhance the representational capacity of the self-attention layer, while having a relatively small impact on inference efficiency. More experiments that combine different settings of DiffQKV are included in Appendix \cref{tab:combine_exp}.

\noindent\textbf{\textsc{Sigma} Model Architecture.}
\label{sec:architecture}
Building on the DiffQKV attention, we construct a pre-trained language model, named \textsc{Sigma}.  
Specifically, we adopt two model scales with 1.5B parameters and 10B parameters, respectively (i.e. \textsc{Sigma}-1.5B and \textsc{Sigma}-10B). They both adopt the aforementioned strategies of differentially compressed KV and augmented Q in the self-attention layer. 
The detailed configurations of the \textsc{Sigma} architecture are included in \cref{sec:appendix_architecture}.

\section{Efficiency Analysis}
\label{sec:efficiency_analysys}

In this section, we analyze \textsc{Sigma}-1.5B's efficiency both theoretically and empirically, using FlashAttention2 \citep{dao2023flashattention2} as the default attention mechanism.

\subsection{Theoretical Analysis} \label{sec:efficiency_analysys:theoretical_analysis}
The efficiency gains of \textsc{Sigma} can be largely attributed to the reduction in the number of key heads $n^h_k$. In \textsc{Sigma}-1.5B, $n^h_k$ is reduced from 16 to 4, whereas $n^h_v$ remains 16. This reduction directly impacts two critical efficiency indicators: \textbf{KV Cache} and \textbf{Attention Computation}.

\begin{figure*}[!ht]
    \centering
    \begin{subfigure}[b]{0.24\textwidth}
        \includegraphics[width=\textwidth]{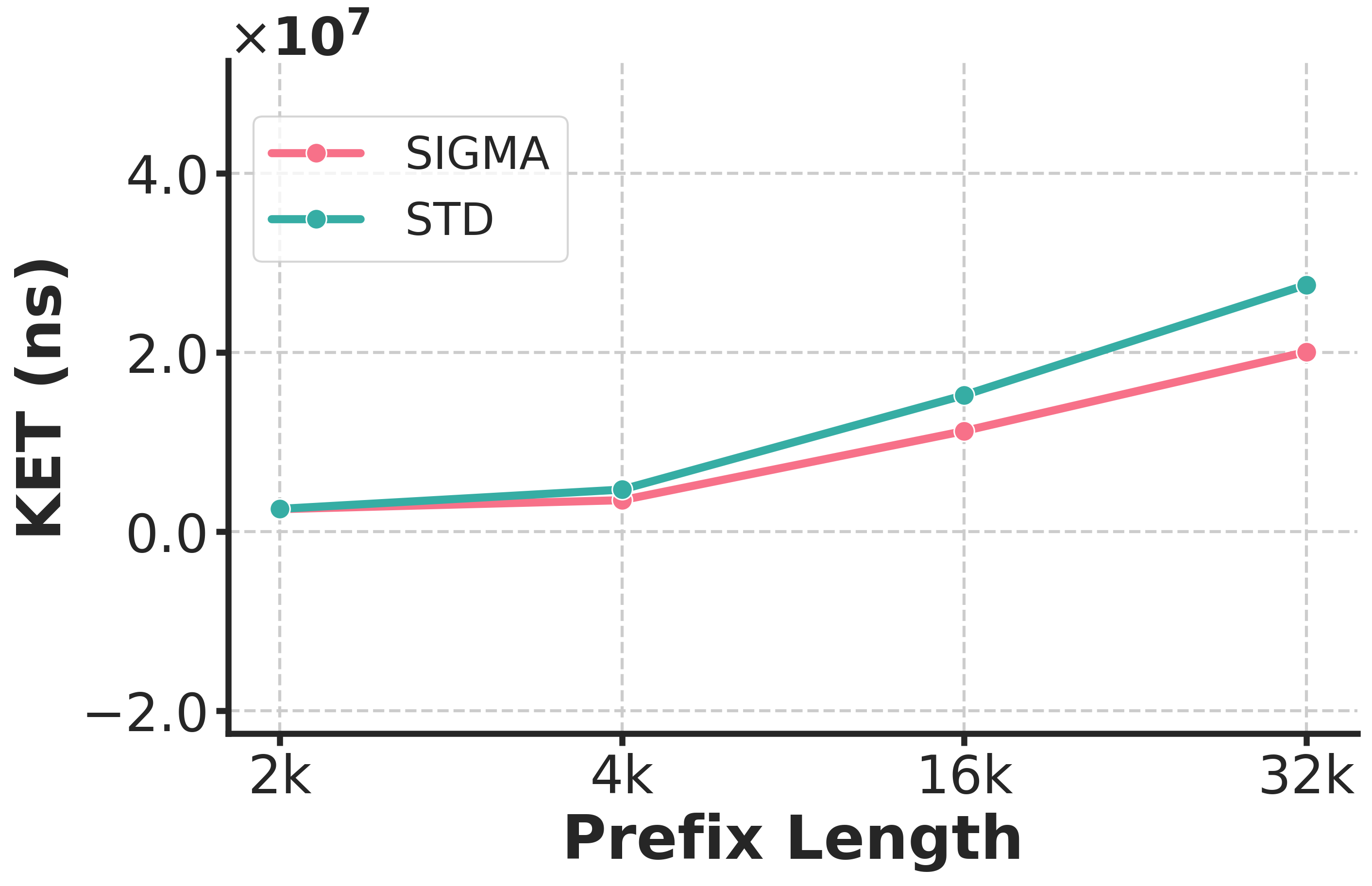}
        \caption{KET of the split kernel.}
        \label{fig:kernel_cost_a}
    \end{subfigure}
    \vspace{-3pt}
    \hfill
    \begin{subfigure}[b]{0.24\textwidth}
        \includegraphics[width=\textwidth]{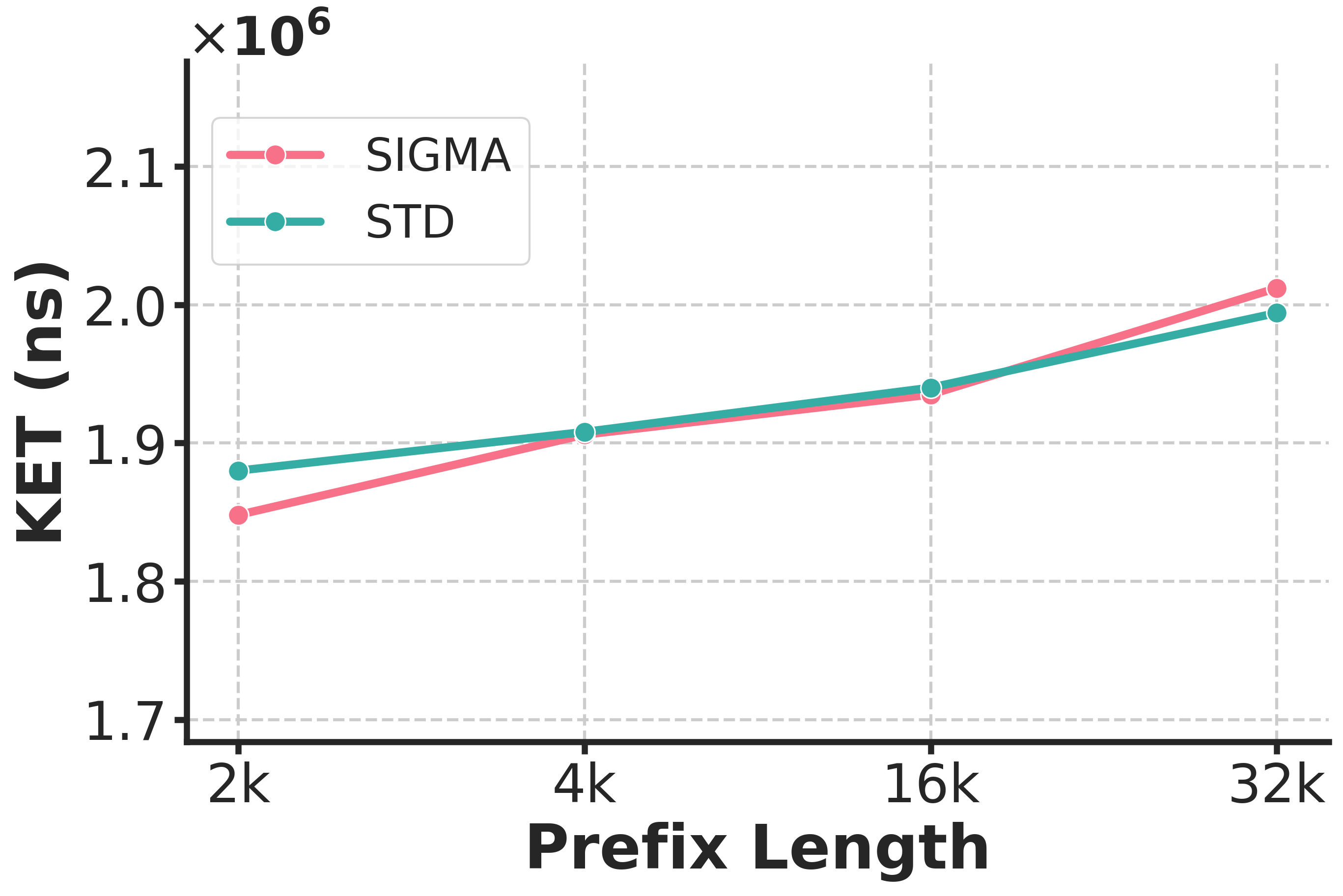}
        \caption{KET of the combine kernel.}
        \label{fig:kernel_cost_b}
    \end{subfigure}
    \vspace{-3pt}
    \hfill
    \begin{subfigure}[b]{0.24\textwidth}
        \includegraphics[width=\textwidth]{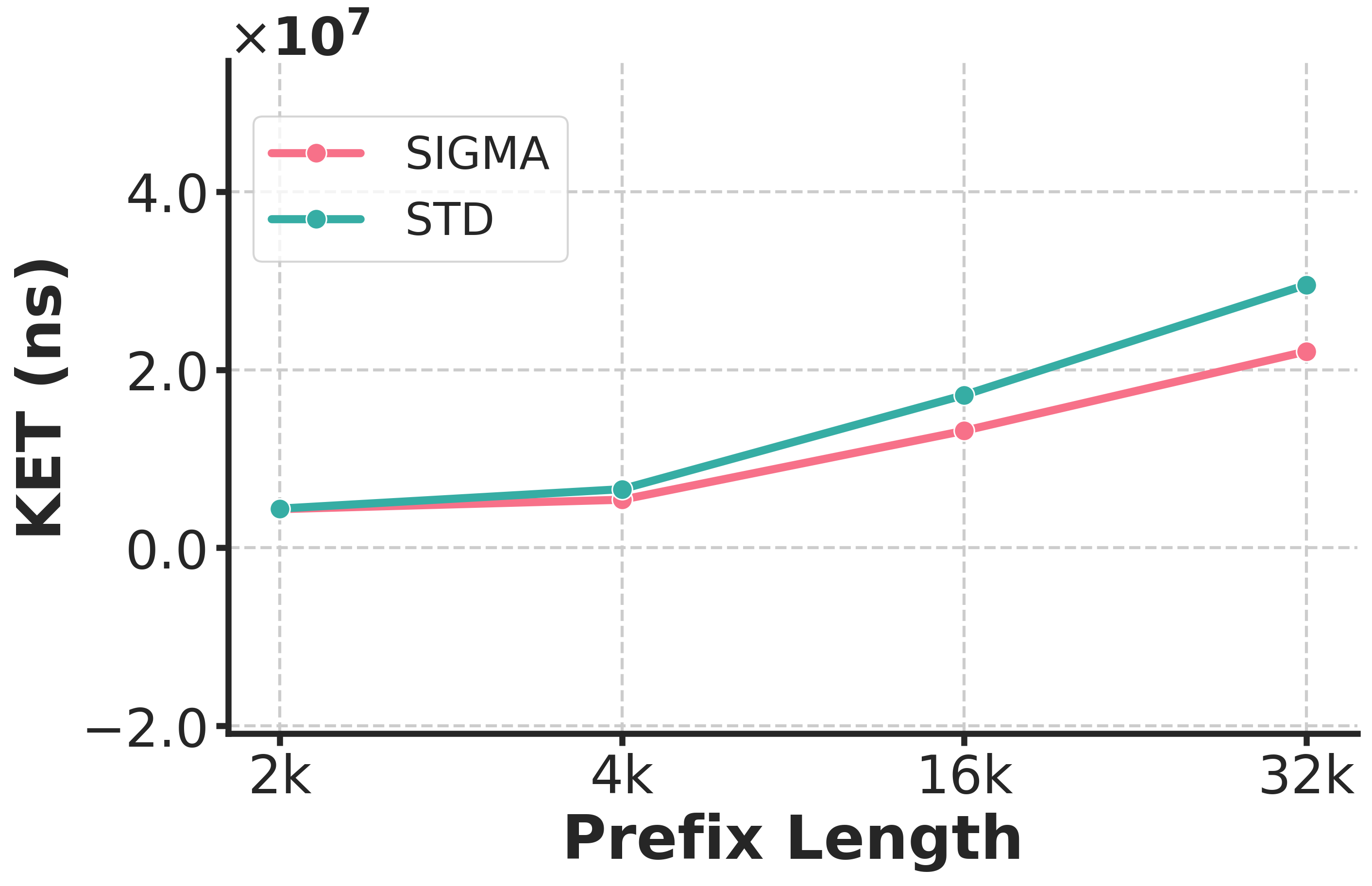}
        \caption{Total KET.}
        \label{fig:kernel_cost_c}
    \end{subfigure}
    \vspace{-3pt}
    \hfill
    \begin{subfigure}[b]{0.24\textwidth}
        \small
        \resizebox{\linewidth}{!}{
        \setlength{\tabcolsep}{2.3pt}
        \begin{tabular}{cccc}
        \toprule
        \bf Prefix & \bf Split & \bf Combine & \bf Total \\
        \midrule
        \bf 2k & 1.17\%  & 1.68\%  & 1.39\% \\
        \bf 4k & 25.33\%  & 0.08\%  & 18.02\% \\
        \bf 16k & 26.30\%  & 0.25\%  & 23.35\% \\
        \bf 32k & 27.21\%  & -0.93\% & 25.31\% \\
        \bottomrule
        \end{tabular}}
        \caption{KET's relative improvement of each kernel}
    \end{subfigure}
    \vspace{-3pt}
    \caption{KET comparison of FlexHeadFA between Standard model(\textsc{Std}) and \textsc{Sigma}.}
    \vspace{-10pt}
    \label{fig:kernel_cost}
\end{figure*}
\noindent\textbf{KV Cache.} Generally, key cache and value cache take dimensions $(\texttt{b}, \texttt{s}, n^h, d^h)$, where \texttt{b} is the batch size, \texttt{s} is the sequence length, $n^h$ denotes the number of attention heads, and $d^h$ represents the dimension of each head. Since KV cache is responsible solely for storing and loading the key and value tensors, we model the cost of KV Cache as being proportional to the total number of elements in the key and value caches. Therefore, it can be expressed as a linear function of the cache size as shown in \cref{eq:cache_0}, with $\alpha$ representing the proportional cost per element and $\beta$ accounting for fixed overheads or unrelated losses.
\vspace{-.05cm}
\begin{equation}
\vspace{-.25cm}
L=\alpha\cdot[\texttt{b}\cdot\texttt{s}\cdot(n^h_k\cdot d^h_k+n^h_v\cdot d^h_v)]+\beta.
\label{eq:cache_0}
\end{equation}

In \textsc{Sigma} 1.5B, the key cache has the dimension of $(\texttt{b}, \texttt{s}, 4, 64)$, whereas the value cache is $(\texttt{b}, \texttt{s}, 16, 64)$. Compared to grouped query attention (GQA), since $n^h_k$ is reduced, the total cost of KV cache operations will decrease accordingly. As the size of cache increases, the reduction rate $r$ of the total cost of KV cache operations converges toward a theoretical value as shown in \cref{eq:imp_t}.
\vspace{-.05cm}
\begin{equation}
\vspace{-.25cm}
r=\lim_{\texttt{s}\rightarrow+\infty}\frac{L^{\rm GQA}-L^{\rm{Sigma}}}{L^{\rm GQA}}=\frac{32-20}{32}=37.5\%.
\label{eq:imp_t}
\end{equation}

\noindent\textbf{Attention Computation.} Here, the cost reduction also originates from the reduction in key heads. The theoretical reduction rate $r$ therefore remains 37.5\%. However, while KV cache operations are primarily I/O-intensive, attention computation is significantly more computation-intensive. Due to the increased complexity of this module, the observed reduction in practice is likely to deviate more significantly from the theoretical rate compared to the KV cache.

Apart from aforementioned improvements, it is important to highlight that the reduction of $n^h_k$ also decreases KV cache's memory consumption. \textsc{Sigma} hence allows for larger batch sizes, enhancing inference efficiency naturally.


\subsection{Implementation}
\label{sec:efficiency_analysys:implementation}
While reducing $n^h_k$ theoretically leads to significant efficiency improvements, practical implementation is hindered by the tight integration of KV cache and attention operations in FlashAttention and existing LLM deployment frameworks (\textit{e.g.}, vLLM \citep{kwon2023efficient} and TensorRT\footnote{\url{https://github.com/NVIDIA/TensorRT-LLM/}}). Here, we propose our current solutions and emphasize the need for broader DiffQKV support.

\noindent\textbf{Differential Management of KV Cache.} In \textsc{Sigma}, since key and value have different number of heads, their cache sizes differ. Generally, it poses no issues for most implementations that store K and V cache separately. However, for frameworks that combine key and value cache into a whole matrix, it becomes challenging. To accommodate these frameworks, one workaround is to duplicate K to match the size of V (named KV Group Sharing). However, this approach is highly discouraged, as it negates the efficiency improvements entirely by degrading to GQA. Its official implementation can be found in \cref{sec:kv_sharing}.

\begin{figure*}[!ht]
    \centering
    \begin{subfigure}[b]{0.24\textwidth}
        \includegraphics[width=\textwidth]{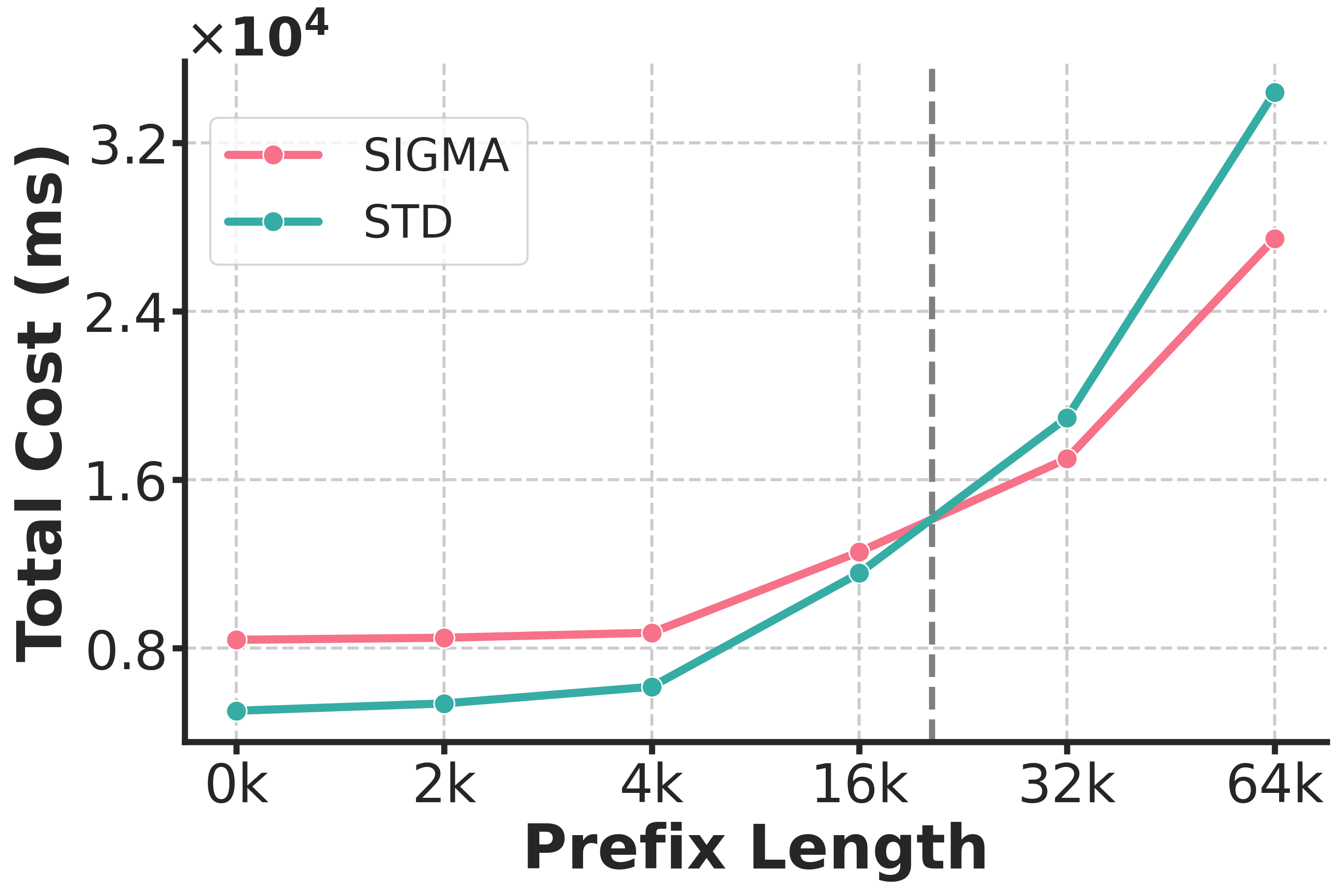}
        \caption{Output Length $= 2k$.}
        \label{fig:total_cost_a}
    \end{subfigure}
    \hfill
    \begin{subfigure}[b]{0.24\textwidth}
        \includegraphics[width=\textwidth]{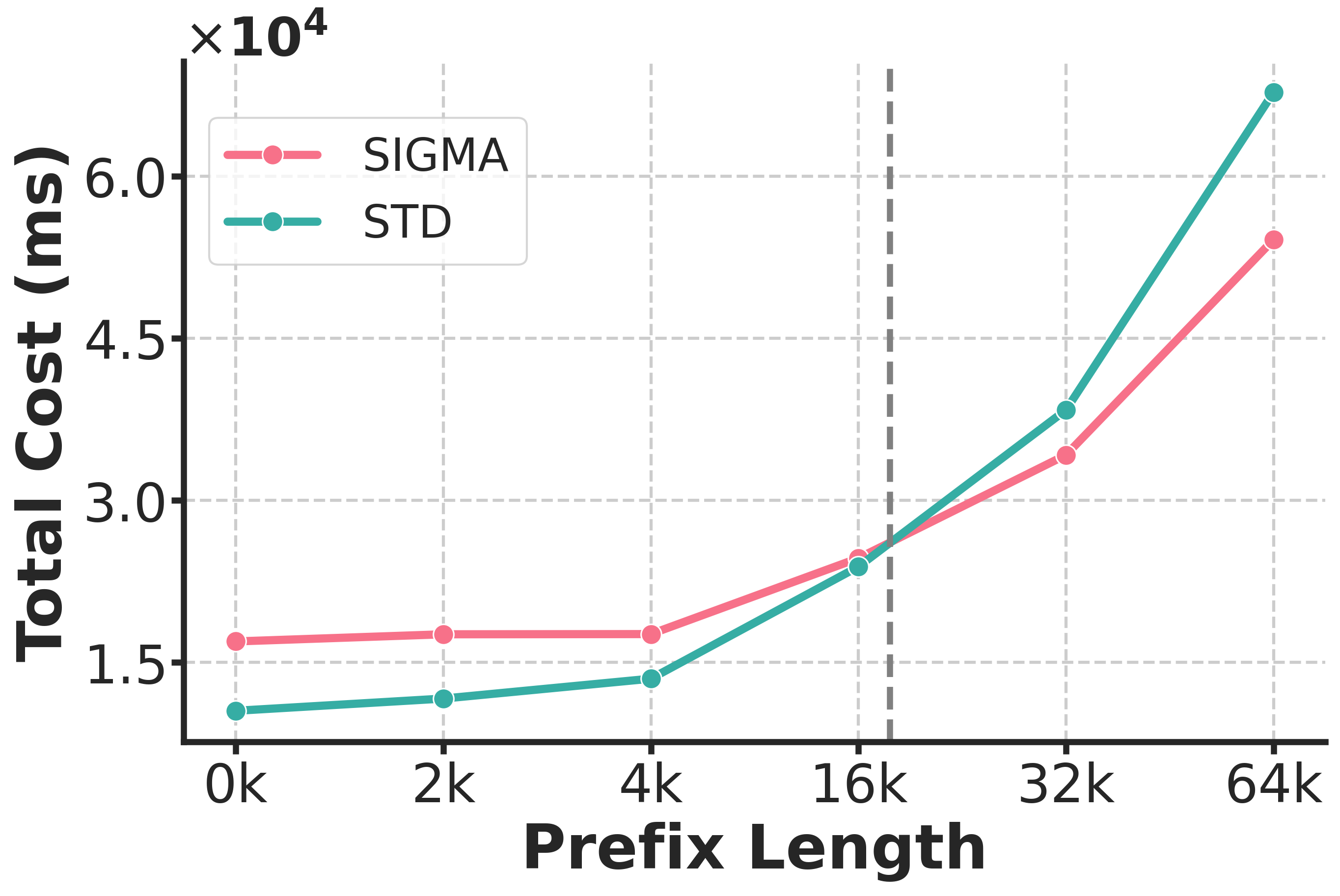}
        \caption{Output Length $= 4k$.}
        \label{fig:total_cost_b}
    \end{subfigure}
    \hfill
    \begin{subfigure}[b]{0.24\textwidth}
        \includegraphics[width=\textwidth]{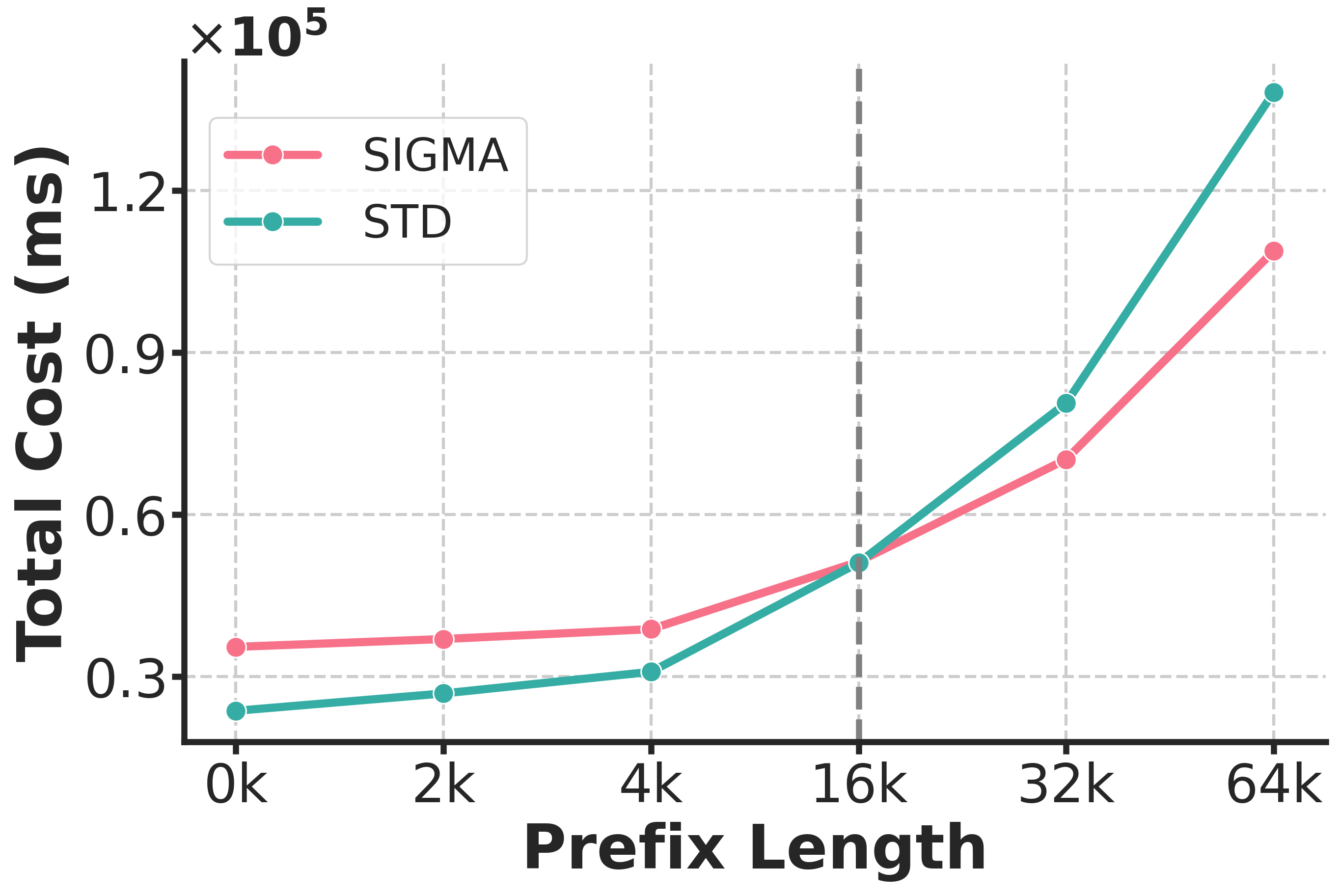}
        \caption{Output Length $= 8k$.}
        \label{fig:total_cost_c}
    \end{subfigure}
    \hfill
    \begin{subfigure}[b]{0.24\textwidth}
        \includegraphics[width=\textwidth]{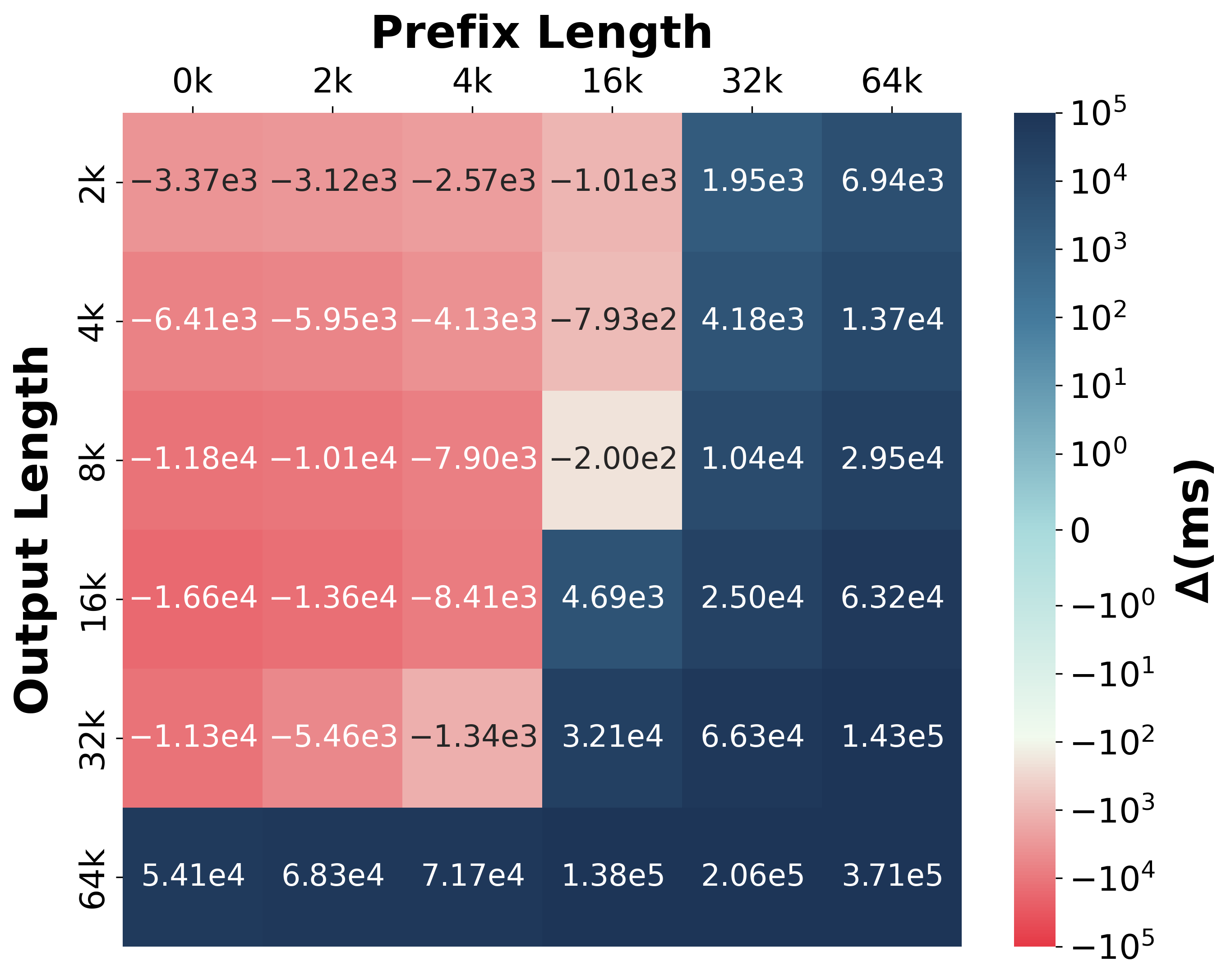}
        \caption{Absolute CEET Improvment.}
        \label{fig:total_cost_g}
    \end{subfigure}

    \vspace{-3pt}

     \centering
    \begin{subfigure}[b]{0.24\textwidth}
        \includegraphics[width=\textwidth]{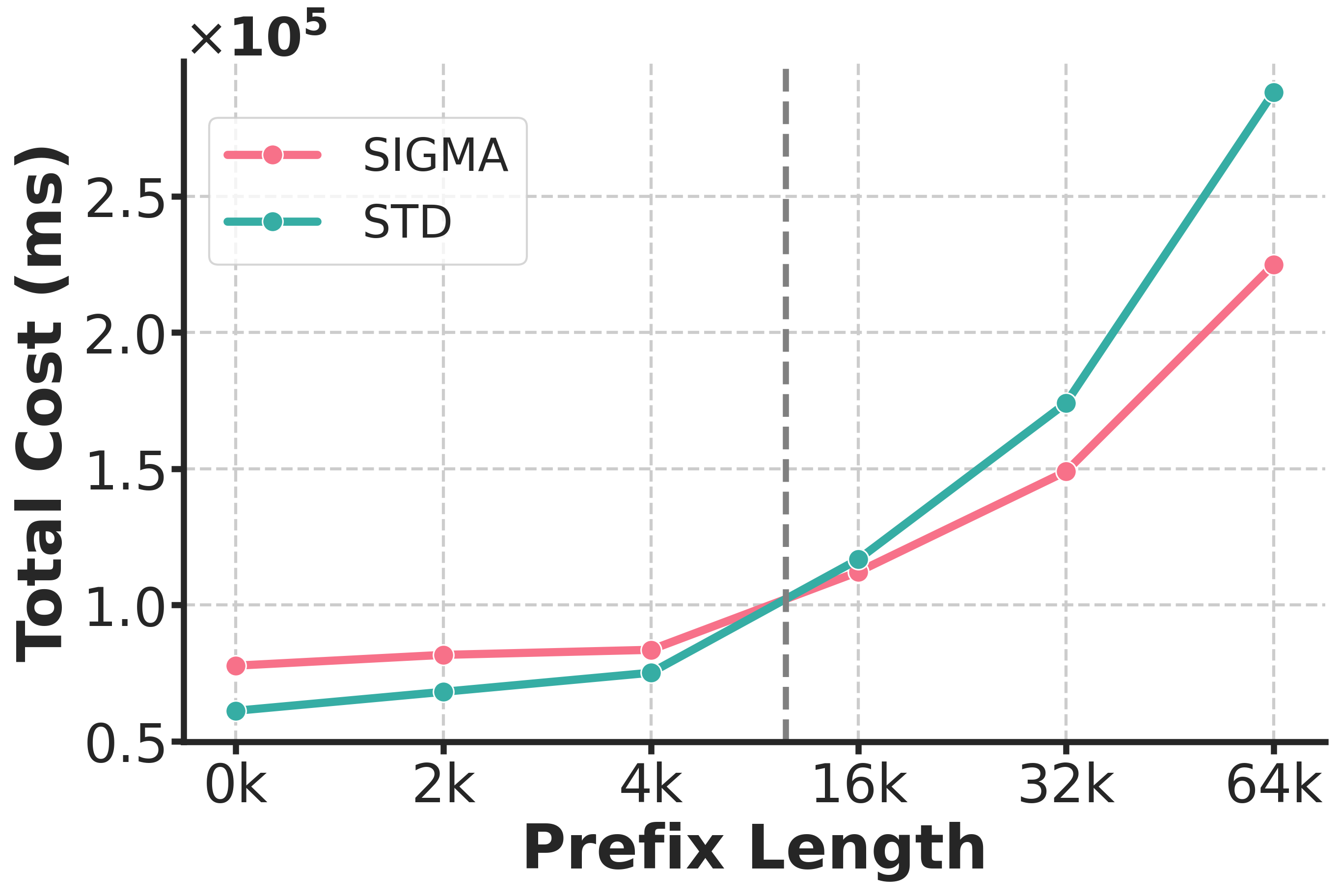}
        \caption{Output Length $= 16k$.}
        \label{fig:total_cost_d}
    \end{subfigure}
    \vspace{-3pt}
    \hfill
    \begin{subfigure}[b]{0.24\textwidth}
        \includegraphics[width=\textwidth]{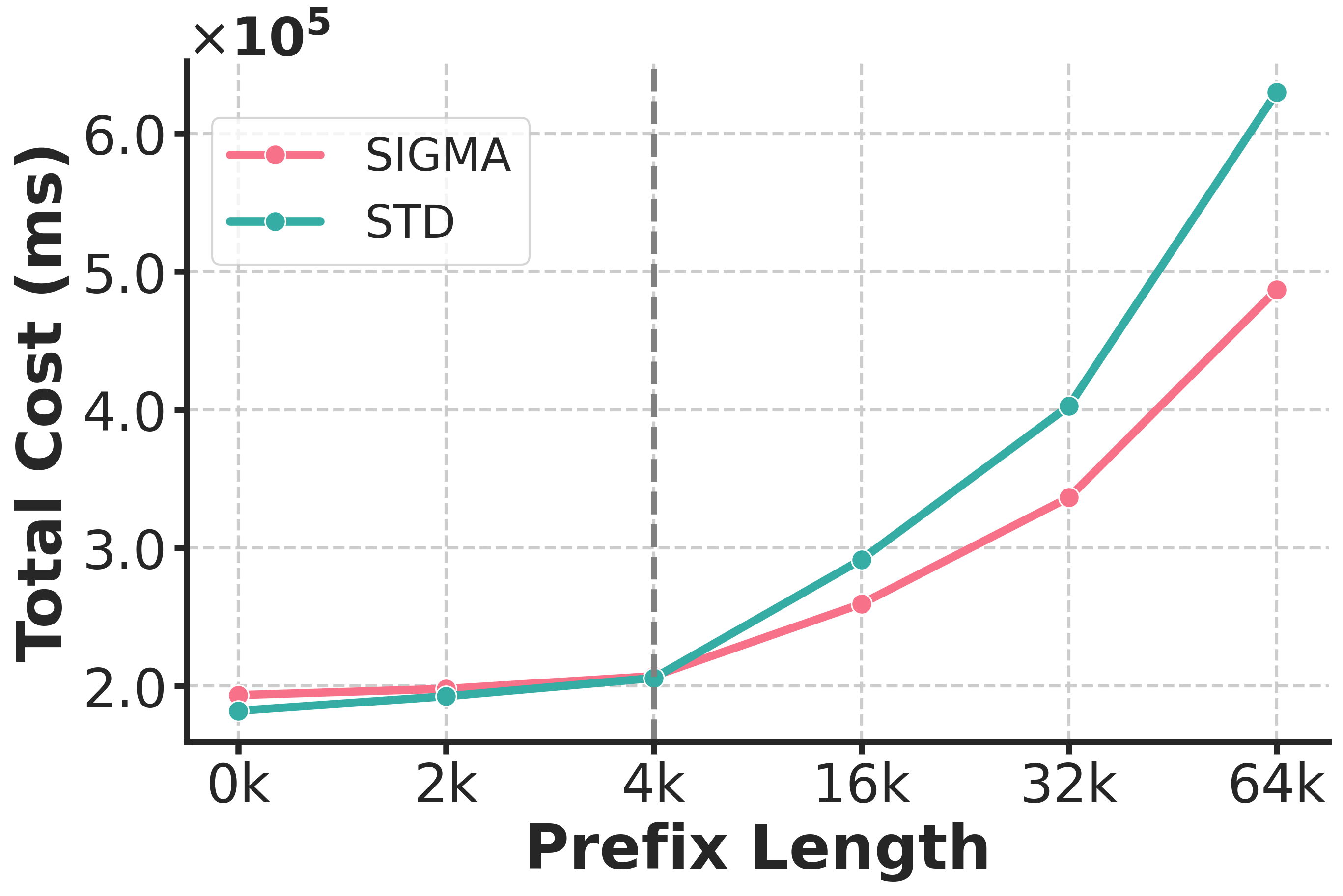}
        \caption{Output Length $= 32k$.}
        \label{fig:total_cost_e}
    \end{subfigure}
    \vspace{-3pt}
    \hfill
    \begin{subfigure}[b]{0.24\textwidth}
        \includegraphics[width=\textwidth]{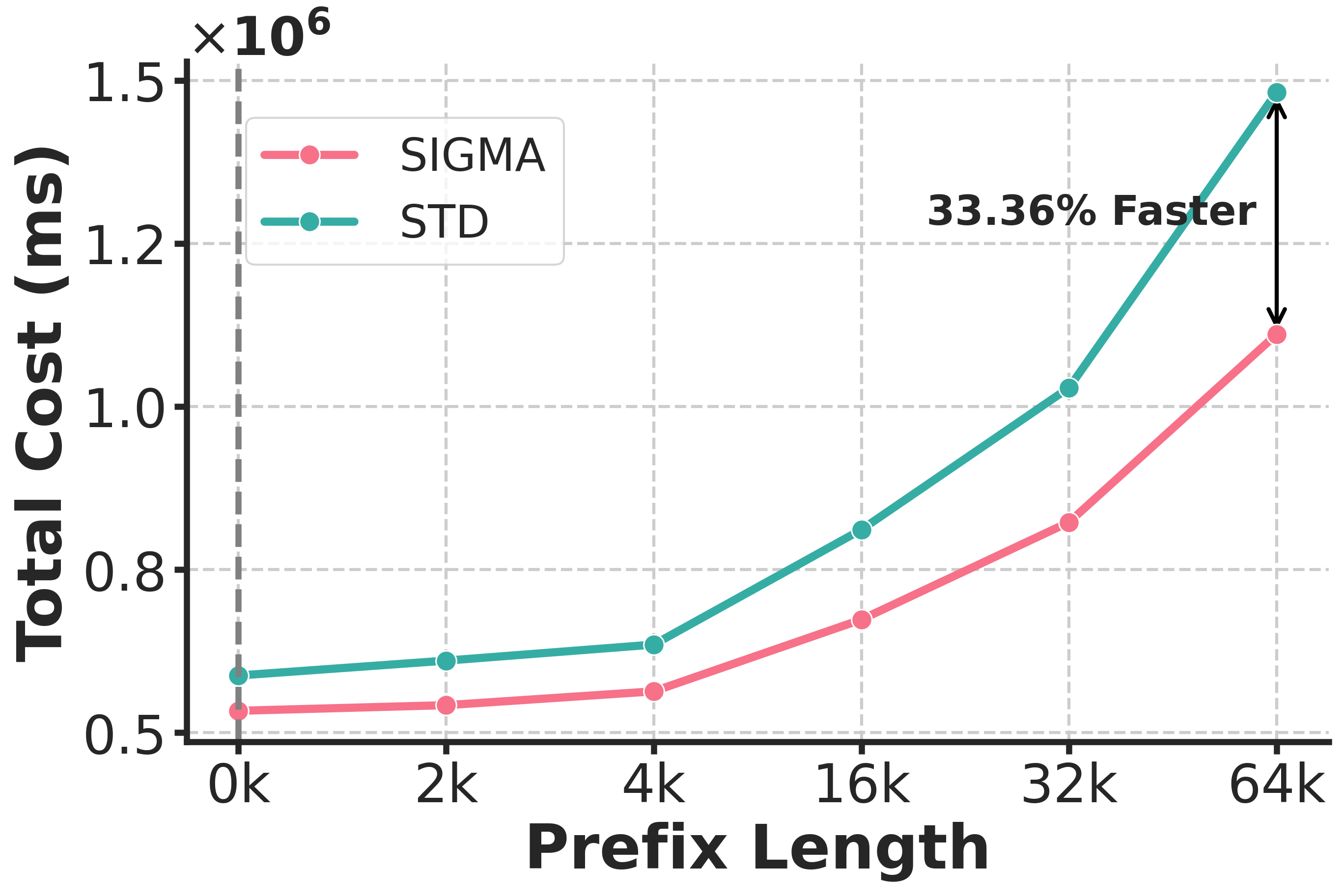}
        \caption{Output Length $= 64k$.}
        \label{fig:total_cost_f}
    \end{subfigure}
    \vspace{-3pt}
    \hfill
    \begin{subfigure}[b]{0.24\textwidth}
        \includegraphics[width=\textwidth]{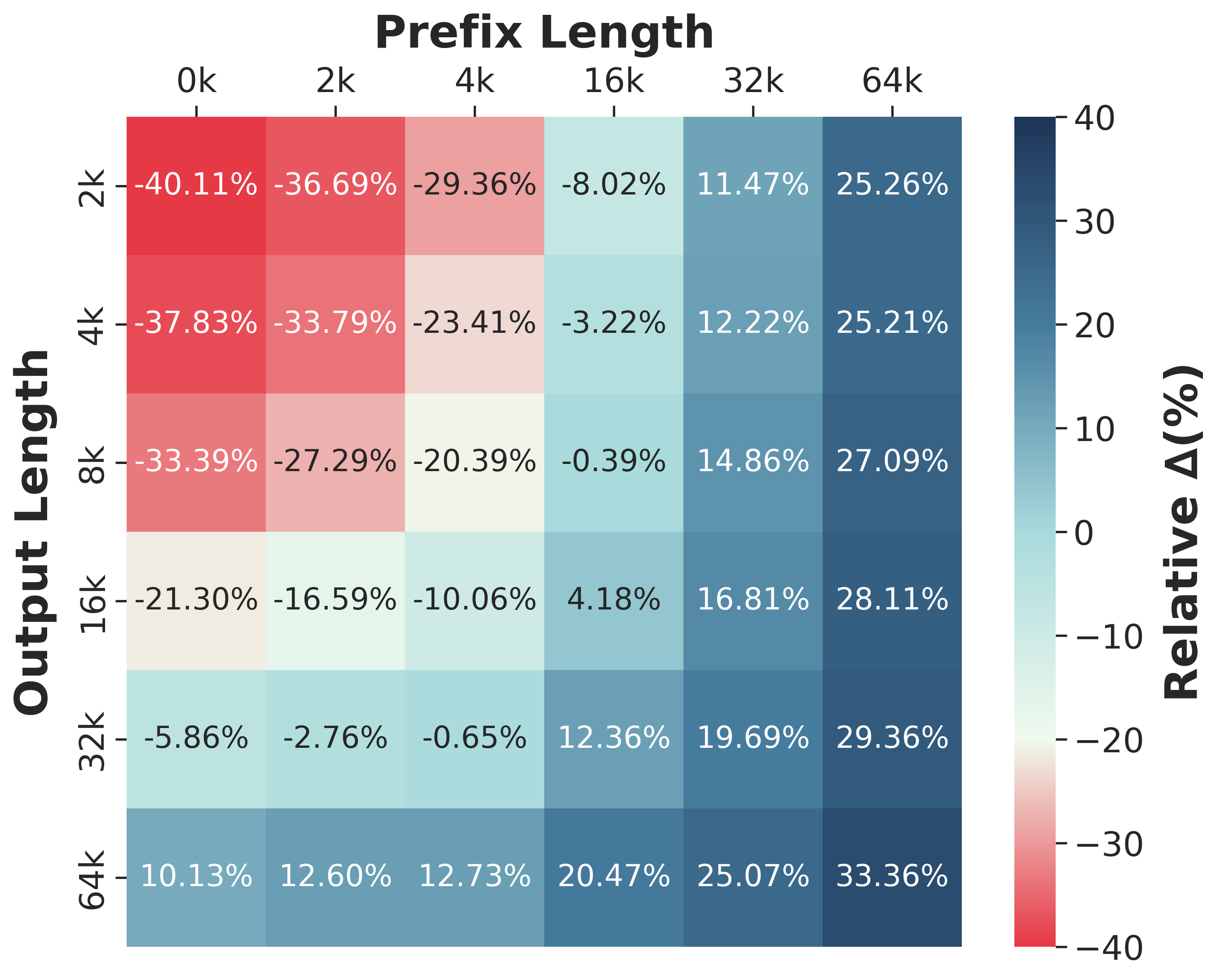}
        \caption{Relative CEET Improvment.}
        \label{fig:total_cost_h}
    \end{subfigure}
    \vspace{-3pt}
    \caption{Comparison of total CEET cost between Standard model(\textsc{Std}) and \textsc{Sigma}. The gray dashed line indicates where the inference costs of both models are equal. As the output length increases, this intersection point moves progressively earlier.}
    \vspace{-6pt}
    \label{fig:total_cost}
\end{figure*}
\noindent\textbf{Flexible Attention Computation.} The official FlashAttention supports attention computations only when $n^h_k$ equals $n^h_v$, and $n^h_q$ is an integer multiple of $n^h_k$ and $n^h_v$, which could not be directly applied to \textsc{Sigma}.
\if\isaccepted{To address this issue, we introduce FlexHeadFA, which leverages \textbf{address probing} to support attention computation with a flexible number of heads. Specifically, FlexHeadFA's implementation follows that of FlashAttention2, consisting of two major GPU kernel operations \lstinline[basicstyle=\ttfamily, breaklines=true]{flash_fwd_splitkv_kernel} and \lstinline[basicstyle=\ttfamily, breaklines=true]{flash_fwd_splitkv_combine_kernel}.}
\else{To address this issue, we introduce FlexHeadFA\footnote{\url{https://shorturl.at/dpzaX}}, which leverages \textbf{address probing} to support attention computation with a flexible number of heads. Specifically, FlexHeadFA's implementation follows that of FlashAttention2, consisting of two major GPU kernel operations \lstinline[basicstyle=\ttfamily, breaklines=true]{flash_fwd_splitkv_kernel} and \lstinline[basicstyle=\ttfamily, breaklines=true]{flash_fwd_splitkv_combine_kernel}.}
The \textbf{split kernel} handles primary attention computations by dividing key and value matrices into chunks, performing multiplications, and generating output chunks. FlexHeadFA separates address calculations for key and value heads, using the query head index $\texttt{idx}_q$ to determine the key and value head indices ($\texttt{idx}_k$, $\texttt{idx}_v$) via \cref{eq:idx}.
\vspace{-.1cm}
\begin{equation}
\vspace{-.25cm}
\texttt{idx}_{i}=\texttt{idx}_q\cdot\frac{n^h_i}{n^h_q}, i\in\{k,v\}.\label{eq:idx}
\end{equation}
This design enables efficient element retrieval for multiplication, removing the constraint of equal head numbers. The \textbf{combine kernel} then assembles these chunks into the final output matrix. 

\subsection{Empirical Analysis}

\noindent\textbf{Experiment Setup.} To demonstrate the efficiency gains achieved by reducing $n^h_k$, we compare \textsc{Sigma} 1.5B with a hypothetical baseline model, referred to as the standard model (\textsc{Std}). To be specific, while \textsc{Sigma} features unbalanced KV heads ($n_k^h=4$ and $n_v^h=16$) and includes an additional augmented Q ($d_q^h=3072$), \textsc{Std} employs a standard Group Query Attention (GQA) mechanism, configured with balanced K and V heads ($n_k^h=16$ and $n_v^h=16$).
To accurately measure efficiency improvements, we employ two metrics: \textbf{Cuda Event Elapsed Time (CEET)} and \textbf{Kernel Execution Time (KET)}. 
CEET measures elapsed time of certain operations by placing checkpoints within the code. While it may be unstable at small absolute time scales, it is well-suited for large-scale experiments. In this test, we record the cost of storing and loading \textbf{KV Cache}, applying \textbf{Attention Computation}, and incorporating \textbf{Augmented Q} (only in \textsc{Sigma}), treating their summation as the total cost of attention layers. Both \textsc{Sigma} 1.5B and \textsc{Std} are evaluated with output and prefix lengths independently increasing in grid patterns of [2k, 4k, 8k, 16k, 32k, 64k] and [0, 2k, 4k, 16k, 32k, 64k], respectively.
KET provides a precise time estimation by recording kernel execution times using \texttt{nsys}. Here, we measure the cost of the combine kernel and the split kernel, treating their summation as the overall cost of applying FlexHeadFA. Although this method ensures high accuracy, it incurs significant memory overhead as the inference scale increases. Therefore, both models are evaluated with 10 output length and prefix length varying in a grid pattern: [2k, 4k, 16k, 32k].
All experiments are conducted on a \texttt{NVIDIA H100 80G HBM3} GPU, with details in \cref{sec:eff_r}.


\noindent\textbf{KET Results.} We present KET results in \cref{fig:kernel_cost} with details recorded in \cref{tab:sigma_eff_k}. The results align closely with our previous analyses. The split kernel, which directly operates on the key and value matrices, benefits naturally from the reduction in key heads in terms of memory load cost. Both its absolute cost and relative improvement increase sharply as the prefix length grows. In contrast, the combine kernel, responsible soly for combining output chunks, is unaffected by changes in the key matrix. Consequently, its improvement ratio consistently stays near zero. The relative improvements for both the split kernel and the total cost become increasingly significant with longer prefix lengths, and we anticipate the absolute improvement to approach 37.5\% as the prefix length continues to expand.

\noindent\textbf{CEET Results - Total Cost.} 
Since \textsc{Std} and \textsc{Sigma} share a largely similar design, the combined cost of above three modules is treated as the total cost (excluding Augmented Q for \textsc{Std}). As illustrated in \cref{fig:total_cost}, \textsc{Sigma} becomes increasingly efficient with the increase in output length and prefix length. Specifically, while \textsc{Sigma} incurs higher inference costs than \textsc{Std} for shorter prefix lengths in \cref{fig:total_cost_a}, the cost crossover point shifts earlier as output length increases. For example, \textsc{Sigma} surpasses \textsc{Std} only when prefix lengths exceed 16k tokens for 2k outputs, but it outperforms \textsc{Std} even without a prefix when generating 64k tokens, reducing total cost by up to 33.36\% for 64k outputs (\cref{fig:total_cost_f}). Detailed cost analyses of the three modules can be found in \cref{sec:dt_eff}. 

In summary, while \textsc{Sigma} is slightly less efficient than the standard model in short-context scenarios ($\sim10^3$ ms), it shows significant advantages in long-context scenarios ($\sim10^6$ ms), which are common and critical in practical applications. This makes \textsc{Sigma} promising for advancing test-time scale-up in LLM, enhancing their scalability and efficiency for more complex, extended inference processes. 


\section{System Domain Pre-training and \textsc{AIMicius} Benchmark}
\label{sec:sigma-system}
In our work, we pre-train \textsc{Sigma} from scratch to be a specialized model tailored for system-related tasks, called \textsc{Sigma-System}, with our meticulously collected domain data. Specifically, our goal is to create LLMs that can automatically \textbf{diagnose AI infrastructure issues and profile AI workloads} (\textit{e.g.}, being able to generate command lines to monitor GPU utilization). These features would allow the LLMs to oversee the training processes of neural models, even including their own, and support automated optimization. 
To achieve this, we compile system-specific pre-training and fine-tuning data and introduce \textsc{AIMicius}, a benchmark for system domain evaluation.




\label{sec: system-data-collection}
\noindent\textbf{System Domain Data Collection.}
We identify 15 primary source categories from over 120 system-related websites to collect data. 
In particular, to enrich general system knowledge, we gather data from a diverse range of sources, including academic papers from arXiv and renowned conferences or journals. StackOverflow is the main source for enhancing debugging skills, while technical blogs and developer forums are crucial for refining system design capability. Additionally, we collect data from related websites to include knowledge on Azure VM, hardware abstraction, Linux commands, GitHub issues, and Stack Exchange. These sources make up the bulk of our pre-training data, supplemented by several other minor resources.
As these data sources exhibit varied formats, we process each source individually to extract publicly available system data. To efficiently cleanse the substantial data, we utilize LLM labeling and various AI tools automatically, including category classification, quality control, and data format conversion. For instance, we employ GPT-3.5 to categorize the data items from StackOverflow, subsequently training a smaller model to classify the remaining data, thereby saving on costs. Ultimately, we gather approximately 19.5B tokens for pre-training. 
\cref{tab:sigma_data} presents detailed statistics of our collected data.

\begin{table}[t]
\centering
\small
\vspace{-2mm}
\caption{\textsc{Sigma}'s system domain pre-training data composition.}
\vspace{0.2em}
\resizebox{0.9\linewidth}{!}{
\begin{tabular}{llrr}
\toprule
\textbf{Data Type} & \textbf{Sources} & \multicolumn{1}{c}{\textbf{Size}} & \textbf{\#$\;$Tokens} \\
\midrule
\multirow{2}{*}{General System} & CCF Ranking list & 14.0 G & 3.3 B \\ 
& arXiv & 33.0 G & 5.4 B \\
\midrule
\multirow{1}{*}{Design Capability} & Blogs \& Forums & 14.5 G & 3.2 B \\ 
\midrule
\multirow{1}{*}{Debug Capability} & Stack Overflow & 38.9 G & 7.6 B \\
\bottomrule
\vspace{-9mm}
\end{tabular}}
\label{tab:sigma_data}
\end{table}

\label{sec: aimicius}
\noindent\textbf{\textsc{AIMicius} Benchmark.}
Currently, \textsc{AIMicius} comprises four tasks: CMDGen, Infrawise, Optiflow, and NL2KQL, with most data sourced from Azure service. Details on evaluation and examples are provided in \cref{sec:detail_system_benchmark,sec:data examples}. 
\label{sec:cmdgen}
\begin{itemize}

\item \textbf{CMDGen} focuses on command-line generation for the NVIDIA and AMD platforms. The target is to generate optimal commands to address specific GPU-related challenges described in the prompts. The commands in CMDGen can be categorized into seven distinct subgroups: \texttt{NCCL}, \texttt{Nvidia-smi}, \texttt{NVCC}, \texttt{RCCL}, \texttt{Rocm-smi}, \texttt{Superbench}, and a general \texttt{others} category.
The data in CMDGen are sourced from a variety of origins, ensuring diversity and realism. Some examples are curated from official documentation, paired with human-written or LLM-generated queries for context; others are directly extracted from Azure service logs or websites like StackOverflow to capture real-world usage patterns. It comprises a total of 2,838 instruction-tuning examples and 395 test cases, with 200 for the NVIDIA platform and 195 for the AMD platform. 

\item \textbf{Infrawise} aims at retrieving benchmark results of a particular model in terms of its infrastructure-wise performance (\textit{e.g.} the inference speed of GPT-3 on a single A100). Performing this task generally involves two critical steps: \textbf{DCW Generation} and \textbf{Benchmark Result Retrieval}. The dataset of Infrawise comprises 422 instruction-tuning samples and 911 end-to-end test cases. 

\item \textbf{Optiflow} targets for optimizing network topology and data flow within a specified multi-GPU SKU and data size to minimize \texttt{all-gather} latency. It is further divided into two subtasks: (1) \textbf{Plan Generation} and (2) \textbf{Plan Improvement}. In Plan Generation, the LLM generates Python codes that represent the optimized network topology and data flow for the given multi-GPU configuration. In Plan Improvement, users provide the latency of their current design, and the LLM is tasked with refining the existing plan to deliver improved code with reduced latency. This task challenges the model’s ability to reason about complex hardware configurations and effectively deliver actionable, efficient solutions. Optiflow comprises a total of 8,000 instruction-tuning examples and 1,258 test cases. Among the instruction-tuning examples, 5,047 are derived from the Plan Generation subtask, while 2,953 focus on the Plan Improvement subtask. The test cases are structured in an end-to-end manner: LLM is first tasked with generating a plan based on the input, followed by refining the plan given the current latency. 

\item \textbf{NL2KQL} is for converting the user instruction in the form of natural language into Kusto Query Language (KQL)\footnote{\url{https://shorturl.at/cvIOc}}, which is specifically designed for querying and analyzing large datasets in Azure Data Explorer and other Microsoft services, such as Log Analytics and Application Insights. The NL2KQL dataset includes a total of 5,166 instruction-tuning examples and 43 test cases. 
\end{itemize}

\section{Performance Evaluations}
\noindent\textbf{Pre-training Settings.}
The pre-training data, totaling 6T tokens, consists of general domain data and domain-specific property data.
For general domain data, we combine DCLM and FineWeb-EDU, remove duplicates to get General Dataset I (about 4T tokens). After quality filtering, the result is General Dataset II (1T tokens). Then, we select higher-scoring data with stricter rules for General Dataset III (about 200B tokens for the annealing phase of \textsc{Sigma}'s pre-training).
For math domain data, we use proof-pile-2 and combine it with 280 billion math-related data from General Dataset I.
For code domain data, following StarcoderV2's filtering method, we select a 500B token dataset.
We also have about 1T tokens of synthesized and rewritten pre-training data, which have passed quality screening and contain multi-domain content, for the later phase of \textsc{Sigma} pre-training. 
More details can be found in \cref{sec:detailed_pretraining_settings}.

\begin{table}[t]
\centering
\small

\setlength{\tabcolsep}{2pt}
\caption{Evaluation results on the \textsc{AIMicius} benchmark. The baselines include GPT-4, Deepseek-Coder-7b-Instruct-v1.5, Qwen2.5-Coder-7B-Instruct, and LLaMA3-8B-Instruct. All metrics are normalized to a scale of 0 to 100, with higher values indicating better performance. \textbf{Bolded metrics} represent the most critical evaluation criteria for each task. \textsc{Sigma-System} 10B is fine-tuned (SFT) using our proprietary dataset.}
\vspace{0.5em}
\resizebox{\linewidth}{!}{
\begin{tabular}{*{2}{l}*{6}{c}}
\toprule
\bf Task & \bf Metric ($\uparrow$) & \bf GPT-4 & \bf DeepSeek & \bf Qwen & \bf Llama & \bf 
\textsc{Sigma} \\
\midrule
\multirow{6}{*}{\makecell[l]{\bf CMDGen \\ \bf NVIDIA}}
& CMD Score & \underline{84.0}  & 59.4 & 81.1 & 33.6 & \bf{87.5} \\ 
& Output Score & \underline{61.0} & 12.9 & 38.3 & 12.0 & \bf{80.9} \\
& Calibration Score & \underline{62.0}  & 0.5 & 52.0 & 5.0 & \bf{78.0} \\
& Exact Match & \underline{13.0}  & - & - & - & \bf 57.0 \\
& Success Ratio & \underline{21.0}  & 0.5 & 15.5 & 3.5 & \bf{74.0} \\ 
& \bf Accuracy & \underline{25.0} & 0.5 & 15.5 & 3.5 & \bf{74.5} \\
\midrule
\multirow{6}{*}{\makecell[l]{\bf CMDGen \\ \bf AMD}}
& CMD Score & 73.0 & 55.0 & \underline{77.2} & 31.4 & \bf 88.9 \\
& Output Score & 49.0  & 20.3 & \underline{49.1} & 14.1 & \bf 78.0 \\
& Calibration Score & 43.0  & 3.1 & \underline{51.3} & 2.6 & \bf 79.3 \\
& Exact Match & \underline{14.0}  & - & - & - & \bf 53.9 \\
& Success Ratio & 13.0 & 3.1 & \underline{35.8} & 1.6 & \bf 69.4 \\ 
& \bf Accuracy & 17.0  & 3.1 & \underline{35.8} & 1.6 & \bf 69.4 \\
\midrule
\multirow{7}{*}{\bf Infrawise}
& Target & 40.7  & 43.5 & \underline{44.9} & 28.1 & \bf 95.2 \\
& Baseline & 34.1 & \underline{43.6} & 42.7 & 34.5 & \bf 92.9 \\
& Criterion & \underline{55.1}  & 36.3 & 44.2 & 11.1 & \bf 75.1 \\
& Workload & \bf 52.1 & 39.9 & 47.8 & 14.2 & \underline{48.4} \\
& DCW & \underline{22.5}  & 16.6 & 20.8 & 4.6 & \bf 40.3 \\
& Bench. Recall & 19.8 & 11.6 & \underline{20.4} & 6.5 & \bf 28.3 \\
& \bf Bench. Acc. & \underline{18.7} & 11.6 & 12.6 & 6.5 & \bf 28.3 \\
\midrule
\multirow{4}{*}{\bf Optiflow}
& Code Detected & 95.8  & 82.2 & \bf 100.0 & \underline{98.2} & \bf 100.0 \\
& Code Executable & 50.3 & 18.6 & 28.3 & \underline{51.2} & \bf{85.9} \\
& \bf Plan Valid & 16.8 & 18.6 & \underline{28.3} & 16.4 & \bf{86.7} \\
& \bf Plan Improved & 0.5  & - & - & \underline{1.1} & \bf 66.7 \\
\midrule
\multirow{6}{*}{\bf NL2KQL}
& \bf Syntax Acc. & \bf 100.0 & 81.4 & \bf 100 & 83.7 & \bf 100.0 \\
& Similarity & 31.8 & 33.9 & \bf 36.7 & 33.6 & \underline{34.9} \\
& \bf Cluster Score & - & - & - & - & \bf{43.0} \\
& \bf Database Score & 2.3  & 2.3 & \underline{4.7} & 2.3 & \bf{40.7} \\
& Table Score & \underline{4.7} & \underline{4.7} & \underline{4.7} & \underline{4.7} & \bf{17.4} \\
& Column Score & 17.3  & 19.3 & \underline{22.2} & 20.8 & \bf{29.0} \\
\bottomrule
\vspace{-6mm}
\end{tabular}
}
\label{tab:sigma_system_b}
\end{table}

\begin{table*}[t]
\centering
\small

\caption{Comparisons with baseline models on commonsense reasoning and text understanding tasks. Differences with original reports in the baseline models are due to our unified re-evaluations for fair comparisons.}
\vspace{0.5em}
\label{tab:main_result}
\resizebox{0.8\linewidth}{!}{
\begin{tabular}{l c c ccccccccc}
\toprule
\multirow{2}{*}{\textbf{Model}} & \multirow{2}{*}{\textbf{Params}} & \multirow{2}{*}{\textbf{Avg.}} & \multicolumn{7}{c}{\textbf{Commonsense \& Comprehension}} & \multicolumn{2}{c}{\textbf{Continued}}\\ \cmidrule(lr){4-10}    \cmidrule(lr){11-12} 
&&& {Hella.} & {ObQA} & {Wino.} & {ARC.e} & {ARC.c} & {PIQA}  & {SciQ} & {Bool.} & {Logi.}\\
\midrule  
Pythia & 1.0B & 49.3 & 47.1 & 31.4 & 53.4 & 49.0 & 27.1 & 69.3 & 76.1 & 60.8 & \underline{29.8}\\
TinyLlama & 1.1B & 54.0 & 61.5 & 36.8 & 59.5 & 55.6 & 32.7 & 73.6 & 84.2 & 56.0 & 25.8\\
Bloom & 1.1B & 46.5 & 43.0 & 29.4 & 55.0 & 45.5 & 25.6 & 67.2 & 74.5 & 59.1 & 18.9\\
OLMo & 1.2B & 54.5 & 63.0 & 36.2 & 59.9 & 57.3 & 30.9 & 75.1 & 78.7 & 61.8 & 27.8\\
OPT & 1.3B & 51.2 & 53.7 & 33.2 & 59.8 & 51.0 & 29.5 & 72.4 & 76.7 & 57.7 & 26.9\\ 
CerebrasGPT & 1.3B & 46.5 & 38.4 & 29.0 & 52.1 & 45.8 & 25.3 & 66.8 & 73.0 & 59.4 & 29.2\\
Phi1 & 1.3B & 39.6 & 30.4 & 25.0 & 49.9 & 34.6 & 23.5 & 56.0 & 64.5 & 45.2 & 27.3\\
Pythia & 1.4B & 51.7 & 52.0 & 33.2 & 57.3 & 53.9 & 28.3 & 70.9 & 79.3 & 63.3 & 27.5\\
DCLM & 1.4B & \textbf{62.8} & \textbf{71.6} & \textbf{42.6} & \underline{66.2} & 71.6 & 43.5 & 77.7 & 90.7 & \underline{71.4} & \underline{29.8}\\
StableLM2 & 1.6B & 61.4 & 69.0 & 38.8 & 63.6 & 68.2 & 38.9 & 76.6 & \textbf{95.3} & \textbf{74.7} & 27.2\\
SmolLM & 1.7B & 61.0 & 65.7 & \underline{42.0} & 60.8 & \underline{73.5} & \textbf{46.3} & 76.1 & 89.6 & 66.0 & 28.7\\
Gemma & 2.0B & \underline{62.2} & \underline{71.4} & 40.0 & 64.6 & \underline{72.3} & 41.8 & \textbf{78.2} & 91.5 & 69.3 & \textbf{30.4}\\ \midrule
\textsc{Sigma} (Ours) & 1.5B & 61.6 & 67.3 & 40.4 & \textbf{67.5} & \underline{72.3} & \underline{43.9} & \underline{77.8} & \underline{94.0} & 63.0 & 28.4\\
\bottomrule
\vspace{-4mm}
\end{tabular}
}
\end{table*}

\label{sec: system-performance}
\noindent\textbf{System Domain Performance.}
We pre-train \textsc{Sigma-System}-10B on our system domain pre-training data and fine-tuned with full-parameter updates on our SFT dataset, tailored individually for each task. 
We consider the following models as the compared baselines: \textbf{GPT-4} \cite{bubeck2023sparks}, DeepSeek-Coder-7b-Instruct-v1.5 (\textbf{DeepSeek}) \cite{xin2024deepseek}, Qwen-Coder-7B-Instruct (\textbf{Qwen}) \cite{yang2024qwen2}, and LLaMA3-8B-Instruct (\textbf{LLaMA}) \cite{dubey2024llama}, to compare them with \textsc{Sigma-System} on the complete \textsc{AIMicius} benchmark. 

As shown in \cref{tab:sigma_system_b},  
\textsc{Sigma-System} showcases superior performance across all tasks in the \textsc{AIMicius} benchmark, substantially outperforming the baseline models. Specifically, \textbf{compared to the second-best model, its absolute improvement is as significant as 49.5\%, 33.6\%, 9.6\%, 65.6\%, and 36\%} on the major metrics of the CMDGen NVIDIA, CMDGen AMD, Infrawise, Optiflow, and NL2KQL tasks, respectively. 
Its high scores in various metrics indicate its remarkable ability to generate high-quality commands, monitor infrastructure effectively, and perform well in other system-related tasks. Though the baseline models excel in general domains, they struggle in \textsc{AIMicius}, especially in CMDGen and Optiflow tasks, highlighting the importance of domain-specific advancements for LLMs.

Although our model surpasses the baselines in nearly all metrics, it is crucial to recognize that its absolute performance in specific tasks, such as Infrawise and NL2KQL, is still subpar.
In Infrawise, \textsc{Sigma-System} performs exceptionally in terms of the Target, Baseline, Criterion scores. Nonetheless, its other metrics do not exceed 50\%, with Benchmark Result Accuracy standing at 28.3\%. The performance of other models on this task is even more disappointing. 
This indicates that there is room for further improvement, particularly in scenarios requiring more nuanced understanding and precise decision-making. As part of our future work, we plan to enrich the training dataset and refine the data generation policy to better capture the complexity and variability of these tasks, aiming to achieve enhanced performance across all evaluation metrics.

\label{sec:sigma-general}
\noindent\textbf{General Domain Performance.}
We pre-train a general domain model using the \textsc{Sigma}-1.5B architecture to investigate its performance in more settings. 
We consider following benchmarks for evaluation of  commonsense reasoning and text-understanding ability: HellaSwag (Hella.)~\citep{zellers2019hellaswag}, OpenBookQA (ObQA)~\citep{OpenBookQA2018}, WinoGrande (Wino.)~\citep{sakaguchi2021winogrande}, ARC easy/Challenge (ARC.e/c)~\citep{clark2018think}, PIQA~\citep{bisk2020piqa}, SciQ~\citep{welbl2017crowdsourcing}, BoolQ (Bool.)~\citep{clark2019boolq}, and LogiQA (Logi.)~\citep{liu2020logiqa}. 
We compare \textsc{Sigma} against the following state-of-the-art models with similar scales: TinyLLaMA-1.1B~\citep{zhang2024tinyllama}, Pythia-(1.0B, 1.4B)~\citep{biderman2023pythia}, OPT-1.3B~\citep{zhang2022opt}, Bloom-1.1B~\citep{muennighoff2022crosslingual}, Cerebras-GPT-1.1B~\citep{dey2023cerebras}, Phi1-1.3B~\citep{gunasekar2023textbooks}, StableLM-2-1.6B~\citep{bellagente2024stable}, SmolLM-1.7B~\citep{allal2024SmolLM}, DCLM-1.4B~\citep{li2024datacomp}, OLMo-1.2B~\citep{groeneveld2024olmo}, and Gemma-2B~\citep{team2024gemma}. All results are obtained via zero-shot prompting.

As shown in \cref{tab:main_result}, \textsc{Sigma}-1.5B achieves an outstanding average performance of 61.6 on 9 benchmarks, significantly outperforming strong baseline models such as OLMo-1.2B and TinyLLaMA-1.1B on various reasoning and comprehension tasks. It also achieves comparable performance to state-of-the-art large language models such as Gemma-2B, and DCLM-1.4B. Specifically, \textsc{Sigma} achieves top-2 performance at WinoGrande, PIQA, ARC.e/c and SciQ, showing its broad and accurate grasp of both intuitive and logical commonsense during its pre-training on large-scale text sequences, which provides reliable knowledge backups for reasonable text completions and further task-specific tuning. Although it excels at common sense reasoning, \textsc{Sigma} shows limited performance on BoolQ and LogiQA, reflecting a moderate level of reading comprehension on option-form questions. We observe a significant decrease in these benchmarks during the annealing stage when math problem-solving ability improves, showing a potential conflict between natural and formal language understanding for small-scale models. 
Additional evaluation results on problem-solving tasks are presented in \cref{sec:appendix_problem_solving}.
\section{Conclusion}


We introduce \textsc{Sigma}, an efficient LLM specialized for the system domain. 
Its architecture features a novel attention module that called DiffQKV, which is equipped with augmented Q for performance improvement and differential KV compression for enhanced inference efficiency. 
Through theoretical and empirical analyses, we demonstrate the competitive efficiency of DiffQKV, with up to 33.36\% speed improvement compared with Grouped-Query Attention(GQA) in long-context scenarios.
Moreover, after pre-training on 6 trillion tokens, \textsc{Sigma} demonstrates performance on par with the latest state-of-the-art models across general domains.
On the \textsc{AIMicius} benchmark, the first comprehensive system domain benchmark that we propose, \textsc{Sigma} substantially surpass all baseline models across all tasks, achieving an absolute improvement up to 52.5\%.







\section*{Impact Statement}

This paper presents work whose goal is to advance the field of 
Machine Learning. There are many potential societal consequences 
of our work, none which we feel must be specifically highlighted here.


\bibliography{reference}
\bibliographystyle{icml2025}

\newpage
\appendix
\onecolumn
\newpage

\section{Related Work}
Over the past few years, Large Language Models (LLMs) have exerted a considerable impact across various domains \citep{bubeck2023sparks, jiang2023mistral, glm2024chatglm, dubey2024llama, yang2024qwen2}. 
Nevertheless, the formidable computational demands and memory requirements of LLM inference present considerable challenges for deployment in many resource-constrained scenarios.
Thus, while the exploration into scaling up a larger model scale to achieve even more advanced level of intelligence is still ongoing \citep{kaplan2020scaling, aghajanyan2023scaling, brown2024large}, 
the development of smaller, more efficient language models has also garnered increasing research interest due to their significantly reduced inference costs and lower deployment requirements \citep{hu2024minicpm, zhang2024tinyllama, abdin2024phi, lu2024small, liu2024mobilellm, team2024gemma, dai2024deepseekmoe,mehta2024openelm}. 
Meanwhile, a variety of methods have been explored to improve the inference efficiency of LLMs \citep{kwon2023efficient,ainslie2023gqa,xiao2023efficient,mu2024learning,zhou2024survey}. Significant efforts have been dedicated to addressing inference bottleneck associated with KV cache \citep{ainslie2023gqa, kwon2023efficient,luohe2024keep,zhang2024h2o}. KV cache is
a common technique adopted by the decoder-only Transformer architecture \citep{vaswani2017attention}, which stores the Key (K) and Value (V) vectors in the attention operation for future reuse at each decoding step to avoid recomputation. 
It can consume a substantial amount of memory that linearly increases with the sequence length \citep{pope2023efficiently}. This consumption practically limits the context length that the model can handle. In addition, repeatedly loading KV cache also places substantial demands on memory bandwidth, which is recognized as the major bottleneck for LLM inference speed \citep{shazeer2019fast,ribar2024sparq}. Notably, most prior studies tend to treat the compression of K and V vectors uniformly, both in terms of the optimization methodology and the compression ratio. 
For instance, Grouped-Query Attention (GQA) \citep{ainslie2023gqa} reduces the number of K and V heads in the attention layer to the same extent by organizing Query (Q) heads into groups, with each group sharing a single K/V head. 
The same applies to methods based on KV cache eviction, which carefully select a fixed number of tokens and cache both their K and V vectors \citep{beltagy2020longformer,xiao2023efficient,han-etal-2024-lm,liu2024scissorhands, zhang2024h2o,ge2024model,ribar2024sparq}. 
However, the different roles K and V vectors play in the attention mechanism might allow for differential treatment in their efficiency optimization, an area that remains underexplored. 
Moreover, existing research rarely takes the optimization of Q into consideration when devising the optimization approach, which might also provide opportunities for further improvement in model performance and efficiency. In this study, we investigate more efficient model architectures for LLMs from the perspective of differentially adjusting the QKV components in the attention mechanism, and introduce \textsc{Sigma}, a pre-trained language model featured by its efficient inference capabilities. 

Apart from KV cache optimization, the studies on improving inference efficiency can be roughly divided into three categories: pre-training stage optimization, post-training stage optimization, and system-level optimization. 
Pre-training stage optimization approaches typically involve refinement on the model architecture \citep{shazeer2019fast,ainslie2023gqa,sun2024you,brandon2024reducing,dai2024deepseekmoe}. 
Among them, Multi-Query Attention (MQA) \citep{shazeer2019fast} and Grouped-Query Attention (GQA) \citep{ainslie2023gqa} are the two most widely-used methods.
MQA is a variant of the standard Multi-Head Attention (MHA) mechanism, where all query heads share a single key head and value head. 
GQA further generalizes MQA by using an intermediate number of shared key and value heads. 
Post-training stage optimization is usually accomplished by selectively evicting part of the KV cache via token elimination \citep{beltagy2020longformer,xiao2023efficient,han-etal-2024-lm,liu2024scissorhands, zhang2024h2o,ge2024model,ribar2024sparq}. These approaches identify and eliminate non-essential tokens by employing criteria based on the attention scores at each decoding step. Another line of research  \citep{mu2024learning, jiang2023longllmlingua} focuses on compressing prompts into a smaller number of ``gisting'' tokens, which can be cached and reused to enhance efficiency. 
The last category of approaches improves efficiency by designing systems specifically tailored to the characteristics of LLM inference \citep{kwon2023efficient,jin2024llm,ye-etal-2024-chunkattention,he2024fastdecode}. In particular, \citet{kwon2023efficient} proposed paged attention and introduced the vLLM framework, effectively mitigating the memory fragmentation issues associated with LLM inference.


\section{Detailed Illustration of DiffQKV Attention}

The overview of DiffQKV attention is presented in \cref{fig:overview}. In all experiments discussed in \cref{sec:diffQKV}, we adopt 100B tokens from FineWeb-Edu \citep{penedo2024fineweb} as the pre-training data and the model are scaled to approximately 1B parameters, which consists of 22-layer transformer blocks with a hidden dimension of 2048. 
To evaluate the model's performance on different experimental settings, we employed the following benchmarks: HellaSwag (Hella.)~\citep{zellers2019hellaswag}, OpenBookQA (ObQA)~\citep{OpenBookQA2018}, WinoGrande (Wino.)~\citep{sakaguchi2021winogrande}, ARC Challenge (ARC.)~\citep{clark2018think}, PIQA~\citep{bisk2020piqa}, SciQ~\citep{welbl2017crowdsourcing}, BoolQ (Bool.)~\citep{clark2019boolq}, LogiQA (Logi.)~\citep{liu2020logiqa}, and LAMBADA~\citep{lambada2016} (LAMB.).
\subsection{Comparison with Existing Attention Mechanisms.}
The three widely used attention mechanisms, MHA, Multi-Query Attention (MQA), and Grouped-Query Attention (GQA), can be viewed as special forms of the above algorithm. 
As shown in \cref{fig:overview}, these three methods all employ the same dimensions for Q, K, and V heads ($d_q^h=d_k^h=d_v^h$). In terms of the head number, MHA sets the same number of Q, K, and V heads ($n_q^h=n_k^h=n_v^h$), MQA only employs a single K head and V head ($n_k^h=n_v^h=1$), and GQA features an intermediate number of K and V heads, yet still aligns their values ($n_k^h=n_v^h$). 

\begin{figure*}[h]
\centering
\includegraphics[width=\linewidth]{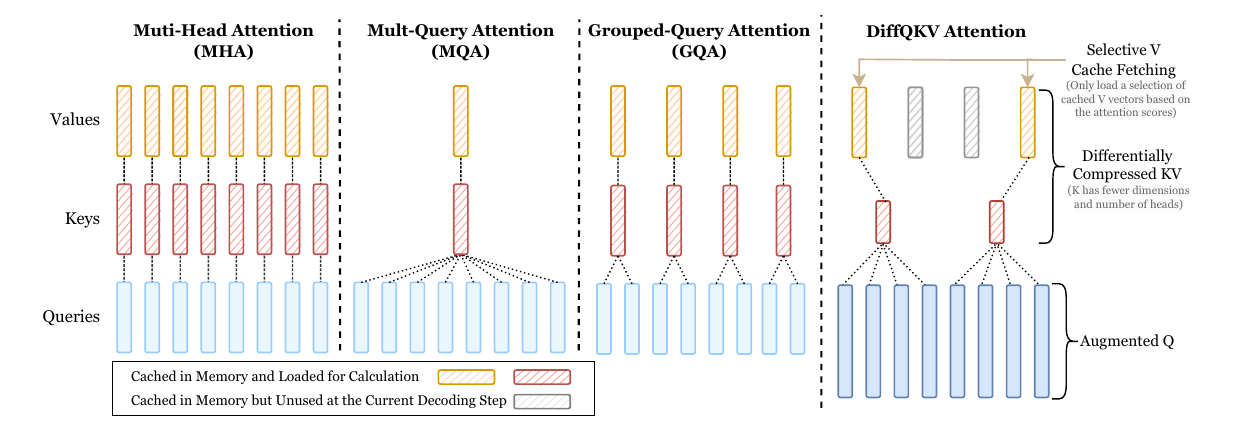}
\caption{Overview of our proposed method for differential rescaling of QKV, compared alongside Multi-Head Attention (MHA), Multi-Query Attention (MQA), and Grouped Query Attention (GQA). Specifically, our method involves: (1) differentially compressed KV: applying more aggressive compression on the number of K heads and their dimensions than on the V components, which more significantly reduces the size of K cache. We can also optionally adopt selective V cache fetching for V compression; (2) augmented Q: adopting a higher dimension for the Q head compared to the KV heads.}
\label{fig:overview}
\end{figure*}

\begin{table*}[h]
\centering
\small
\setlength{\tabcolsep}{5pt}
\caption{The ablation studies of the model performance when only selectively loading the V vectors corresponding to the highest attention scores for approximate calculation. This operation significantly enhances inference efficiency by reducing memory usage. The number of Q heads is 32 for all models in the table ($n_q^h=32$). }
\vspace{.5em}
\label{tab:topkV}
\begin{tabular}{l l ccccccccc}
\toprule
\multirow{2}{*}{\textbf{Model}} & \multirow{2}{*}{\textbf{Overall}}& \multicolumn{6}{c}{\textbf{Commonsense \& Comprehension}} & \multicolumn{2}{c}{\textbf{Continued}}& \textbf{LM} \\ \cmidrule(lr){3-8}    \cmidrule(lr){9-10}   \cmidrule(lr){11-11}    
&& {Hella.} & {ObQA} & {Wino.} & {ARC.}  & {PIQA}  & {SciQ} & {Bool.} & {Logi.} & {LAMB.} \\
\midrule
\textbf{MHA} ($n_k^h$=$n_v^h$=32) & \textbf{52.40}& 55.6 & 37.6 & 57.6 & 36.0 & 73.9 & 85.5 & 59.6 & 28.9 & 36.8 \\ 
\rowcolor{customgray!15}
+ Sel.V-top100
& \textbf{52.10} \tiny{($\downarrow$0.30)}& 55.6 & 37.6 & 57.6 & 36.0 & 74.0 & 84.8 & 59.3 & 27.0 & 36.9 \\ 
\midrule
\textbf{GQA} ($n_k^h$=$n_v^h$=16)
& \textbf{52.14}& 55.1 & 39.6 & 56.3 & 35.4 & 71.9 & 85.0 & 61.4 & 27.8 & 36.8 \\ 
\rowcolor{customgray!15}
+ Sel.V-top100 & \textbf{52.08} \tiny{($\downarrow$0.06)}& 55.2 & 39.6 & 56.3 & 35.4 & 71.8 & 84.4 & 61.6 & 27.8 & 36.7 \\ 
\midrule
\textbf{GQA} ($n_k^h$=$n_v^h$=4)
& \textbf{51.66}& 54.0 & 38.0 & 56.0 & 37.5 & 72.3 & 82.0 & 61.3 & 28.6 & 35.4 \\ 
\rowcolor{customgray!15}
+ Sel.V-top100 & \textbf{51.67} \tiny{($\uparrow$0.01)}& 54.0 & 38.2 & 55.9 & 37.5 & 72.2 & 82.0 & 61.2 & 28.6 & 35.4 \\ 
\bottomrule
\end{tabular}
\end{table*}

\subsection{Selective V Cache Fetching}
Apart from differentially compresssed KV and augmented Q, we can further optimize this part by loading V vectors selectively during inference.  
Due to the high sparsity exhibited by the attention scores \citep{xiao2023efficient, zhang2024h2o}, it can still preserve the model performance with only a small number of V vectors to approximate the results, while largely reducing the memory usage. We term this strategy as \textbf{selective V cache fetching} \citep{xiao2023efficient, zhang2024h2o}. Due to the high sparsity in attention scores, efficiency can be further enhanced by loading V vectors selectively during inference. 

\begin{table*}[h]
\centering
\small

\caption{Comparisons of model performance when reducing the same number of K heads versus V heads. The number of Q heads is 32 for all models ($n_q^h=32$). The results show that compressing the number of K heads has a relatively smaller impact on the overall model performance.}
\vspace{0.5em}
\label{tab:full_kv_head_compression}
\begin{tabular}{l l ccccccccc}
\toprule
\multirow{2}{*}{\textbf{Model}} & \multirow{2}{*}{\textbf{Overall}}& \multicolumn{6}{c}{\textbf{Commonsense \& Comprehension}} & \multicolumn{2}{c}{\textbf{Continued}}& \textbf{LM} \\ \cmidrule(lr){3-8}    \cmidrule(lr){9-10}   \cmidrule(lr){11-11}    
&& {Hella.} & {ObQA} & {Wino.} & {ARC.}  & {PIQA}  & {SciQ} & {Bool.} & {Logi.} & {LAMB.} \\
\midrule
\textbf{MHA} ($n_k^h$=$n_v^h$=32)& \textbf{52.40}& 55.6 & 37.6 & 57.6 & 36.0 & 73.9 & 85.5 & 59.6 & 28.9 & 36.8 \\ 
\rowcolor{customgray!10}
-50\% V Heads ($n_v^h$=16)
& \textbf{51.74} \tiny{($\downarrow$0.66)}& 55.5 & 39.6 & 55.0 & 35.9 & 71.6 & 85.9 & 56.9 & 28.3 & 37.0 \\ 
\rowcolor{customgray!20}
-50\% K Heads ($n_k^h$=16)& \textbf{52.83} \tiny{(\underline{$\uparrow$0.43})}& 55.1 & 38.8 & 56.4 & 35.8 & 71.9 & 85.2 & 63.6 & 29.0 & 39.7 \\ 
\midrule
\textbf{GQA} ($n_k^h$=$n_v^h$=16)& \textbf{52.14} & 55.1 & 39.6 & 56.3 & 35.4 & 71.9 & 85.0 & 61.4 & 27.8 & 36.8 \\ 
\rowcolor{customgray!10}
-75\% V Heads ($n_v^h$=4) & \textbf{51.76} \tiny{($\downarrow$0.38)}& 54.0 & 38.2 & 55.6 & 34.8 & 72.7 & 85.0 & 60.3 & 29.9 & 35.3 \\ 
\rowcolor{customgray!20}
-75\% K Heads ($n_k^h$=4) & \textbf{51.97} \tiny{(\underline{$\downarrow$0.17})}& 54.6 & 37.8 & 57.1 & 35.1 & 72.3 & 84.1 & 62.5 & 28.3 & 36.0 \\ 
\midrule
\textbf{GQA} ($n_k^h$=$n_v^h$=4)
& \textbf{51.66}& 54.0 & 38.0 & 56.0 & 37.5 & 72.3 & 82.0 & 61.3 & 28.6 & 35.4 \\ 
\rowcolor{customgray!10}
-75\% V Heads ($n_v^h$=1) & \textbf{51.03}  \tiny{($\downarrow$0.63)}& 53.5 & 38.4 & 57.0 & 35.1 & 72.1 & 82.6 & 56.9 & 28.4 & 35.1 \\ 
\rowcolor{customgray!20}
-75\% K Heads ($n_k^h$=1)
& \textbf{51.67}  \tiny{(\underline{$\uparrow$0.01})}& 53.9 & 36.2 & 58.6 & 36.9 & 71.1 & 83.5 & 60.7 & 28.7 & 35.5 \\ 

\hline
\end{tabular}
\end{table*}
\begin{table*}[h]
\centering
\small
\caption{The ablation studies of halving the K head dimension. The results indicate that this adjustment, while largely improving the inference efficiency by reducing the size of KV cache,  does not significantly compromise performance. The number of Q heads is 32 for all models ($n_q^h=32$). }
\vspace{0.3em}
\label{tab:full_k_headdim_compression}
\begin{tabular}{l l ccccccccc}
\toprule
\multirow{2}{*}{\textbf{Model}} & \multirow{2}{*}{\textbf{Overall}}& \multicolumn{6}{c}{\textbf{Commonsense \& Comprehension}} & \multicolumn{2}{c}{\textbf{Continued}}& \textbf{LM} \\ \cmidrule(lr){3-8}    \cmidrule(lr){9-10}   \cmidrule(lr){11-11}    
&& {Hella.} & {ObQA} & {Wino.} & {ARC.}  & {PIQA}  & {SciQ} & {Bool.} & {Logi.} & {LAMB.} \\
\midrule
\textbf{MHA} ($n_k^h$=$n_v^h$=32) & \textbf{52.40} & 55.6 & 37.6 & 57.6 & 36.0 & 73.9 & 85.5 & 59.6 & 28.9 & 36.8 \\ 
\rowcolor{customgray!15}
\emph{w/} Half K Dim. & \textbf{52.56} \tiny{($\uparrow$0.16)}& 55.2 & 39.4 & 56.9 & 36.9 & 72.7 & 84.1 & 63.3 & 27.8 & 36.8 \\ 
\midrule

\textbf{GQA} ($n_k^h$=$n_v^h$=16)
& \textbf{52.14}& 55.1 & 39.6 & 56.3 & 35.4 & 71.9 & 85.0 & 61.4 & 27.8 & 36.8 \\ 
\rowcolor{customgray!15}
\emph{w/} Half K Dim.  & \textbf{52.06} \tiny{($\downarrow$0.08)} & 54.3 & 39.8 & 56.9 & 36.7 & 72.0 & 83.9 & 59.5 & 29.2 & 36.2 \\ 
\midrule
\textbf{GQA} ($n_k^h$=$n_v^h$=4)
& \textbf{51.66} & 54.0 & 38.0 & 56.0 & 37.5 & 72.3 & 82.0 & 61.3 & 28.6 & 35.4 \\ 
\rowcolor{customgray!15}
\emph{w/} Half K Dim.& \textbf{51.92} \tiny{($\uparrow$0.26)} & 53.4 & 39.2 & 56.6 & 35.6 & 72.5 & 84.0 & 62.3 & 28.9 & 34.8 \\ 
\bottomrule
\end{tabular}
\end{table*}

\begin{table*}[!ht]
\centering
\small
\setlength{\tabcolsep}{5pt}
\caption{Comparisons between the baseline model architectures and those incorporating augmented Q. $d_q^h$ refers to the intermediate Q head dimension. The number of Q heads is 32 for all models in the table ($n_q^h=32$). For the baseline without AugQ, the intermediate dimension of Q head is $d_q^h=2048$. }
\vspace{0.3em}
\label{tab:full_large_q}
\begin{tabular}{l l ccccccccc}
\toprule
\multirow{2}{*}{\textbf{Model}} & \multirow{2}{*}{\textbf{Overall}}& \multicolumn{6}{c}{\textbf{Commonsense \& Comprehension}} & \multicolumn{2}{c}{\textbf{Continued}}& \textbf{LM} \\ \cmidrule(lr){3-8}    \cmidrule(lr){9-10}   \cmidrule(lr){11-11}    
&& {Hella.} & {ObQA} & {Wino.} & {ARC.}  & {PIQA}  & {SciQ} & {Bool.} & {Logi.} & {LAMB.} \\
\midrule
\textbf{MHA} 
& \textbf{52.40}& 55.6 & 37.6 & 57.6 & 36.0 & 73.9 & 85.5 & 59.6 & 28.9 & 36.8 \\ 
\rowcolor{customgray!20}
+ AugQ ($d_q^h$=5632)
& \textbf{53.03} \tiny{($\uparrow$0.63)}& 57.4 & 38.0 & 57.9 & 39.4 & 72.9 & 85.9 & 60.1 & 27.3 & 38.3 \\ 
\midrule
\textbf{GQA} ($n_k^h$=$n_v^h$=16)
& \textbf{52.14}& 55.1 & 39.6 & 56.3 & 35.4 & 71.9 & 85.0 & 61.4 & 27.8 & 36.8 \\ 
\rowcolor{customgray!5}
+ AugQ ($d_q^h$=3072)
& \textbf{53.38} \tiny{(\underline{$\uparrow$1.24})}& 56.6 & 40.2 & 56.1 & 40.0 & 73.2 & 87.3 & 61.0 & 28.3 & 37.7 \\ 
\rowcolor{customgray!10}
+ AugQ ($d_q^h$=4096)
& \textbf{52.93} \tiny{($\uparrow$0.79)}& 56.7 & 40.8 & 56.9 & 37.1 & 73.6 & 83.5 & 61.7 & 28.0 & 38.1 \\ 
\rowcolor{customgray!20}
+ AugQ ($d_q^h$=5632)
& \textbf{53.07} \tiny{($\uparrow$0.93)}& 57.3 & 39.8 & 57.3 & 36.4 & 74.2 & 83.6 & 61.4 & 28.7 & 39.0 \\ 
\midrule
\textbf{GQA} ($n_k^h$=$n_v^h$=4)
& \textbf{51.66}& 54.0 & 38.0 & 56.0 & 37.5 & 72.3 & 82.0 & 61.3 & 28.6 & 35.4 \\ 
\rowcolor{customgray!20}
+ AugQ ($d_q^h$=5632)
& \textbf{53.13} \tiny{($\uparrow$1.47)}& 56.5 & 40.8 & 58.2 & 37.6 & 73.6 & 84.7 & 61.2 & 27.5 & 37.9 \\ 
\bottomrule
\end{tabular}
\vspace{-.8cm}
\end{table*}

Specifically, we observe that \textbf{using only a small proportion of V vectors, specifically those corresponding to the highest attention scores, in the attention output calculation can still preserve the model performance.}
As in previous work, topk V can improve the memory efficiency of the model~\citep{gupta2021memory}, inspiring us to adopt this strategy in this KV imbalance scenario.
As shown in \cref{tab:topkV}, even when we only use 100 V vectors corresponding to the top100 highest attention scores (denoted as ``Sel.V-top100''), the model can still maintain a comparable effect to the original setting. 
These results indicate that we only need to retrieve a small proportion of V vectors from the cache during inference to achieve competitive performance. It significantly reduces data transfer, and thereby improves inference speed.

\begin{table*}[!ht]
\centering
\small
\setlength{\tabcolsep}{5pt}
\caption{Comparisons of the model performance when incorporating the augmented Q component (AugQ) with different sizes and enlarging the FFN module (AugF). The baseline method is GQA, with the FFN dimension being 5632 and $n_k^h$=$n_v^h$=16. $\Delta d_F$ denotes the enlarged dimension for the FFN module, while $d_q^h$ represents the intermediate Q head dimension ($\delta$=$3072$). }
\vspace{0.3em}
\label{tab:full_large_q_vs_mlp}
\begin{tabular}{l l ccccccccc}
\toprule
\multirow{2}{*}{\textbf{Model}} & \multirow{2}{*}{\textbf{Overall}}& \multicolumn{6}{c}{\textbf{Commonsense \& Comprehension}} & \multicolumn{2}{c}{\textbf{Continued}}& \textbf{LM} \\ \cmidrule(lr){3-8}    \cmidrule(lr){9-10}   \cmidrule(lr){11-11}    
\scriptsize{ } &\scriptsize{ }& \scriptsize{{Hella.}} & \scriptsize{ObQA} & \scriptsize{Wino.} & \scriptsize{ARC.}  & \scriptsize{PIQA}  & \scriptsize{SciQ} & \scriptsize{Bool.} & \scriptsize{Logi.} & \scriptsize{LAMB.} \\
\midrule
\textbf{GQA}
& \textbf{52.14}& 55.1 & 39.6 & 56.3 & 35.4 & 71.9 & 85.0 & 61.4 & 27.8 & 36.8 \\ 
\rowcolor{customgray!8}
+ AugF ($\Delta d_{\text{F}}$=$\delta$)
& \textbf{53.26} \tiny{($\uparrow$1.12)}& 57.6 & 39.6 & 57.3 & 38.5 & 73.2 & 87.3 & 59.0 & 27.6 & 39.2 \\ 
\rowcolor{customgray!20}
+ AugQ ($d_q^h$=$\delta$)
& \textbf{53.38} \tiny{(\underline{$\uparrow$1.24})}& 56.6 & 40.2 & 56.1 & 40.0 & 73.2 & 87.3 & 61.0 & 28.3 & 37.7 \\ 
\midrule
\rowcolor{customgray!8}
+ AugF ($\Delta d_{\text{F}}$=$2\delta$)
& \textbf{53.16} \tiny{($\uparrow$1.02)}& 59.3 & 40.0 & 57.0 & 38.3 & 73.9 & 85.2 & 59.8 & 24.9 & 40.1 \\ 
\rowcolor{customgray!20}
+ AugF ($\Delta d_{\text{F}}$=$\delta$) \& AugQ ($d_q^h$=$\delta$)
& \textbf{54.55} \tiny{(\underline{$\uparrow$2.41})}& 58.8 & 41.8 & 57.5 & 39.7 & 74.4 & 86.9 & 62.4 & 28.1 & 41.5 \\ 
\midrule
\rowcolor{customgray!8}
+ AugF ($\Delta d_{\text{F}}$=$3\delta$)
& \textbf{54.50} \tiny{($\uparrow$2.36)}& 60.5 & 42.4 & 59.8 & 39.8 & 74.7 & 87.3 & 59.9 & 27.0 & 39.2 \\ 
\rowcolor{customgray!20}
+ AugF ($\Delta d_{\text{F}}$=$2\delta$) \& AugQ ($d_q^h$=$\delta$)
& \textbf{54.67} \tiny{(\underline{$\uparrow$2.53})}& 60.5 & 41.6 & 57.4 & 39.9 & 74.7 & 87.3 & 59.6 & 27.3 & 43.8 \\ 
\midrule
\rowcolor{customgray!8}
+ AugF ($\Delta d_{\text{F}}$=$5\delta$)
& \textbf{55.08} \tiny{($\uparrow$2.94)}& 62.4 & 41.4 & 57.3 & 41.3 & 75.3 & 88.3 & 60.6 & 26.6 & 42.5 \\ 
\rowcolor{customgray!20}
+ AugF {($\Delta d_{\text{F}}$=$3\delta$)} \& AugQ {($d_q^h$=$\delta$)}  & \textbf{55.09} \tiny{(\underline{$\uparrow$2.95})}& 61.6 & 39.8 & 61.0 & 40.5 & 75.1 & 88.7 & 59.6 & 28.3 & 41.2 \\ 
\bottomrule
\end{tabular}
\end{table*}
\begin{table*}[!ht]
\centering
\small
\setlength{\tabcolsep}{5pt}
\caption{Combinations of three strategies for optimizing the self-attention architecture: augmented Q, compressing the number of K heads, and compressing K head dimension. \ding{51} and \ding{55} represent whether the corresponding strategy is used or not respectively. If the strategy is not used, the standard model setting is adopted (i.e. $n_k^h$=$n_v^h$=16, and Q is not augmented).}
\vspace{0.3em}
\label{tab:combine_exp}
\begin{tabular}{cc c c ccccccccc}
\toprule
\multicolumn{3}{c}{\textbf{Model Config}($n_k^h$=$n_v^h$=16)} & \multirow{3}{*}{\textbf{Overall}} & \multicolumn{6}{c}{\textbf{Commonsense \& Comprehension}} & \multicolumn{2}{c}{\textbf{Continued}}& \textbf{LM} \\\cmidrule(lr){1-3} \cmidrule(lr){5-10}    \cmidrule(lr){11-12}   \cmidrule(lr){13-13}   

\textbf{AugQ}&\textbf{-75\% K Heads }&\textbf{Half K Dim}&& \multirow{2}{*}{Hella.} & \multirow{2}{*}{ObQA} & \multirow{2}{*}{Wino.} & \multirow{2}{*}{ARC.}  & \multirow{2}{*}{PIQA}  & \multirow{2}{*}{SciQ} & \multirow{2}{*}{Bool.} & \multirow{2}{*}{Logi.} & \multirow{2}{*}{LAMB.} \\
($d_q^h$=3072) & ($n_k^h$=4) & ($d^h_k$=$d^h_v/2$) &&&&&&&&&& \\
\midrule




\ding{55}  & \ding{55}  & \ding{55} & \textbf{52.14} & 55.1 & 39.6 & 56.3 & 35.4 & 71.9 & 85.0 & 61.4 & 27.8 & 36.8 \\ 
\midrule
\ding{51} & \ding{51} & \ding{55} & \textbf{52.97} & 56.0 & 38.4 & 57.7 & 37.0 & 72.5 & 83.8 & 62.5 & 30.1 & 38.8 \\ 
\midrule
\ding{51} & \ding{55} & \ding{51} & \textbf{52.72} & 55.9 & 39.4 & 59.1 & 37.5 & 73.1 & 84.3 & 60.6 & 27.0 & 37.4 \\ 
\midrule
\ding{55} & \ding{51} & \ding{51} & \textbf{51.74} & 54.3 & 37.4 & 57.5 & 36.3 & 72.9 & 85.5 & 60.5 & 26.3 & 35.1 \\ 
\midrule
\ding{51} & \ding{51} & \ding{51} & \textbf{52.61} & 55.8 & 40.2 & 54.9 & 38.5 & 74.0 & 84.6 & 61.1 & 26.9 & 37.6 \\ 

\bottomrule
\end{tabular}
\end{table*}
\subsection{Detailed Evaluation Results}
\label{sec:observation_detailed}
Due to the space limitation of the main text, in Table \ref{tab:kv_head_compression}-\ref{tab:large_q_vs_mlp}, we only include the average results across nine benchmarks.
Here, we report the detailed evaluation results on each benchmark in \cref{tab:full_kv_head_compression,tab:full_k_headdim_compression,tab:full_large_q,tab:full_large_q_vs_mlp}.

\subsection{Combination of Multiple Optimization Strategies}
\begin{wraptable}{R}{.5\linewidth}
\centering
\small
\setlength{\tabcolsep}{1pt}
\vspace{-.6cm}
\caption{Key configurations and hyperparameters of \textsc{Sigma}.}
\label{tab:hyperparameters}
\begin{tabular}{l|cc}
\toprule
\bf Parameter $\backslash$ Scale & \bf 1.5B & \bf 10B \\
\midrule
Layers & 26 & 32  \\
Hidden Dimension & 2,048 & 4,096  \\
FFN Dimension & 6,144 & 14,336  \\
Aug Q Dimension & 3,072 & 6,144 \\
Attention Heads & 32 & 32  \\
Key Heads & 4 & 4  \\
Value Heads & 16 & 16 \\
Peak Learning Rate & 4.0e-4 & 1.5e-4  \\
Activation Function & SwiGLU & SwiGLU \\
Vocabulary Size & 128,256 & 128,256 \\
Positional Embeddings & ROPE ($\theta=$50,000) & ROPE ($\theta=$500,000) \\
\bottomrule
\end{tabular}
\vspace{-.8cm}
\end{wraptable}
As shown in \cref{tab:combine_exp}, by integrating three optimization strategies (\textit{i.e.}, -75\% K Heads ($n_k^h$=4), Half K Dim($d^h_k$=
$d^h_v/2$), and AugQ ($d_q^h$=3072), we observe that the model's performance surpasses that of the original model and also exhibits advantages in terms of inference speed and KV cache utilization.

\subsection{\textsc{Sigma} Model Architecture}
\label{sec:appendix_architecture}

Building on the DiffQKV attention, we construct a pre-trained language model, named \textsc{Sigma}.  
Specifically, we adopt two model scales with 1.5B parameters and 10B parameters, respectively (i.e. \textsc{Sigma}-1.5B and \textsc{Sigma}-10B). 
For the sake of balancing the model performance and the cost of the KV cache, during the training of \textsc{Sigma}-1.5B and \textsc{Sigma}-10B, no dimension compression is applied to the K heads. Only the number of K heads was decreased. Concretely, the K head is set to 4, and the number of V heads, which is half of the number of Q heads, is set to 16.
For \textsc{Sigma}-1.5B, we set $d_q^h=3072$, and for \textsc{Sigma}-10B, we set $d_q^h=6144$, corresponding to 1.5 times the dimension of the hidden state, so as to extend the representational space of Q.
We utilize the same vocabulary as Llama3 \citep{dubey2024llama}, with a vocabulary size of 128k.
For more detailed configurations of the \textsc{Sigma} architecture, please refer to \cref{tab:hyperparameters}.

\section{Evaluation Procedure on \textsc{AIMicius} Benchmark} \label{sec:detail_system_benchmark}

\subsection{CMDGen}



We consider the following evaluation metrics on the CMDGen task:
\begin{itemize}
    \item \textbf{CMD Score} computes the cosine similarity between the embeddings of the generated command and the ground-truth command. The embeddings are encoded with all-MiniLM-L6-v2.\footnote{\url{https://huggingface.co/sentence-transformers/all-MiniLM-L6-v2}}
    \item \textbf{Output Score} evaluates the cosine similarity between the execution results of the generated command and that of the ground-truth command.  
    The embeddings are also obtained with all-MiniLM-L6-v2.
    \item \textbf{Calibration Score} serves as an approximate measure of accuracy. For a given test sample, the calibration score is assigned as one if either the CMD Score or the Output Score exceeds a predefined threshold. 
    \item \textbf{Exact Match} checks if the output command exactly matches the ground-truth command. 
    \item \textbf{Success Ratio} measures if the execution result of the generated command is similar enough to that of the ground-truth command. 
    \item \textbf{Accuracy} serves as a comprehensive measure of the model's overall performance. For a given test sample, a generated command is deemed accurate if it either exactly matches the ground-truth command (Exact Match = 1) or produces highly similar execution results to the ground-truth (Success Ratio = 1). 
    We consider it as \textbf{the primary metric} on the CMDGen task.
\end{itemize}

\subsection{Infrawise}


DCW generation aims to produce a JSON object, referred to as a DCW, based on the user instructions. Specifically, DCW stands for: 
\begin{itemize}
    \item \textbf{Design}, a kind of virtual machine (probably with a specific feature) or a GPU type.  It defines the testbed to be evaluated and should be inferred based on the user’s intentions. The design has two sub-keys: baseline and target, implying the baseline design and target design user wants to retrieve. Commonly used values include \texttt{NDv4}, \texttt{NDmv4}, \texttt{NDv4\_MI200}, \texttt{NDv5}, \texttt{NCv3}, \texttt{A100}, \texttt{H100}, \texttt{MI200} and \texttt{MI300x}.
    \item \textbf{Workload}: the benchmarks, models, algorithms, or performance measurement tools that the user intends to run on the design testbed.
    \item \textbf{Criterion}: the evaluation principles or performance metrics the user intends to apply. It specifies how the effectiveness, quality, or performance between the two designs should be compared. Common criteria include evaluating both testbeds’ peak performance, assessing cost-effectiveness, or measuring results within a specified time duration.
\end{itemize}

Benchmark result retrieval entails two substeps for enhanced performance. In the first phase, a set of benchmark results (typically 10 samples) is retrieved based on initial filtering criteria. In the second phase, LLMs are instructed to select all the appropriate benchmark results from this set, guided by the generated DCW. 
If none of the retrieved results align with the DCW specifications, the final answer would be marked as ``None.''

We adopt the following metrics to conduct evaluation on Infrawise:
\begin{itemize}
    \item \textbf{Target, Baseline, Criterion, Workload}: These four metrics evaluate individual components of the generated results. It equals one if the generated result (\textit{e.g.}, Target Design, Baseline Design, Criterion, Workload) exactly matches the ground truth.
    \item \textbf{DCW}: This composite metric evaluates the correctness of the entire DCW. It equals one if all four components: Target, Baseline, Criterion, and Workload, are correct (\textit{i.e.}, each equals one individually).
    \item \textbf{Benchmark Result Recall}: This metric measures the completeness of the retrieval phase. It equals one if all suitable benchmark results are retrieved without omission.
    \item \textbf{Benchmark Result Accuracy}: The most critical metric. For a given test sample, its benchmark result accuracy is assigned as one only if all suitable benchmark results are retrieved and no irrelevant results are included.
\end{itemize}


\subsection{Optiflow}


The evaluation of Optiflow focuses on assessing the quality of the generated Python code using four key metrics, described as follows:
\begin{itemize}
    \item \textbf{Code Detected}: This metric verifies whether the output includes a Python code block. If a code block is present, the metric is set to one.
    \item \textbf{Code Executable}: This metric evaluates the validity of the Python code. If the generated code can be executed without errors, the metric equals one.
    \item \textbf{Plan Valid}: This metric assesses whether the generated Python code provides a valid plan based on the specified SKU. The metric equals one if the plan meets the given requirements.
    \item \textbf{Plan Improved}: Each test case involves both Plan Generation and Plan Improvement tasks. The evaluator first measures the latency of the initial output provided by the LLM and then requests an improved plan. This metric equals one if the latency of the second output is reduced compared to the first.
\end{itemize}


\subsection{NL2KQL}

A typical KQL query consists of four key components: \texttt{CLUSTER}, \texttt{DATABASE}, \texttt{TABLE}, and \texttt{COLUMN}, taking the structure of:

\vspace{-3mm}
\begin{verbatim}
    cluster(CLUSTER).database(DATABASE).TABLE
    | where COLUMN
\end{verbatim}
\vspace{-2mm}

The evaluation of NL2KQL focuses on the exact match of each KQL component and the overall syntax accuracy of the generated queries. 
Specifically, the evaluation metrics include:
\begin{itemize}
    \item \textbf{Syntax Accuracy}, which measures whether the generated KQL query adheres to the correct syntax;
    \item \textbf{Similarity}, which assesses the Jaccard similarity between the generated KQL query and the ground-truth query;
    \item \textbf{Cluster Score, Database Score, Table Score, Column Score}, used to assess the correctness of individual components (\texttt{CLUSTER}, \texttt{DATABASE}, \texttt{TABLE}, and \texttt{COLUMN}) by checking for the recall rate with the corresponding elements in the ground-truth query.
\end{itemize}


\clearpage
\section{Data Examples of \textsc{AIMicious} Benchmark}
\label{sec:data examples}
\subsection{CMDGen Examples} \label{sec:cmdgen_ex}
\vspace{-.4cm}
\begin{figure}[!h]
    \centering
    \includegraphics[width=0.9\linewidth]{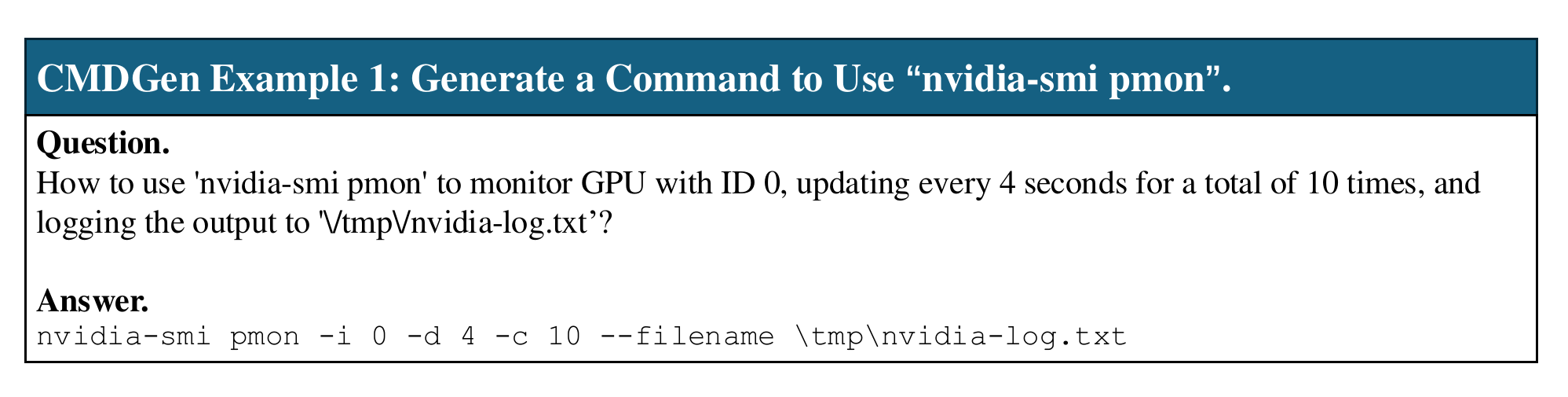}
    \label{fig:cmdgen_ex1}
\end{figure}

\vspace{-.4cm}

\begin{figure}[h]
    \centering
    \includegraphics[width=0.9\linewidth]{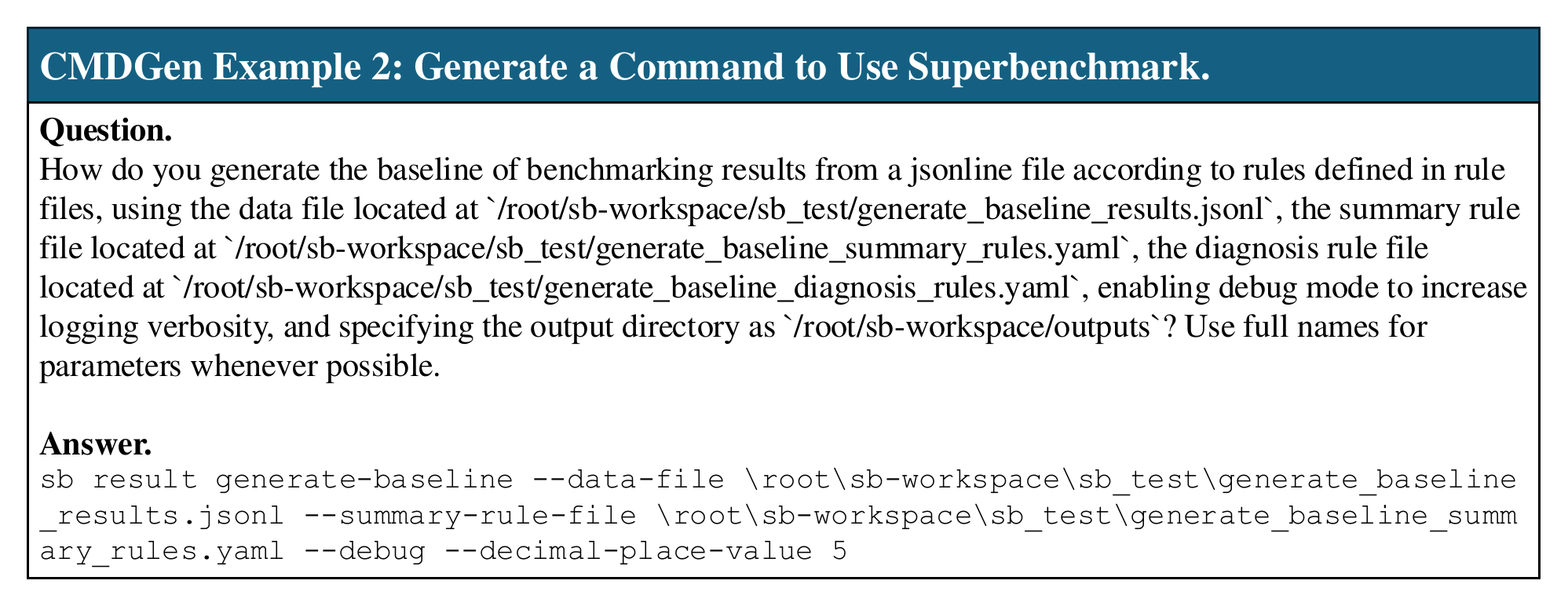}
    \label{fig:cmdgen_ex2}
\end{figure}

\vspace{-.4cm}

\subsection{NL2KQL Examples} \label{sec:nl2kql_ex}
\begin{figure}[h]
    \centering
    \includegraphics[width=0.9\linewidth]{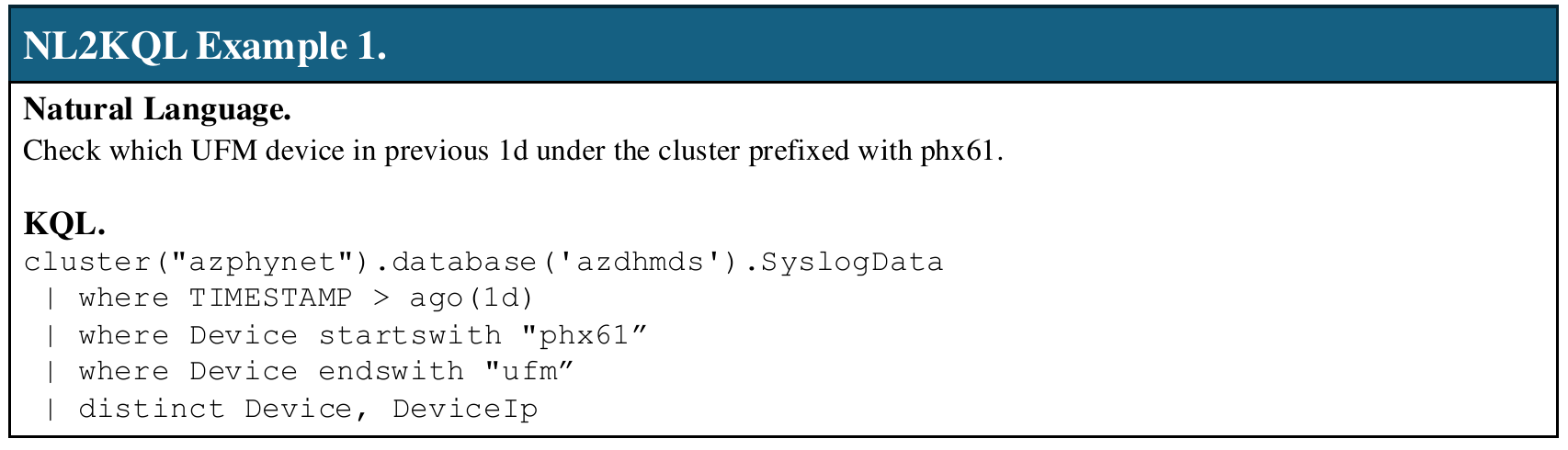}
    \label{fig:nl2kql_ex1}
\end{figure}

\vspace{-.4cm}

\subsection{Infrawise Examples} \label{sec:infrawise_ex}

\begin{figure}[H]
    \centering
    \includegraphics[width=0.9\linewidth]{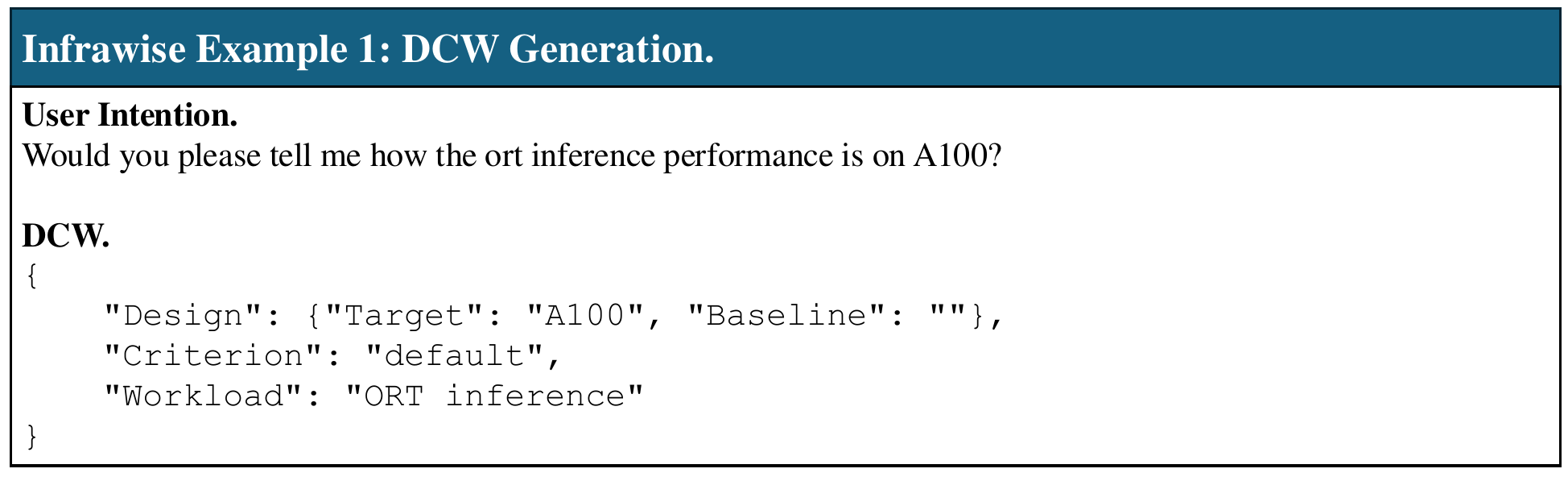}
    \label{fig:infrawise_ex1}
\end{figure}

\clearpage
\begin{figure}[h]
    \centering
    \includegraphics[width=0.9\linewidth]{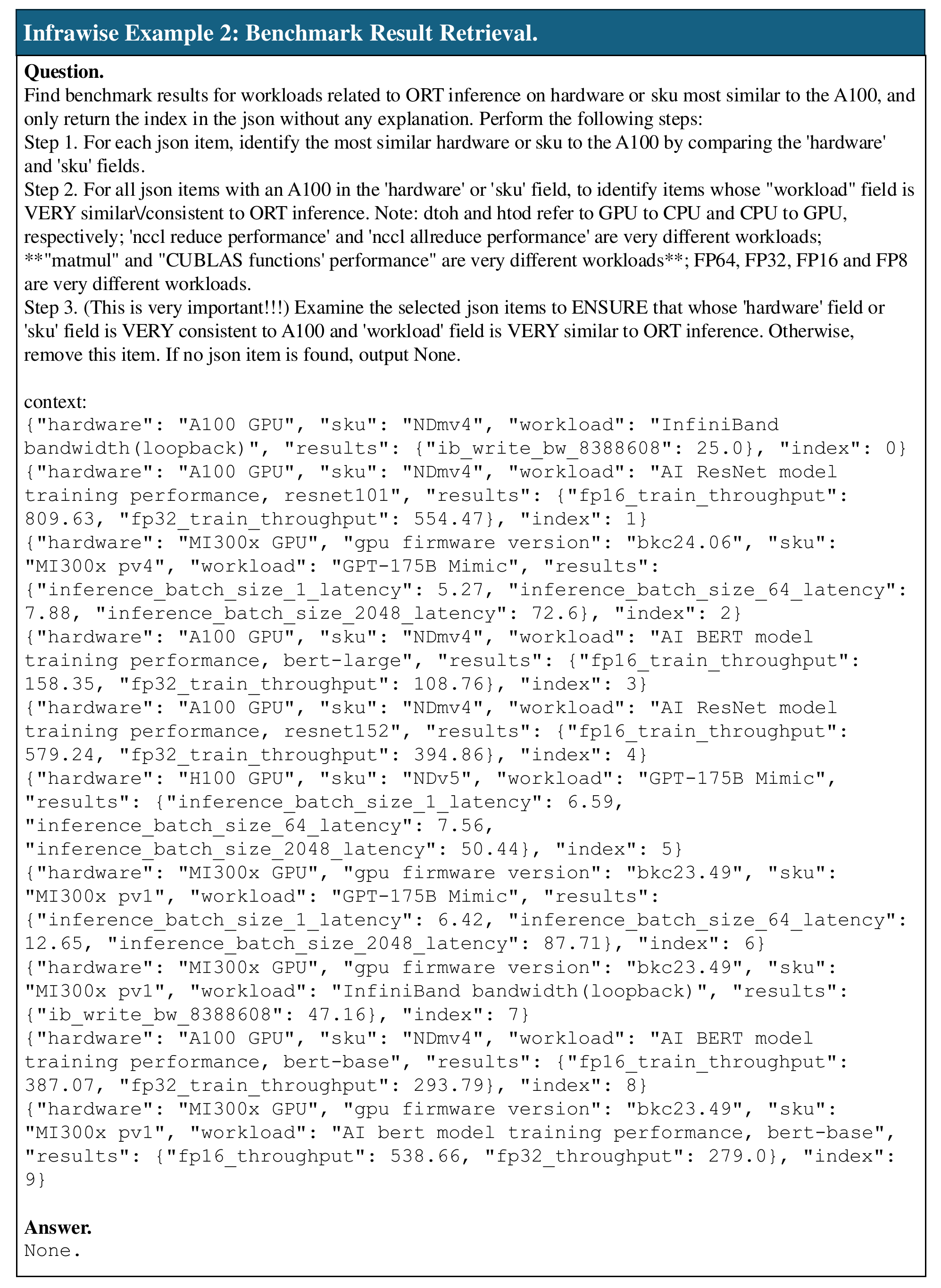}
    \label{fig:infrawise_ex2}
\end{figure}

\clearpage
\subsection{Optiflow Examples} \label{sec:optiflow_ex}
\vspace{-.4cm}
\begin{figure}[h]
    \centering
    \includegraphics[width=0.88\linewidth]{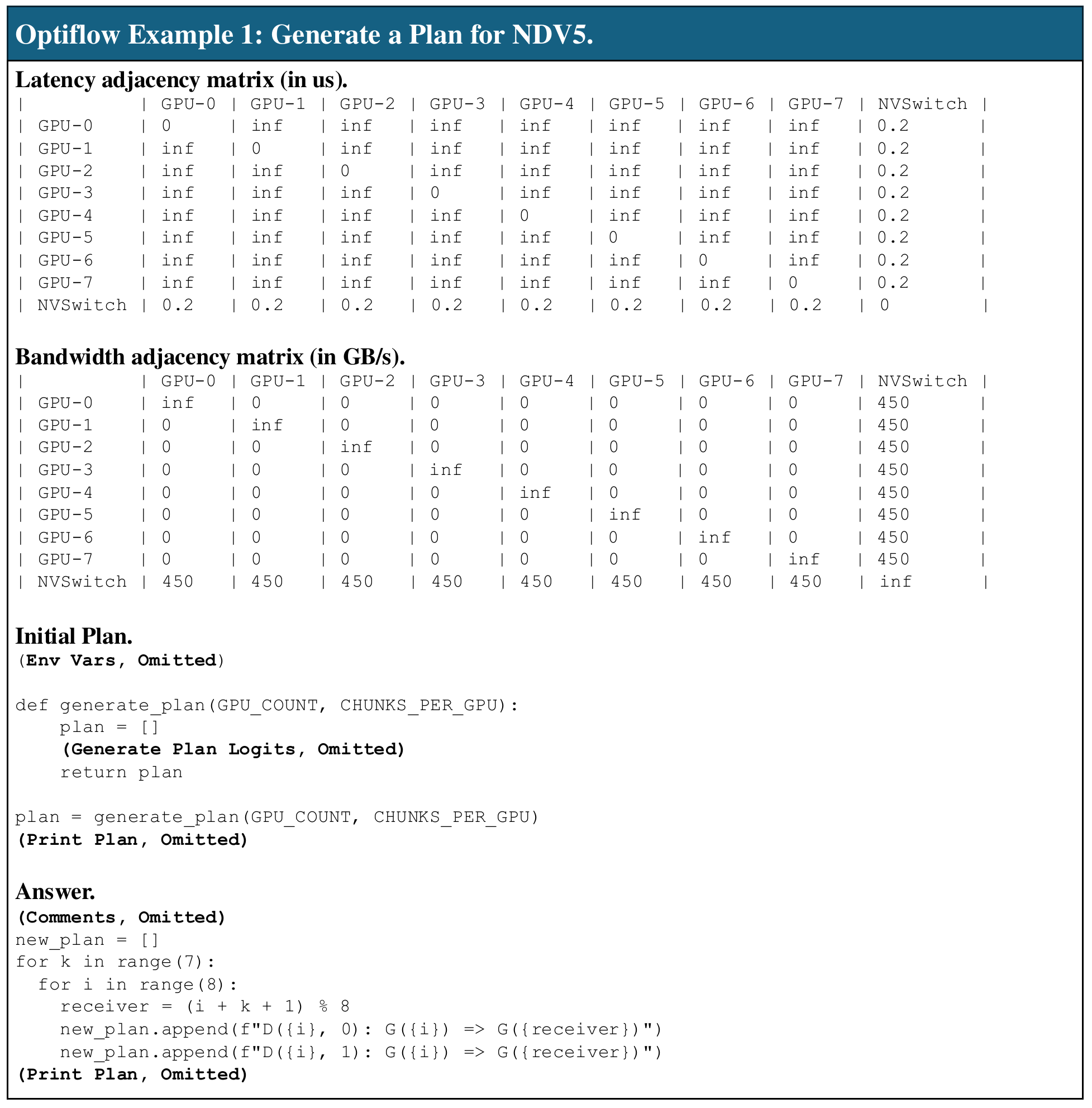}
    \label{fig:optiflow_ex1}
\end{figure}

\vspace{-.7cm}

\begin{figure}[H]
    \centering
    \includegraphics[width=0.88\linewidth]{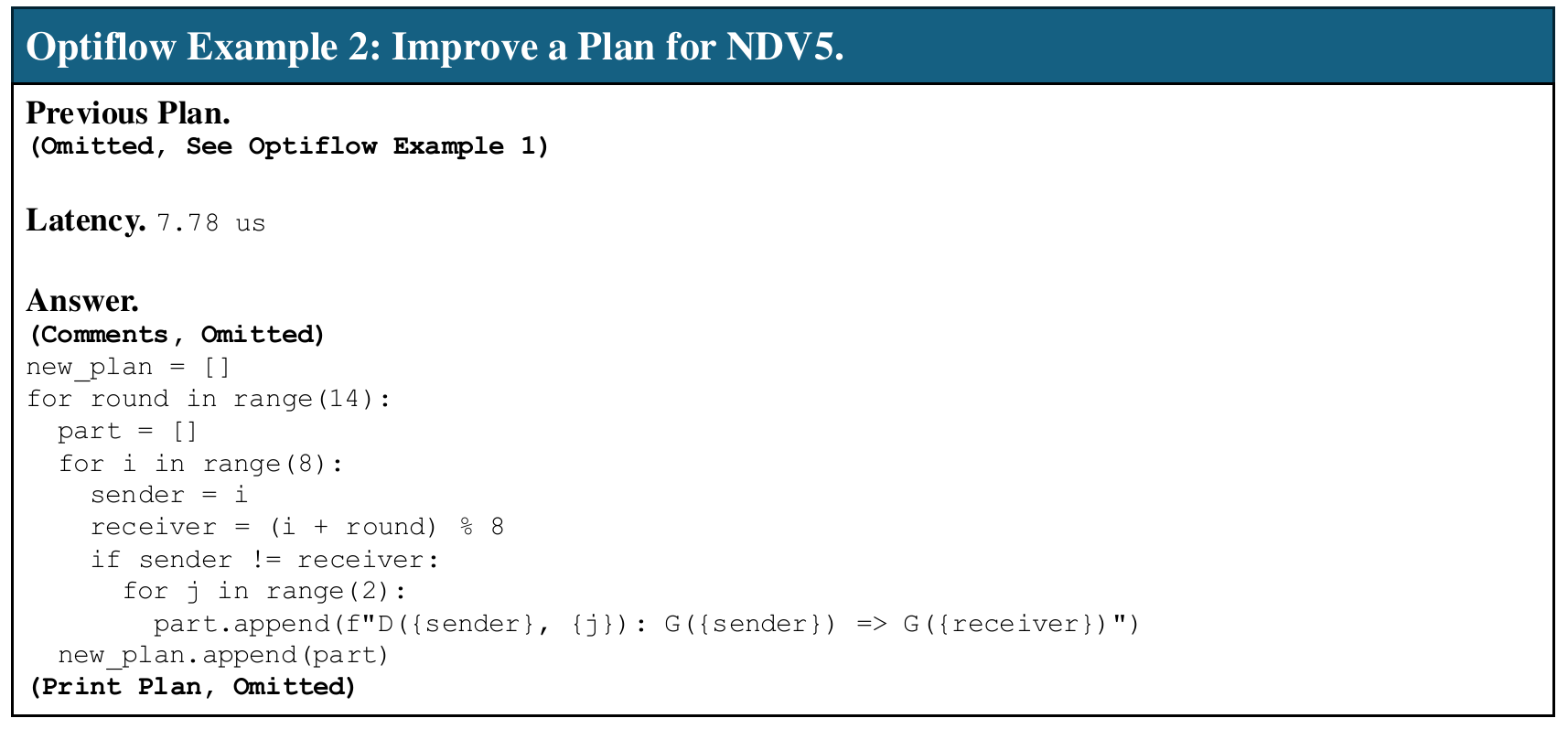}
    \label{fig:optiflow_ex2}
\end{figure}

\section{Official Implementation of KV Group Sharing} \label{sec:kv_sharing}

Here, we release the official implementation of KV Group Sharing to ensure compatibility with mainstream LLM frameworks.


\lstnewenvironment{pythoncode}[1][]{%
  \lstset{
    language=Python,
    numbers=left,
    numberstyle=\small\color{gray},
    stepnumber=1,
    breaklines=true,
    breakatwhitespace=false,
    frame=single,
    tabsize=2,
    basicstyle=\ttfamily\small,
    keywordstyle=\color{blue},
    stringstyle=\color{orange},
    linewidth=1.0\textwidth,
    xleftmargin=3em,
    commentstyle=\color{green!50!black},
    moredelim=[is][\bfseries\color{red}]{@}{@},
    #1 
  }
}{}

\begin{pythoncode}
def kv_group_sharing(self, key_states, value_states):
    """
    This is for balance the number of KV heads to facilitate subsequent flash attention calculations
    """
    batch, _, slen, head_dim = key_states.shape
    
    if self.num_key_heads > self.num_value_heads:
        n_rep = int(self.num_key_heads / self.num_value_heads)
        value_states = value_states[:, :, None, :, :].expand(batch, self.num_value_heads, n_rep, slen, head_dim)
        return key_states, value_states.reshape(batch, self.num_value_heads * n_rep, slen, head_dim)
    else:
        n_rep = int(self.num_value_heads / self.num_key_heads)
        if n_rep == 1:
            return key_states, value_states
        key_states = key_states[:, :, None, :, :].expand(batch, self.num_key_heads, n_rep, slen, head_dim)
        return key_states.reshape(batch, self.num_key_heads * n_rep, slen, head_dim), value_states
\end{pythoncode}

You can use this code in the \texttt{forward} function like this:

\begin{pythoncode}
def forward(
    self,
    hidden_states: torch.Tensor,
    attention_mask: Optional[torch.LongTensor] = None,
    position_ids: Optional[torch.LongTensor] = None,
    past_key_value: Optional[Cache] = None,
    output_attentions: bool = False,
    use_cache: bool = False,
    cache_position: Optional[torch.LongTensor] = None,
) -> Tuple[torch.Tensor, Optional[torch.Tensor], Optional[Tuple[torch.Tensor]]]:
    if isinstance(past_key_value, StaticCache):
        raise ValueError(
            "`static` cache implementation is not compatible with `attn_implementation==flash_attention_2` "
            "make sure to use `sdpa` in the mean time, and open an issue at https://github.com/huggingface/transformers"
        )
    output_attentions = False
    bsz, q_len, _ = hidden_states.size()

    query_states = self.q_proj(hidden_states)
    
    key_states = self.k_proj(hidden_states)
    
    query_states = self.q_down_proj(self.act_fn(self.q_gate_proj(query_states)) * self.q_up_proj(query_states))
    
    value_states = self.v_proj(hidden_states)

    # Flash attention requires the input to have the shape
    # batch_size x seq_length x head_dim x hidden_dim
    # therefore we just need to keep the original shape
    
    query_states = query_states.view(bsz, q_len, self.num_heads, self.head_dim).transpose(1, 2)
    key_states = key_states.view(bsz, q_len, self.num_key_heads, self.head_dim).transpose(1, 2)
    value_states = value_states.view(bsz, q_len, self.num_value_heads, self.head_dim).transpose(1, 2)
    
    cos, sin = self.rotary_emb(value_states, position_ids)
    query_states, key_states = apply_rotary_pos_emb(query_states, key_states, cos, sin)

    if past_key_value is not None:
        # sin and cos are specific to RoPE models; cache_position needed for the static cache
        cache_kwargs = {"sin": sin, "cos": cos, "cache_position": cache_position}
        key_states, value_states = past_key_value.update(key_states, value_states, self.layer_idx, cache_kwargs)
    
    @key_states, value_states = self.kv_group_sharing(key_states, value_states)@
    query_states = query_states.transpose(1, 2)
    key_states = key_states.transpose(1, 2)
    value_states = value_states.transpose(1, 2)

    (Code Omitted.)

    '''
    flash attention
    '''
    attn_output = self._flash_attention_forward(
        query_states, key_states, value_states, attention_mask, q_len, dropout=dropout_rate
    )

    attn_output = attn_output.reshape(bsz, q_len, -1).contiguous()
    attn_output = self.o_proj(attn_output)
        
    
    if not output_attentions:
        attn_weights = None
    
    return attn_output, attn_weights, past_key_value
\end{pythoncode}

\section{Official Implementation of Efficiency Recording} \label{sec:eff_r}
Here, we offers an official implementation to use Cuda Event Elapsed Time to record the cost of attention computation in \texttt{modeling\_sigma.py}.
\begin{pythoncode}
global total_attention_cost
start = torch.cuda.Event(enable_timing=True)
end = torch.cuda.Event(enable_timing=True)
start.record()
attn_output = flash_attn_func(
    query_states, key_states, value_states, dropout, softmax_scale=softmax_scale, causal=causal
)
end.record()
end.synchronize()
latency = start.elapsed_time(end)
total_attention_cost += latency
\end{pythoncode}

To use Kernel Execution Time to record the cost, there is no need to edit the modeling python file, we can directly use a command line tool to implement this: \texttt{nsys profile --output xxx --stats=true -t cuda python xxx.py}.

\clearpage
\section{Detailed Efficiency Results} \label{sec:dt_eff}
\begin{table*}[!h]
\vspace{-10pt}
\centering
\small
\caption{KET Results (ns) with the prefix length increase from 2k to 32k, keeping the output length as 10. \textbf{Split} represents the split kernel and \textbf{Combine} represents the combine kernel.}
\begin{tabular}{c|ccc|ccc|ccc}
\toprule
\bf Prefix & \multicolumn{3}{c|}{\bf \textsc{Std}} & \multicolumn{3}{c|}{\bf \textsc{Sigma} 1.5B} & \multicolumn{3}{c}{\bf Relative Improvement} \\
\cmidrule{2-10}
\bf Length & \bf Split & \bf Combine & \bf Total Cost & \bf \bf Split & \bf Combine & \bf Total Cost & \bf Split & \bf Combine  & \bf Total Cost\\
\midrule
\bf 2k & 2.53E+6 & 1.88E+6 & 4.41E+6 & 2.50E+6 & 1.85E+6 & 4.34E+6 & 1.17\%  & 1.68\%  & 1.39\% \\
\bf 4k & 4.68E+6 & 1.91E+6 & 6.59E+6 & 3.49E+6 & 1.91E+6 & 5.40E+6 & 25.33\%  & 0.08\%  & 18.02\% \\
\bf 16k & 1.52E+7 & 1.94E+6 & 1.72E+7 & 1.12E+7 & 1.94E+6 & 1.31E+7 & 26.30\%  & 0.25\%  & 23.35\% \\
\bf 32k & 2.75E+7 & 1.99E+6 & 2.95E+7 & 2.00E+7 & 2.01E+6 & 2.21E+7 & 27.21\%  & -0.93\% & 25.31\% \\
\bottomrule
\end{tabular}
\label{tab:sigma_eff_k}
\end{table*}

\begin{figure*}[!h]
    \centering
    \begin{subfigure}[b]{0.32\textwidth}
        \includegraphics[width=\textwidth]{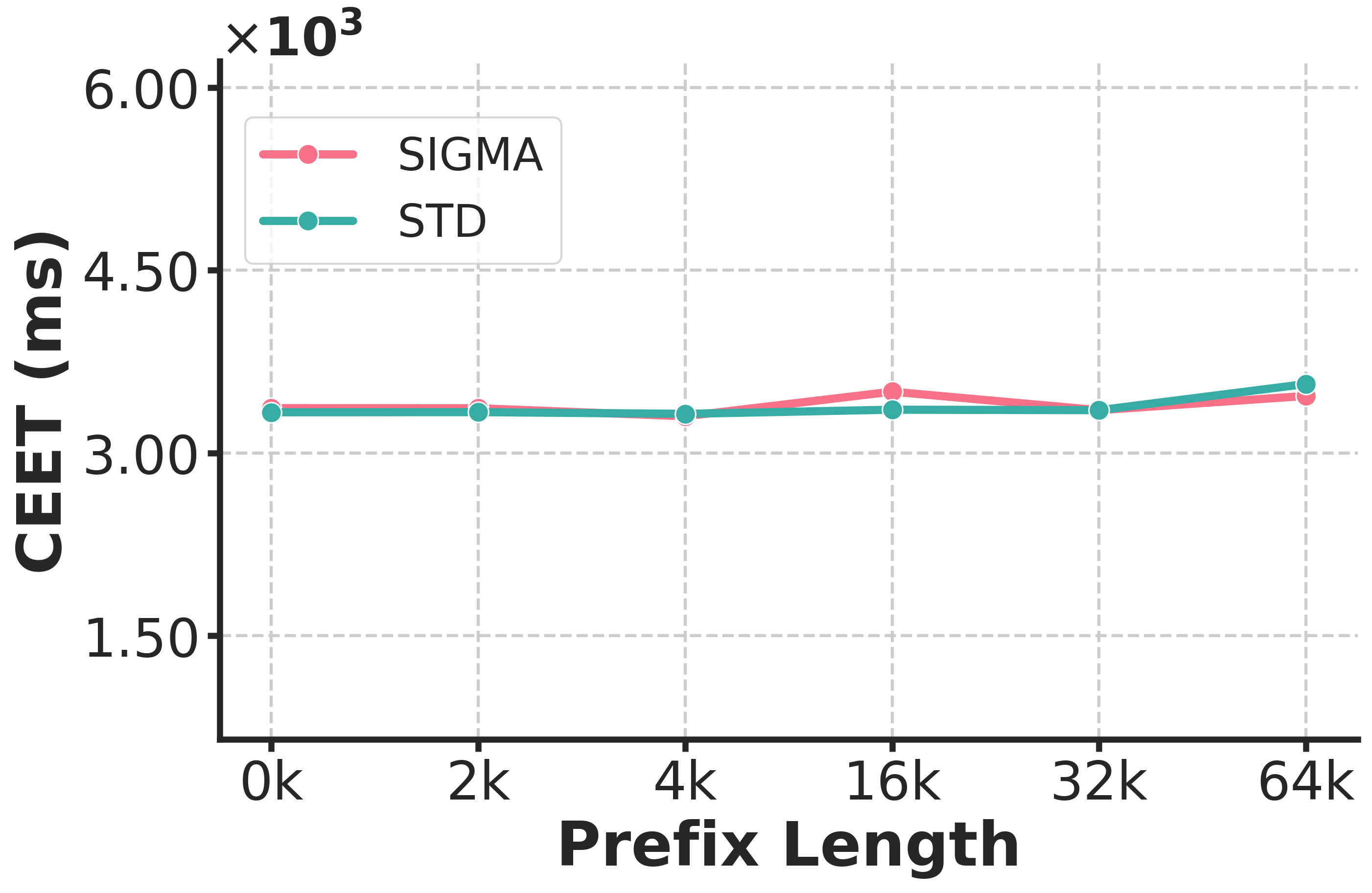}
        \caption{$\text{Output Length }= 2k$.}
        \label{fig:augq_cost_a}
    \end{subfigure}
    \hfill
    \begin{subfigure}[b]{0.32\textwidth}
        \includegraphics[width=\textwidth]{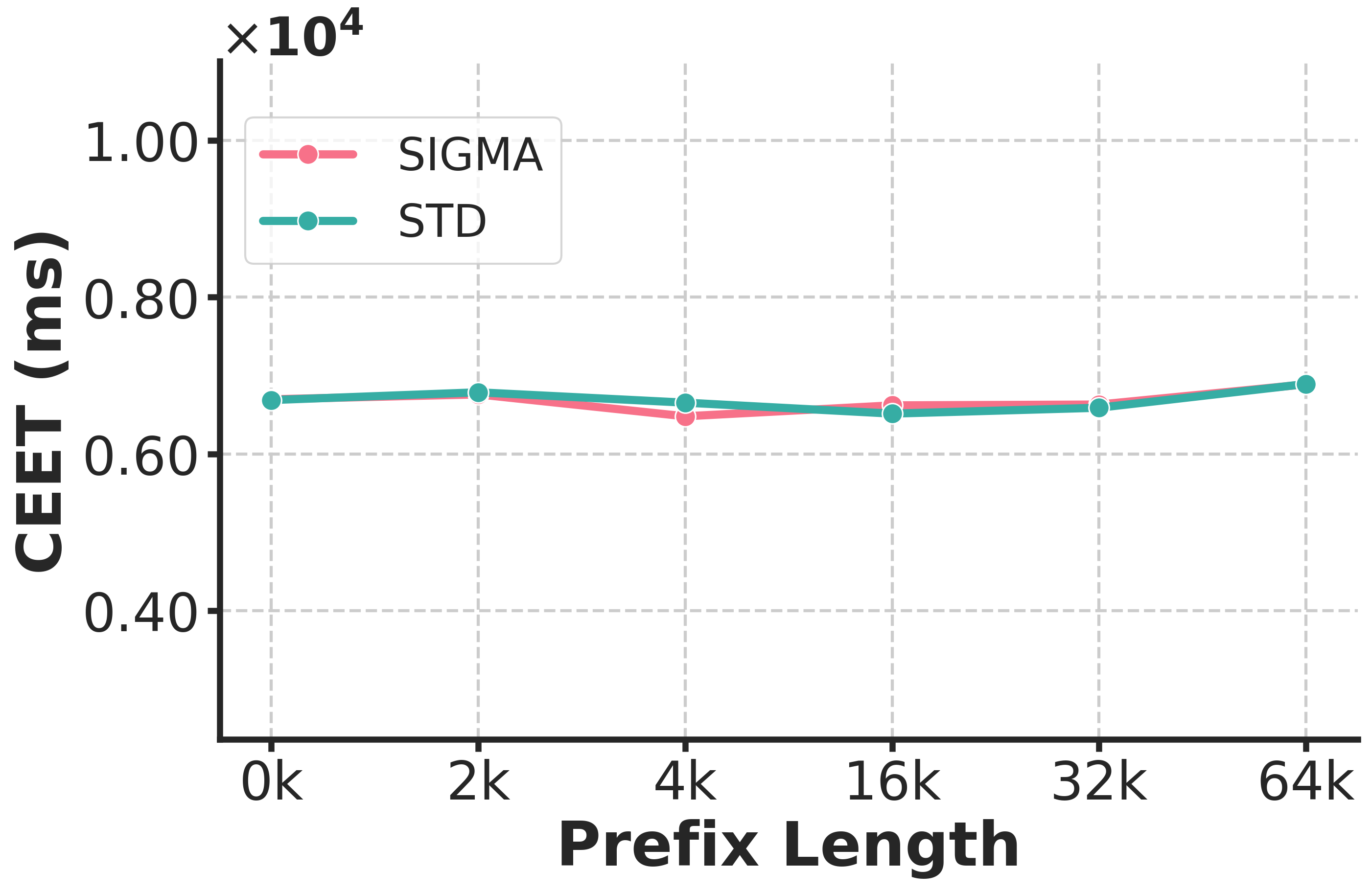}
        \caption{Output Length $= 4k$.}
        \label{fig:augq_cost_b}
    \end{subfigure}
    \hfill
    \begin{subfigure}[b]{0.32\textwidth}
        \includegraphics[width=\textwidth]{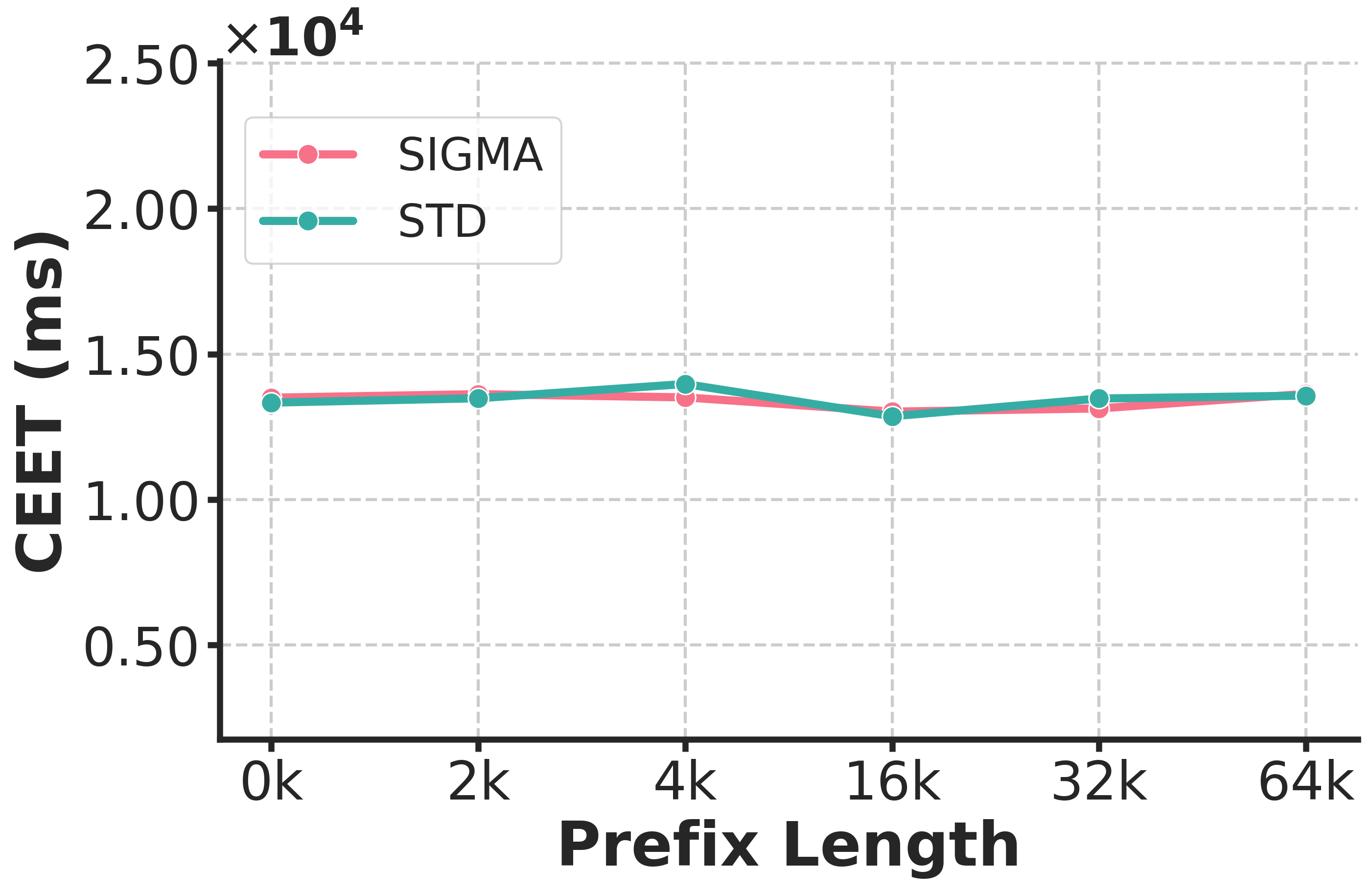}
        \caption{Output Length $= 8k$.}
        \label{fig:augq_cost_c}
    \end{subfigure}

    \centering
    \begin{subfigure}[b]{0.32\textwidth}
        \includegraphics[width=\textwidth]{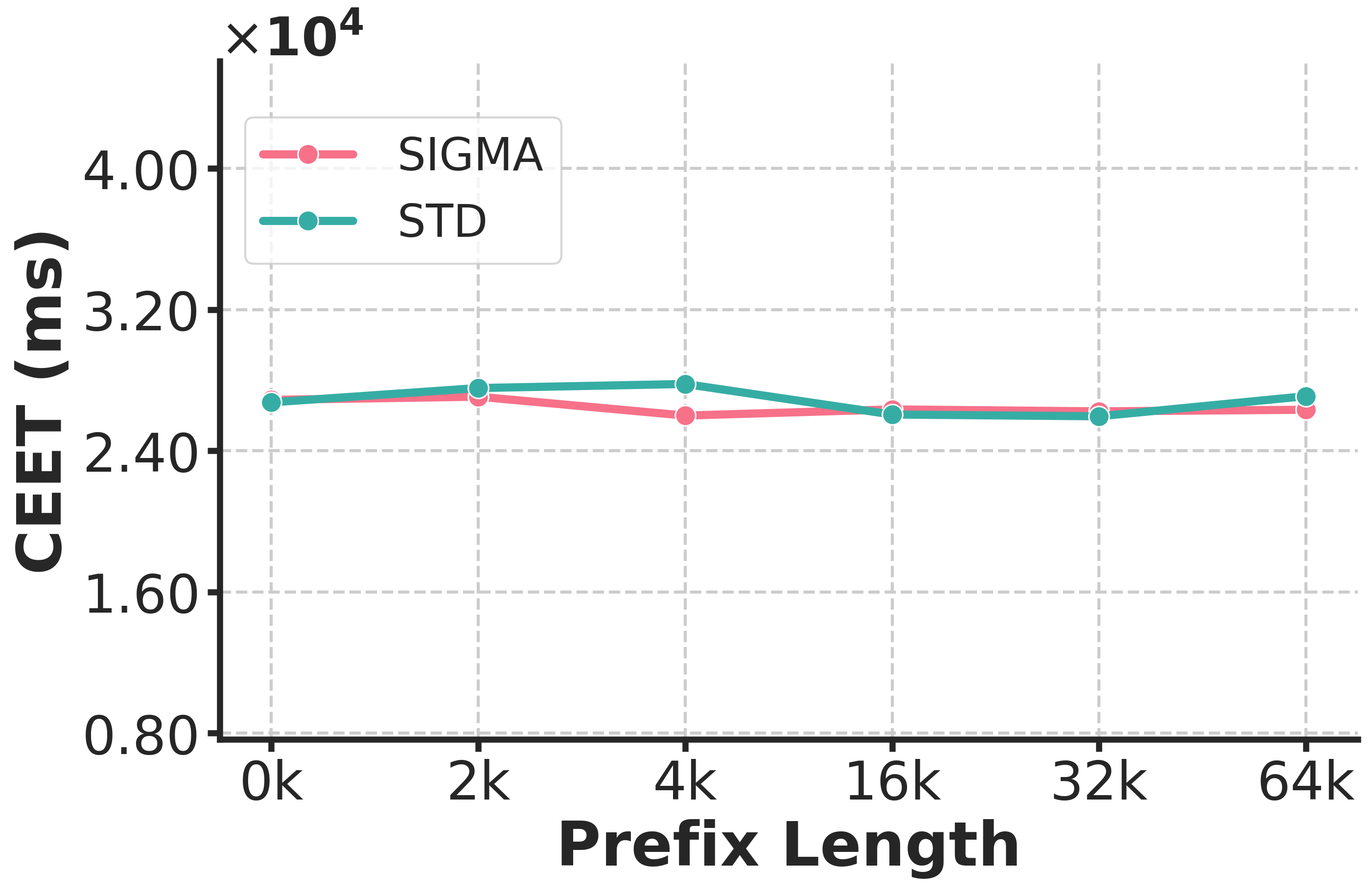}
        \caption{Output Length $= 16k$.}
        \label{fig:augq_cost_d}
    \end{subfigure}
    \hfill
    \begin{subfigure}[b]{0.32\textwidth}
        \includegraphics[width=\textwidth]{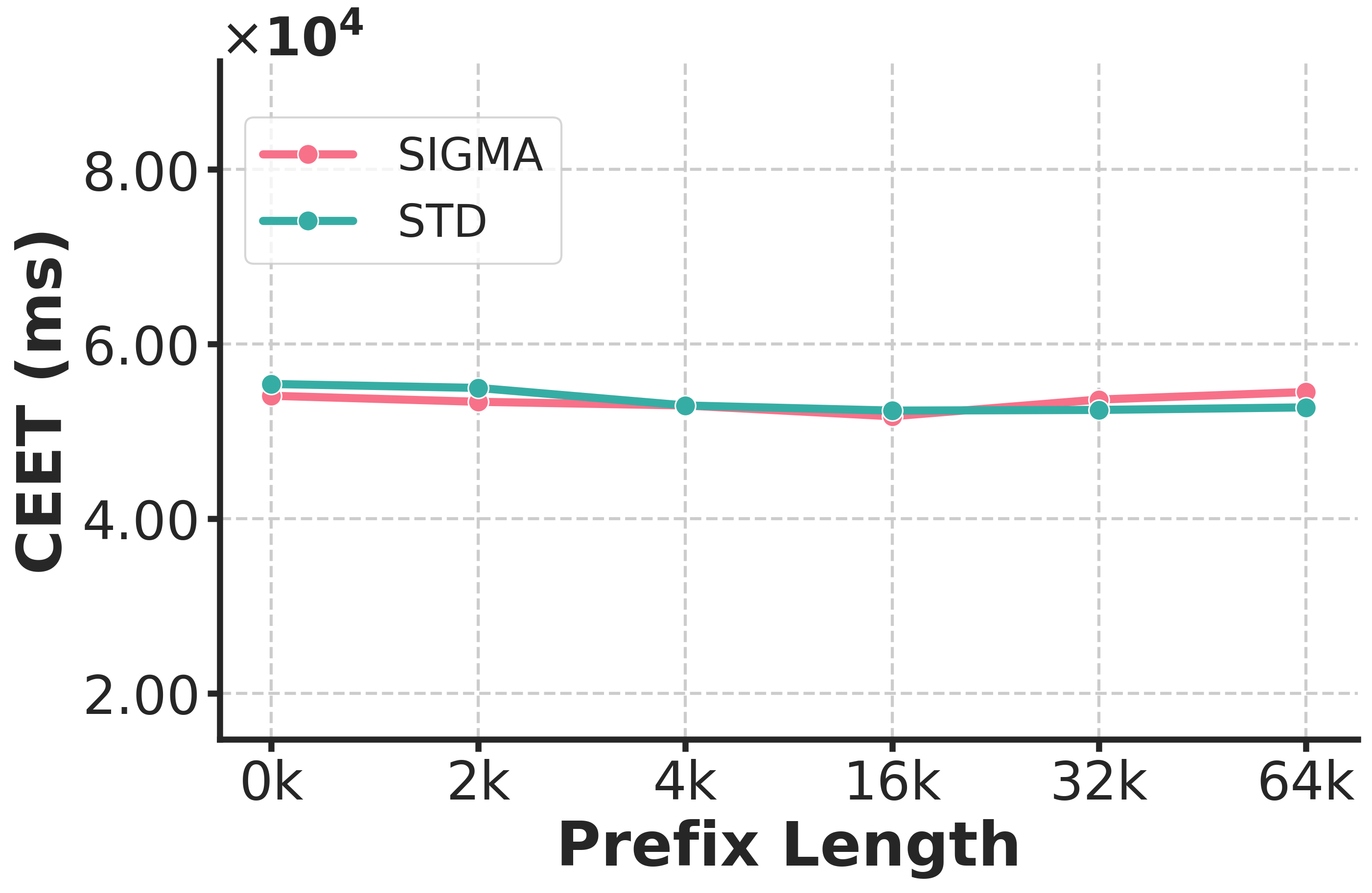}
        \caption{Output Length $= 32k$.}
        \label{fig:augq_cost_e}
    \end{subfigure}
    \hfill
    \begin{subfigure}[b]{0.32\textwidth}
        \includegraphics[width=\textwidth]{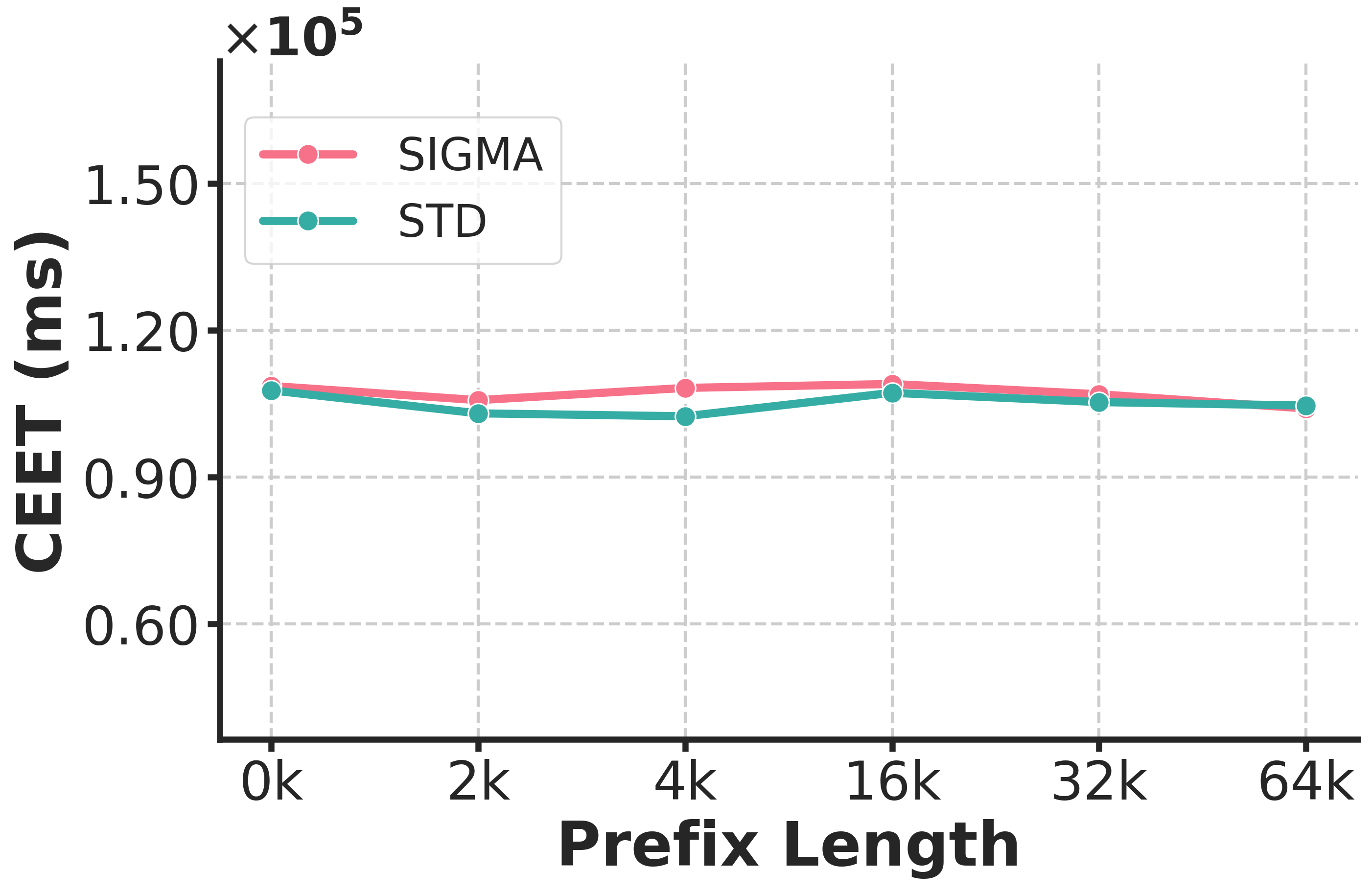}
        \caption{Output Length $= 64k$.}
        \label{fig:augq_cost_f}
    \end{subfigure}

    \centering
    \begin{subfigure}[b]{0.49\textwidth}
        \includegraphics[width=\textwidth]{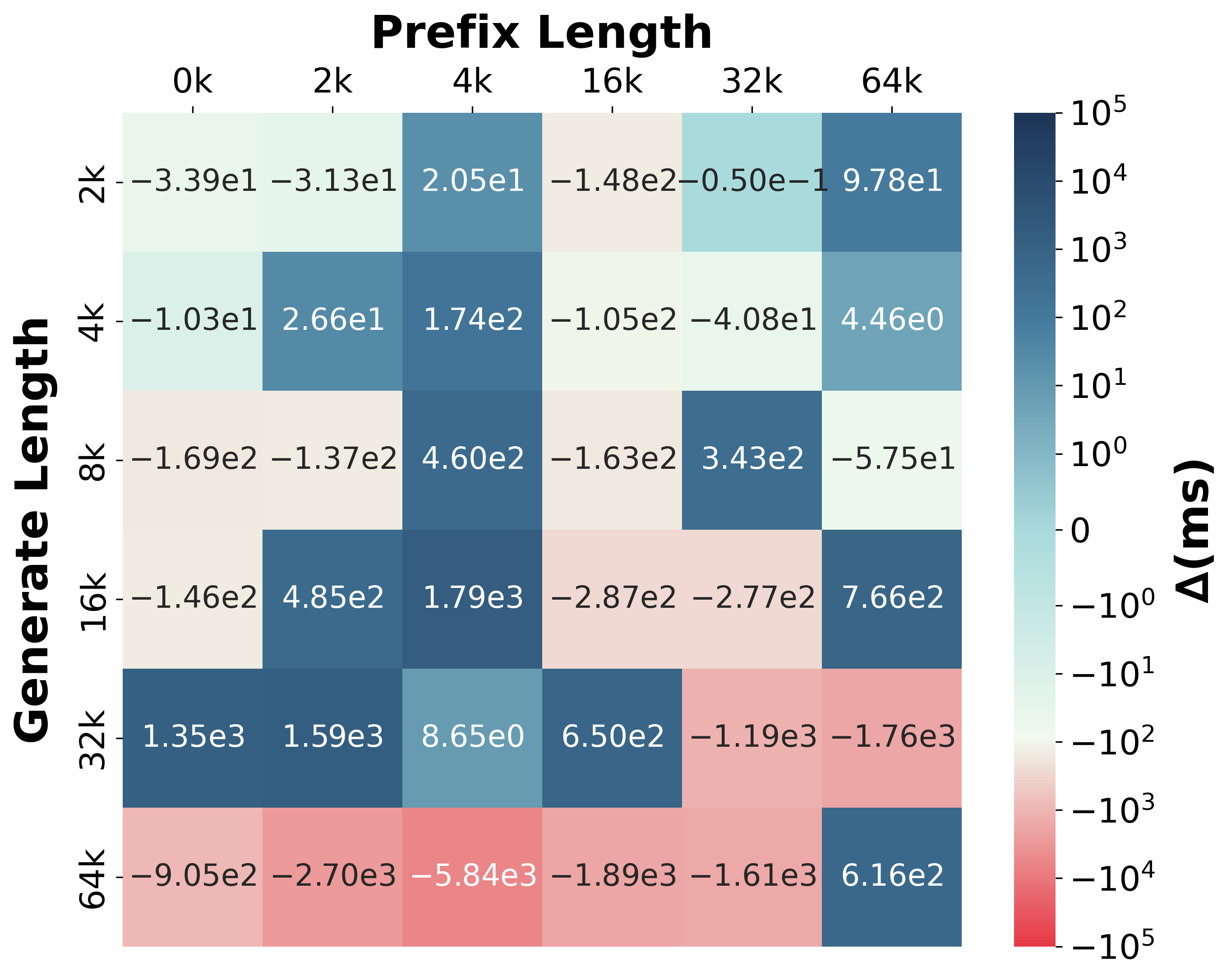}
        \caption{\textsc{Sigma}'s absolute CEET improvment vs \textsc{Std}.}
        \label{fig:augq_cost_g}
    \end{subfigure}
    \begin{subfigure}[b]{0.49\textwidth}
        \includegraphics[width=\textwidth]{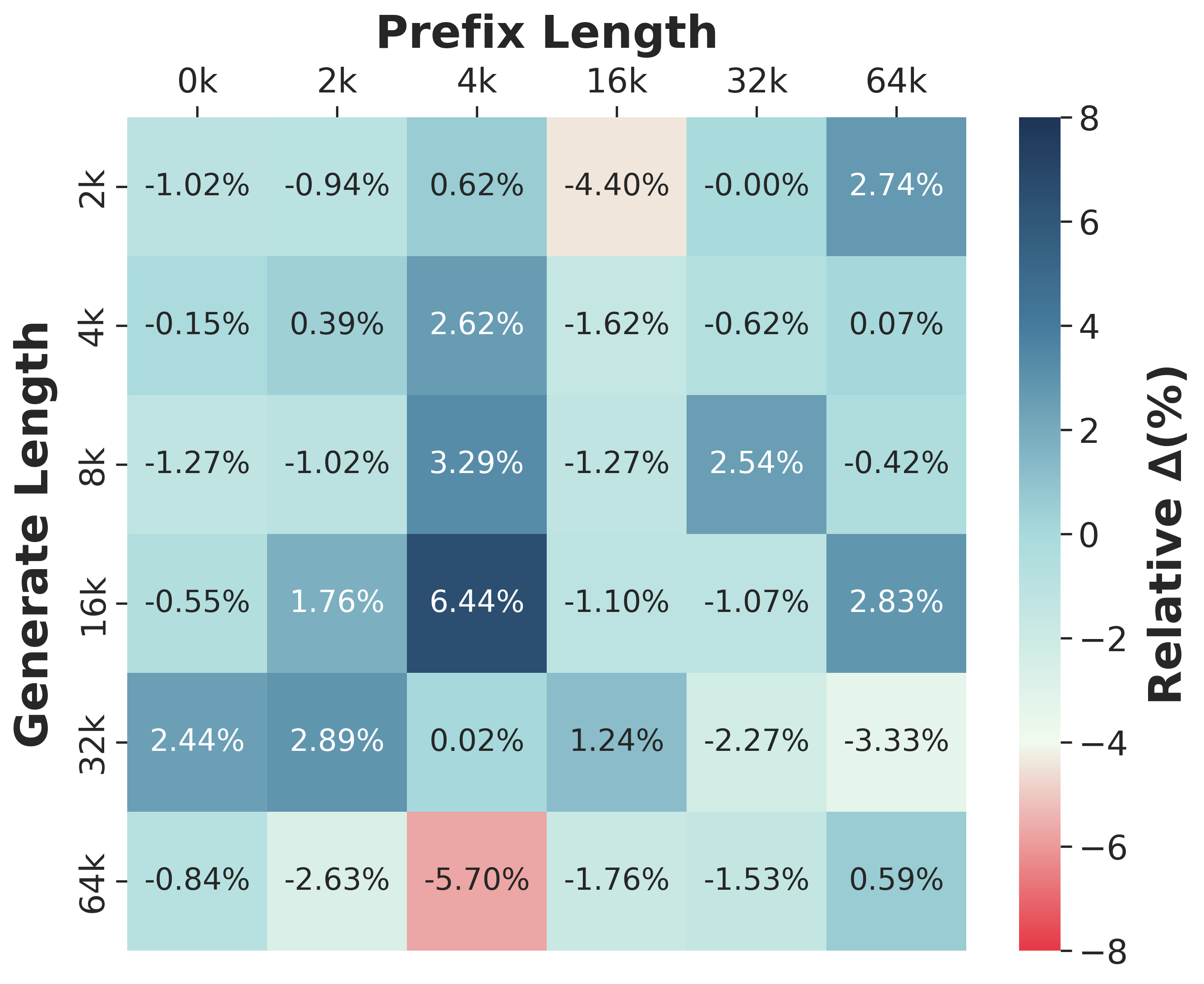}
        \caption{\textsc{Sigma}'s relative CEET improvment vs \textsc{Std}.}
        \label{fig:augq_cost_h}
    \end{subfigure}

    \caption{CEET comparison of augmented Q between Standard model(\textsc{Std}) and \textsc{Sigma}. From (a) to (f), the output length increases progressively from 2k to 64k tokens.}
    \label{fig:augq_cost}

\end{figure*}
Here, we report detailed KET results and CEET results of three modules: KV Cache, Attention Computation, and Augmented Q, as shown in \cref{fig:attn_cost,fig:kvcache_cost,fig:augq_cost,tab:sigma_eff_k}.

\begin{figure*}[!ht]
    \centering
    \begin{subfigure}[b]{0.32\textwidth}
        \includegraphics[width=\textwidth]{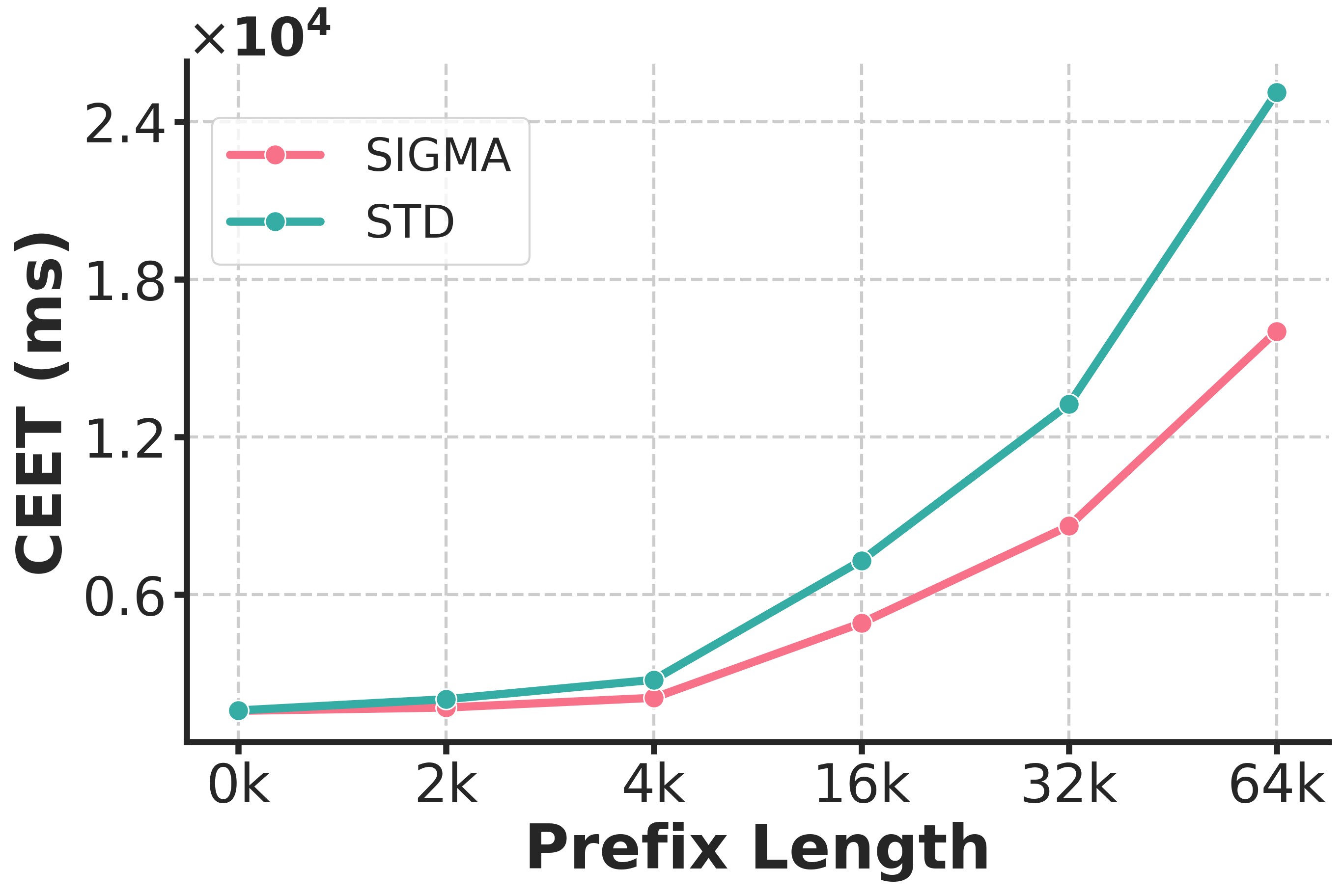}
        \caption{Output Length $= 2k$.}
        \label{fig:kvcache_cost_a}
    \end{subfigure}
    \hfill
    \begin{subfigure}[b]{0.32\textwidth}
        \includegraphics[width=\textwidth]{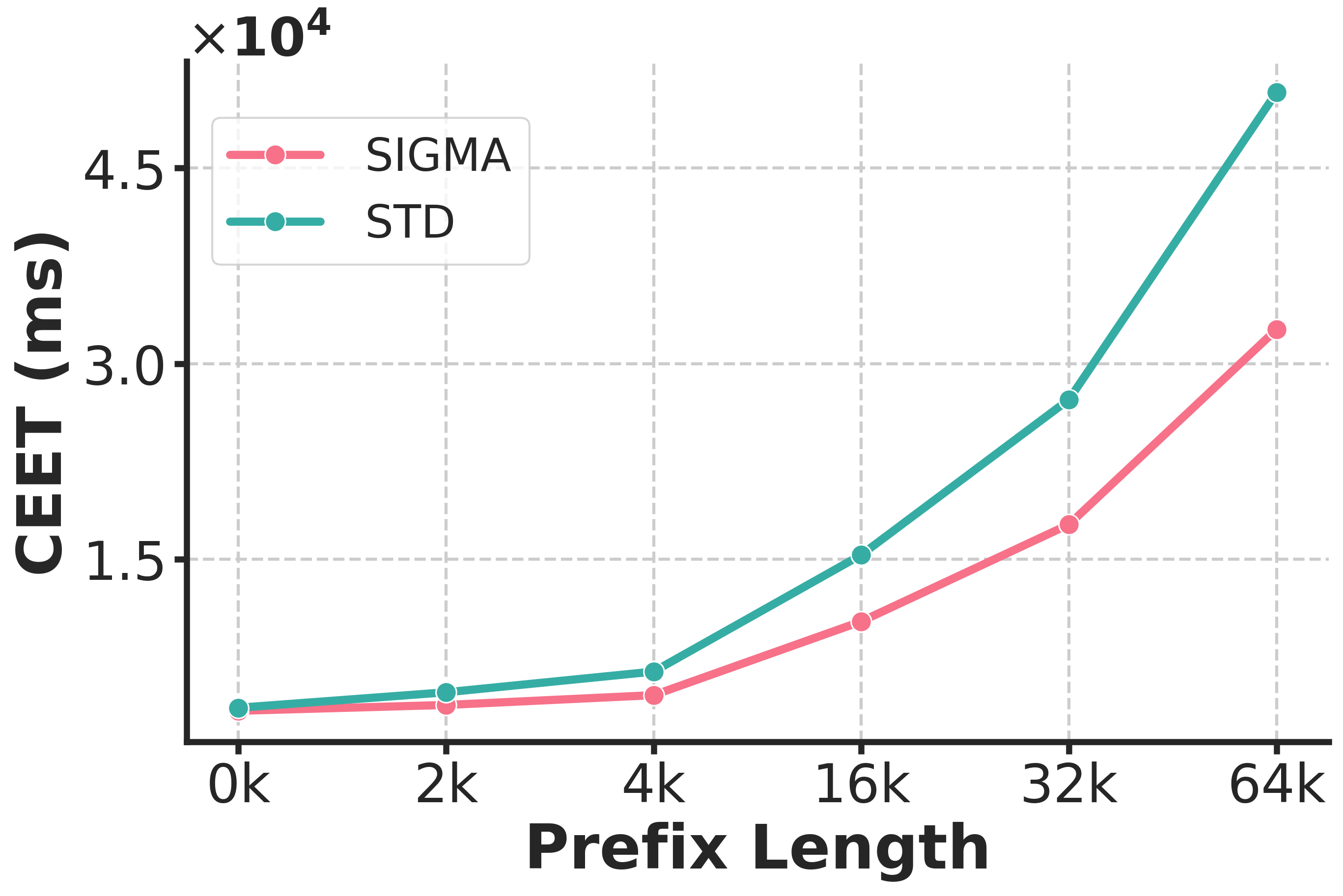}
        \caption{Output Length $= 4k$.}
        \label{fig:kvcache_cost_b}
    \end{subfigure}
    \hfill
    \begin{subfigure}[b]{0.32\textwidth}
        \includegraphics[width=\textwidth]{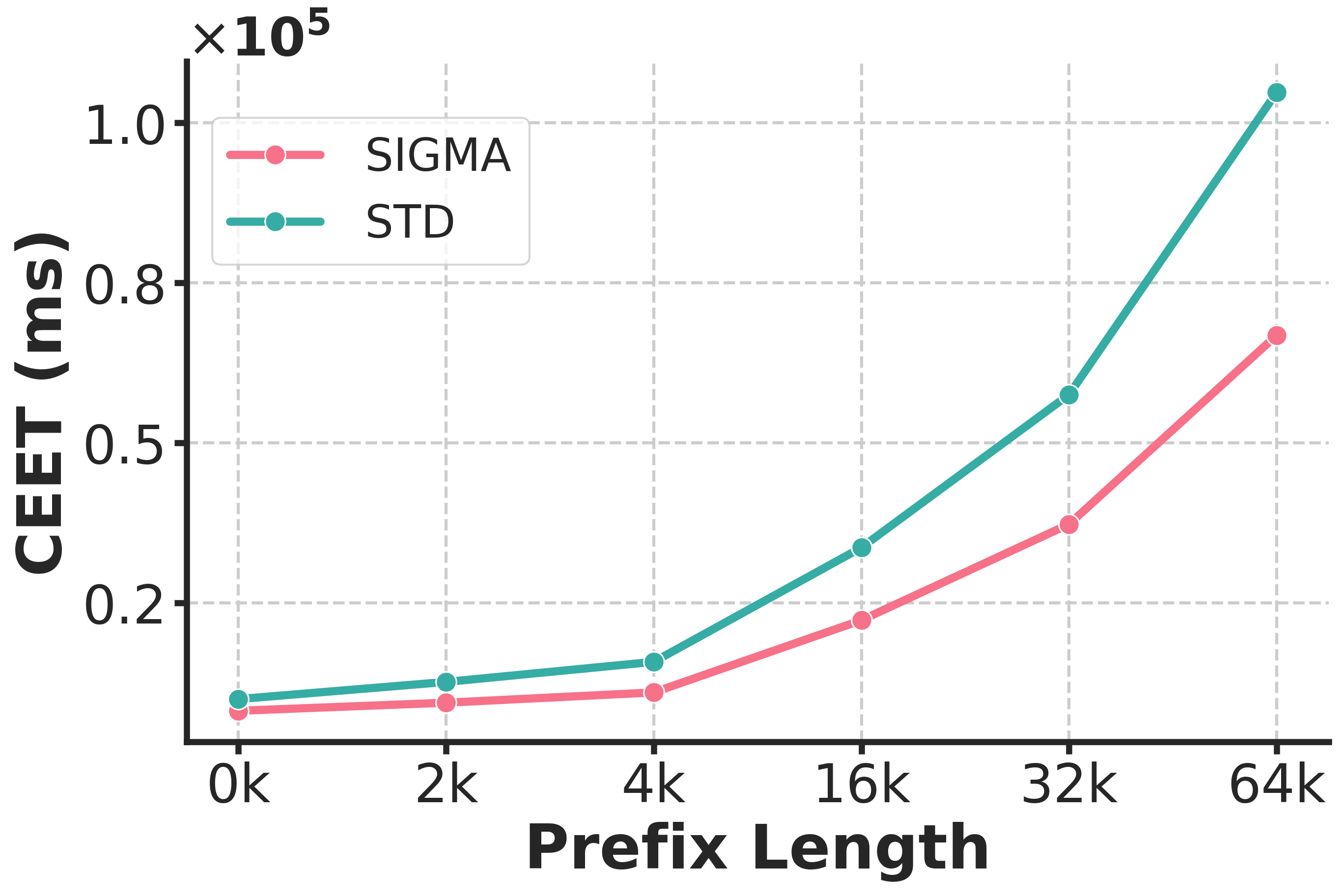}
        \caption{Output Length $= 8k$.}
        \label{fig:kvcache_cost_c}
    \end{subfigure}

    \centering
    \begin{subfigure}[b]{0.32\textwidth}
        \includegraphics[width=\textwidth]{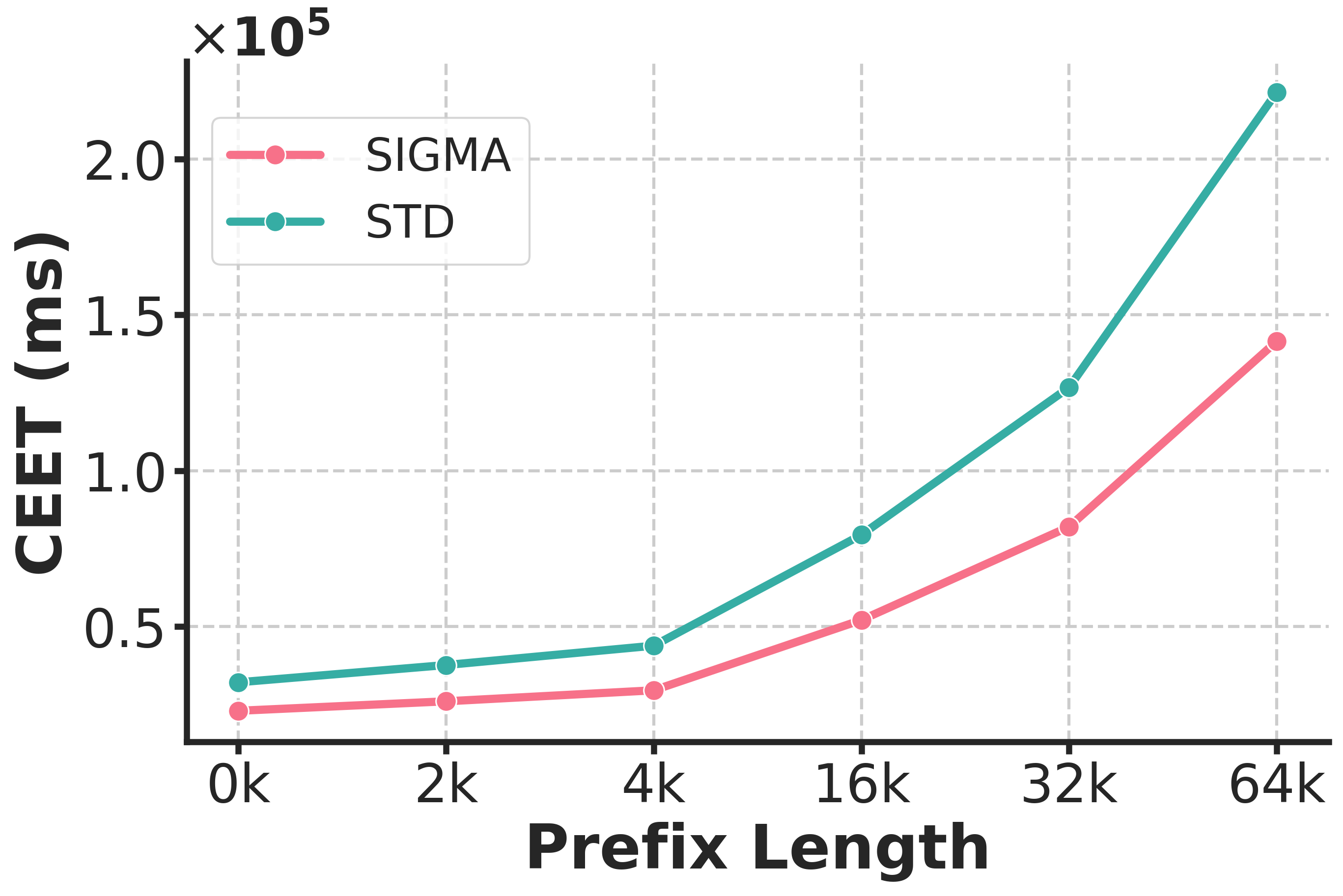}
        \caption{Output Length $= 16k$.}
        \label{fig:kvcache_cost_d}
    \end{subfigure}
    \hfill
    \begin{subfigure}[b]{0.32\textwidth}
        \includegraphics[width=\textwidth]{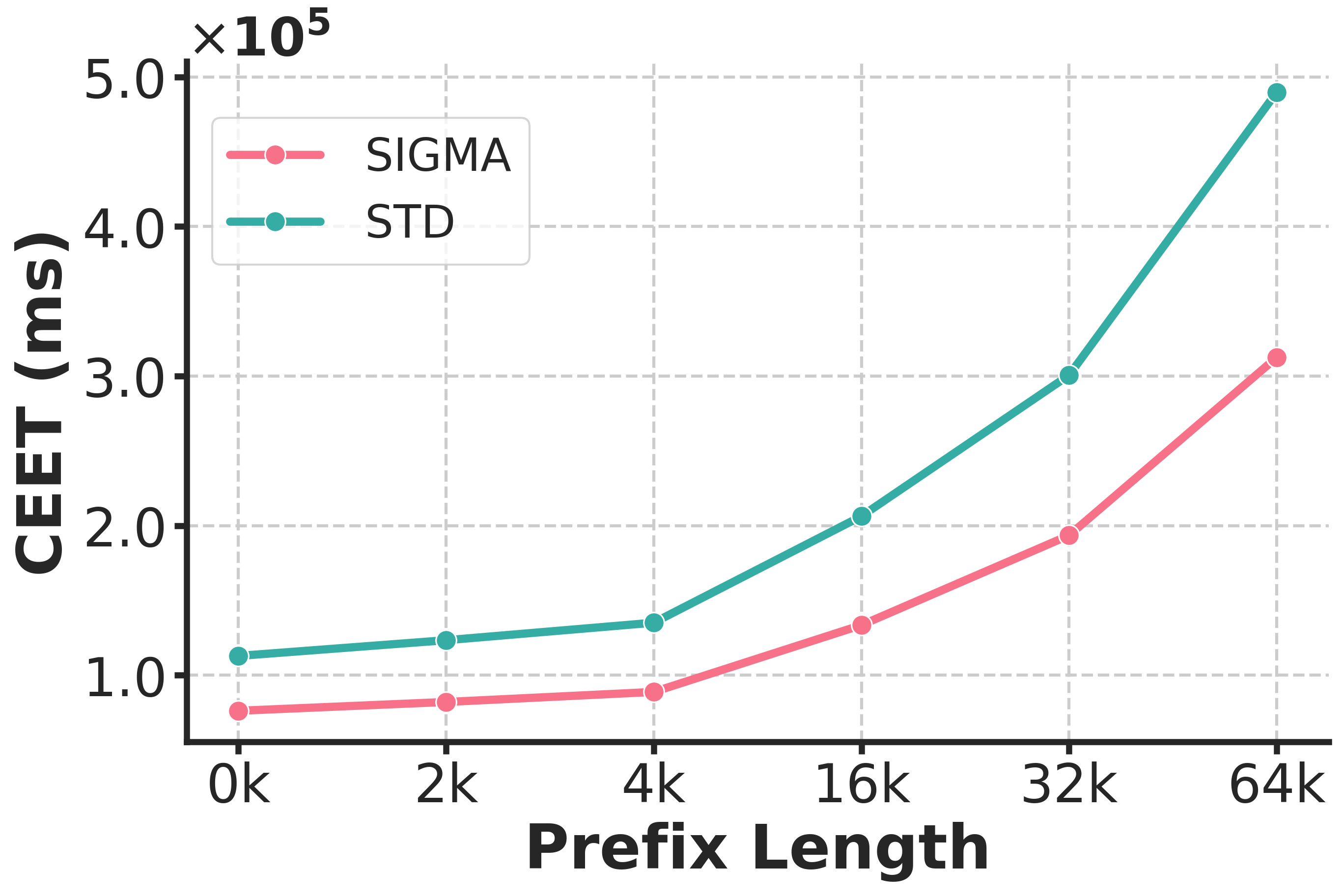}
        \caption{Output Length $= 32k$.}
        \label{fig:kvcache_cost_e}
    \end{subfigure}
    \hfill
    \begin{subfigure}[b]{0.32\textwidth}
        \includegraphics[width=\textwidth]{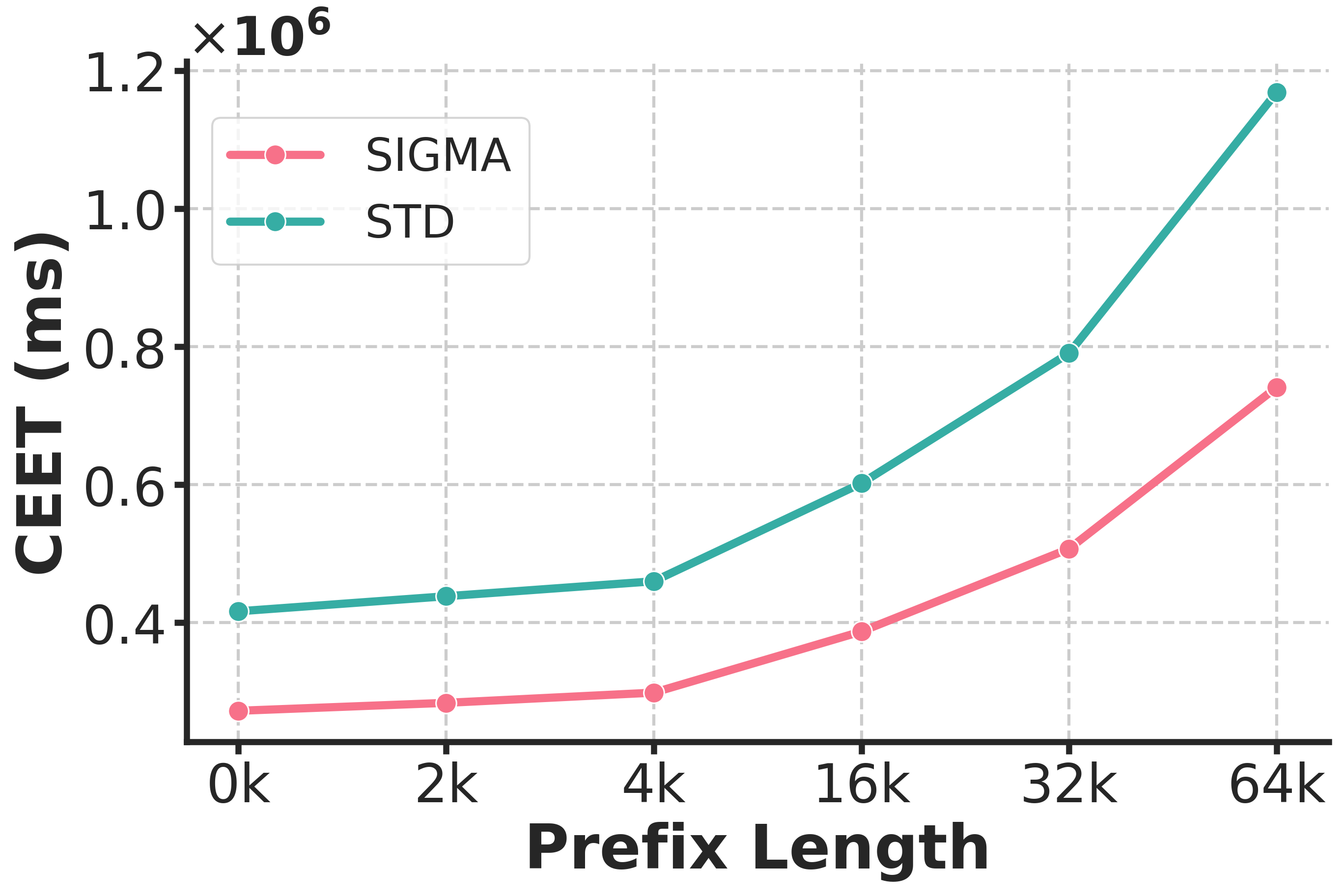}
        \caption{Output Length $= 64k$.}
        \label{fig:kvcache_cost_f}
    \end{subfigure}

    \centering
    \begin{subfigure}[b]{0.49\textwidth}
        \includegraphics[width=\textwidth]{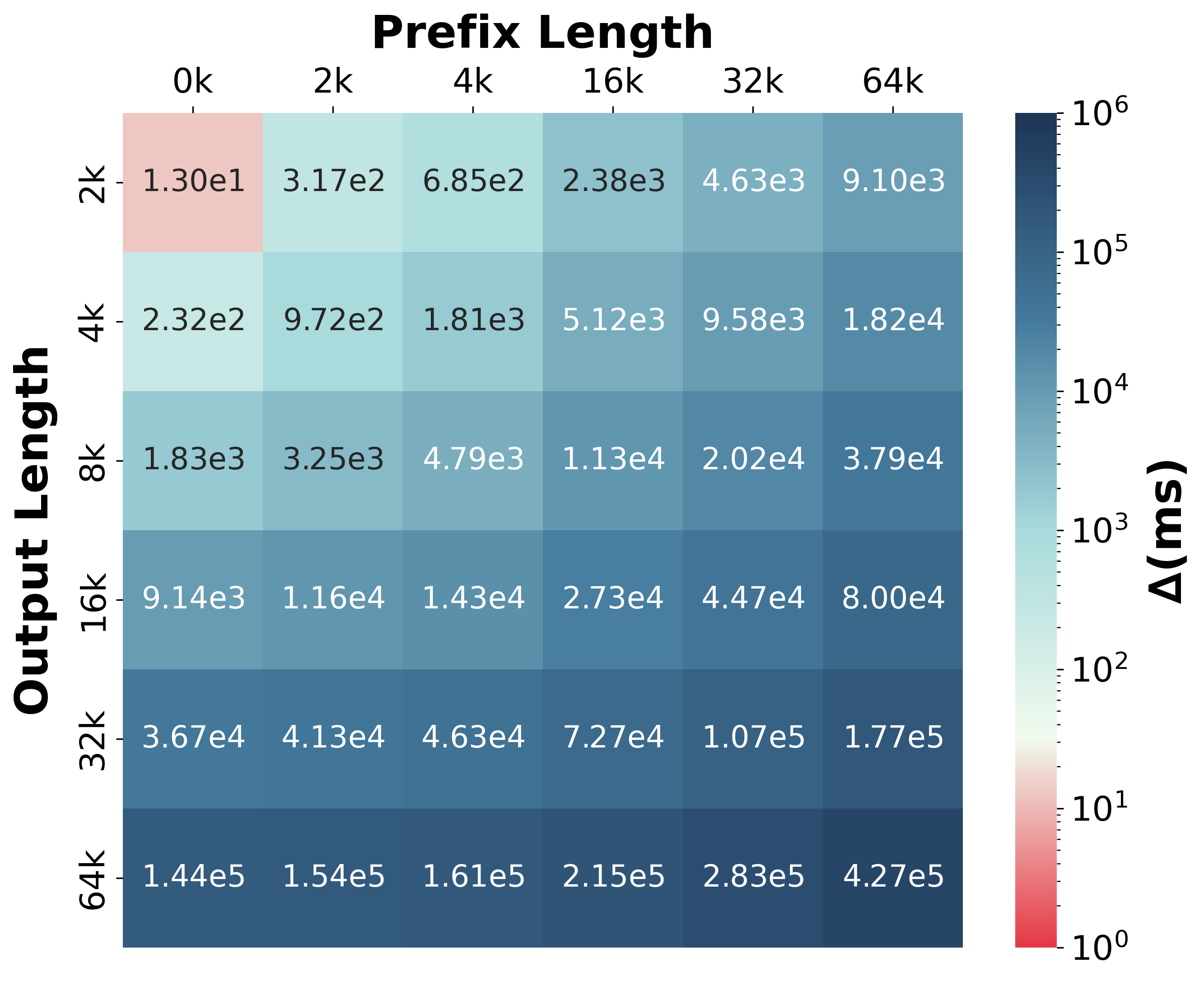}
        \caption{\textsc{Sigma}'s absolute CEET improvment vs \textsc{Std}.}
        \label{fig:kvcache_cost_g}
    \end{subfigure}
    \begin{subfigure}[b]{0.49\textwidth}
        \includegraphics[width=\textwidth]{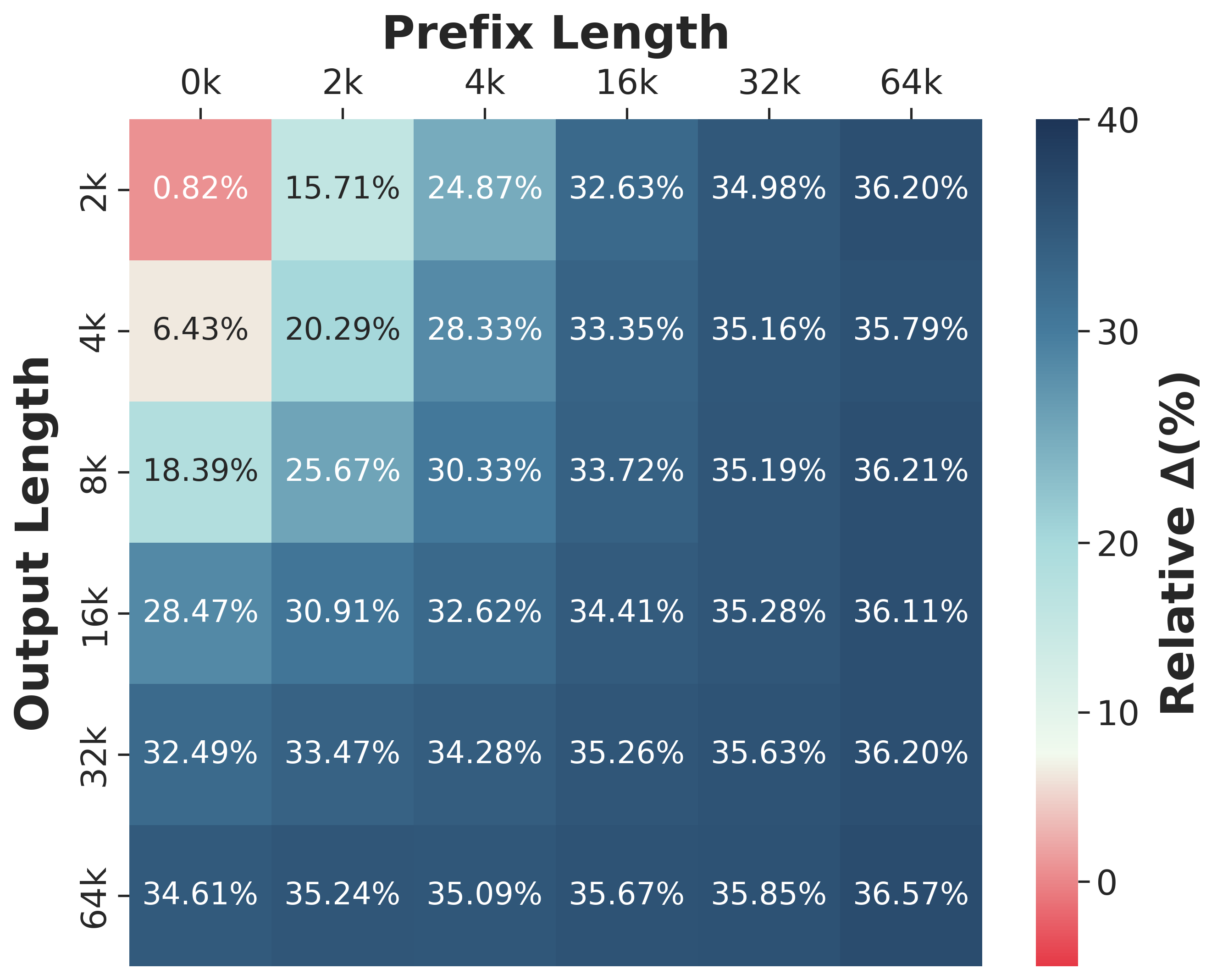}
        \caption{\textsc{Sigma}'s relative CEET improvment vs \textsc{Std}.}
        \label{fig:kvcache_cost_h}
    \end{subfigure}

    \caption{CEET comparison of KV cache between Standard model(\textsc{Std}) and \textsc{Sigma}. From (a) to (f), the output length increases progressively from 2k to 64k tokens.}
    \label{fig:kvcache_cost}
\end{figure*}
\paragraph{CEET Results - Augmented Q.} The augmented Q module is not inherently included within the standard model(\textsc{Std}). In this experiment, we attach the augmented Q module to \textsc{Std} to demonstrate that the augmented Q module's cost remains consistent regardless of variations in other parts of the model or different context lengths. This setup ensures the reliability of efficiency evaluations for other modules.
The CEET results shown in \cref{fig:augq_cost} can strongly testify to it. The relative improvement ratio in \cref{fig:augq_cost_h} for Augmented Q offers clearer support to our hypothesis, as it consistently hovers near zero across nearly all settings. To provide a more rigorous validation, we perform a t-test to evaluate whether the relative improvement ratio of Augmented Q has an expectation of zero. The computed T-value is $-0.214$, and the corresponding P-value is $0.832$, indicating no evidence to reject our hypothesis.

\paragraph{CEET Results - KV Cache.} CEET results of KV cache can be found in \cref{fig:kvcache_cost}. CEET here primarily reflects the cost of loading and storing the KV cache. These operations are largely proportional to the size of the key and value matrices with minimal extraneous influencing factors. Therefore, the results for KV cache align most closely with our theoretical analysis. In \cref{fig:kvcache_cost_a,fig:kvcache_cost_b,fig:kvcache_cost_c,fig:kvcache_cost_d,fig:kvcache_cost_e,fig:kvcache_cost_f}, \textsc{Sigma} demonstrates a significantly lower cost for KV cache operations. As the output length increases from \cref{fig:kvcache_cost_a} to \cref{fig:kvcache_cost_e}, the CEET gap between \textsc{SIGMA} and \textsc{Std} in KV cache becomes increasingly pronounced. As illustrated in \cref{fig:kvcache_cost_h}, 
when the prefix and output lengths reach 64k, \textsc{Sigma} achieves a speedup of 36.57\% compared to \textsc{Std}, which is very close to the theoretically calculated improvement rate of 37.5\%.

\begin{figure*}[!ht]
    \centering
    \begin{subfigure}[b]{0.32\textwidth}
        \includegraphics[width=\textwidth]{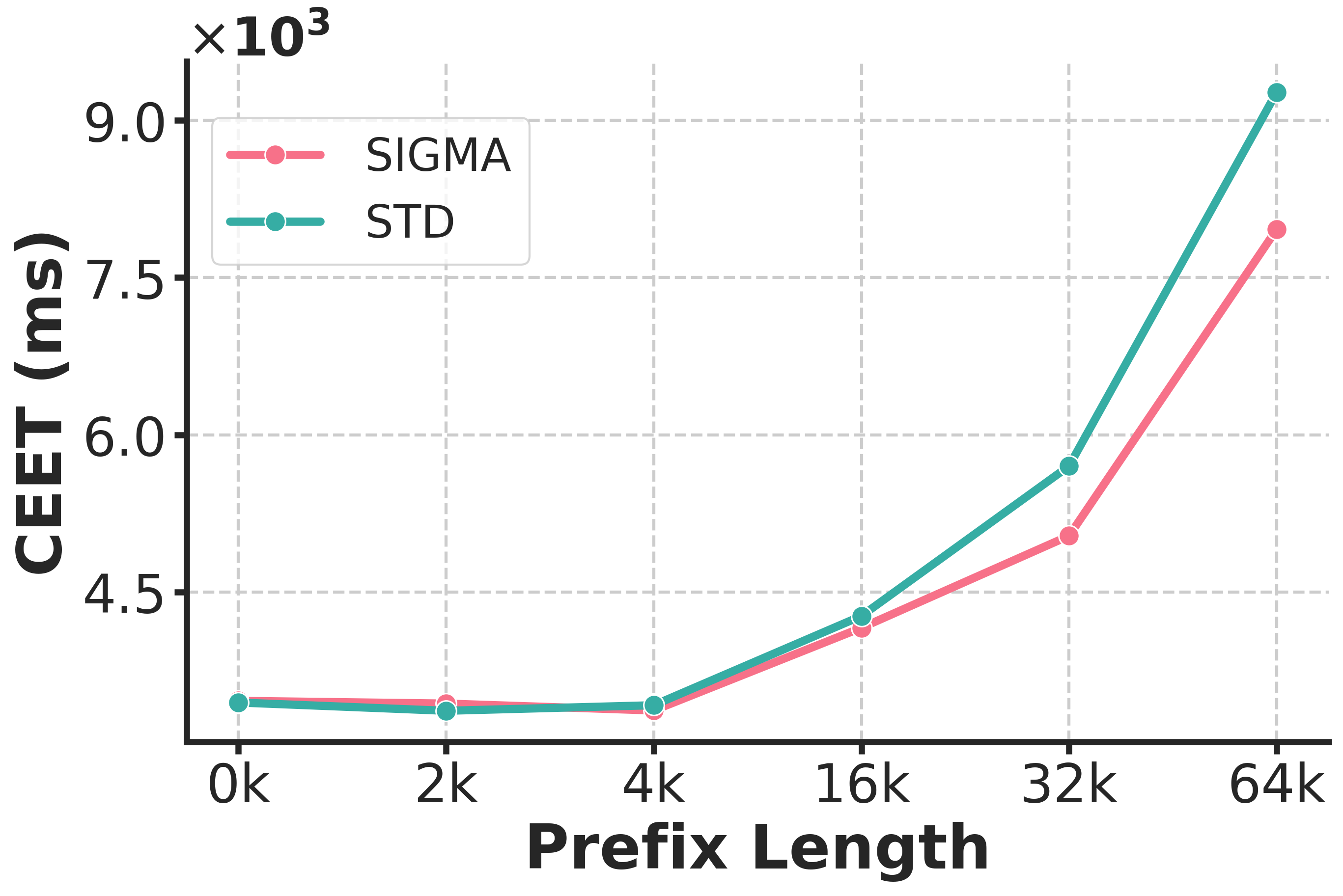}
        \caption{Output Length $= 2k$.}
        \label{fig:attn_cost_a}
    \end{subfigure}
    \hfill
    \begin{subfigure}[b]{0.32\textwidth}
        \includegraphics[width=\textwidth]{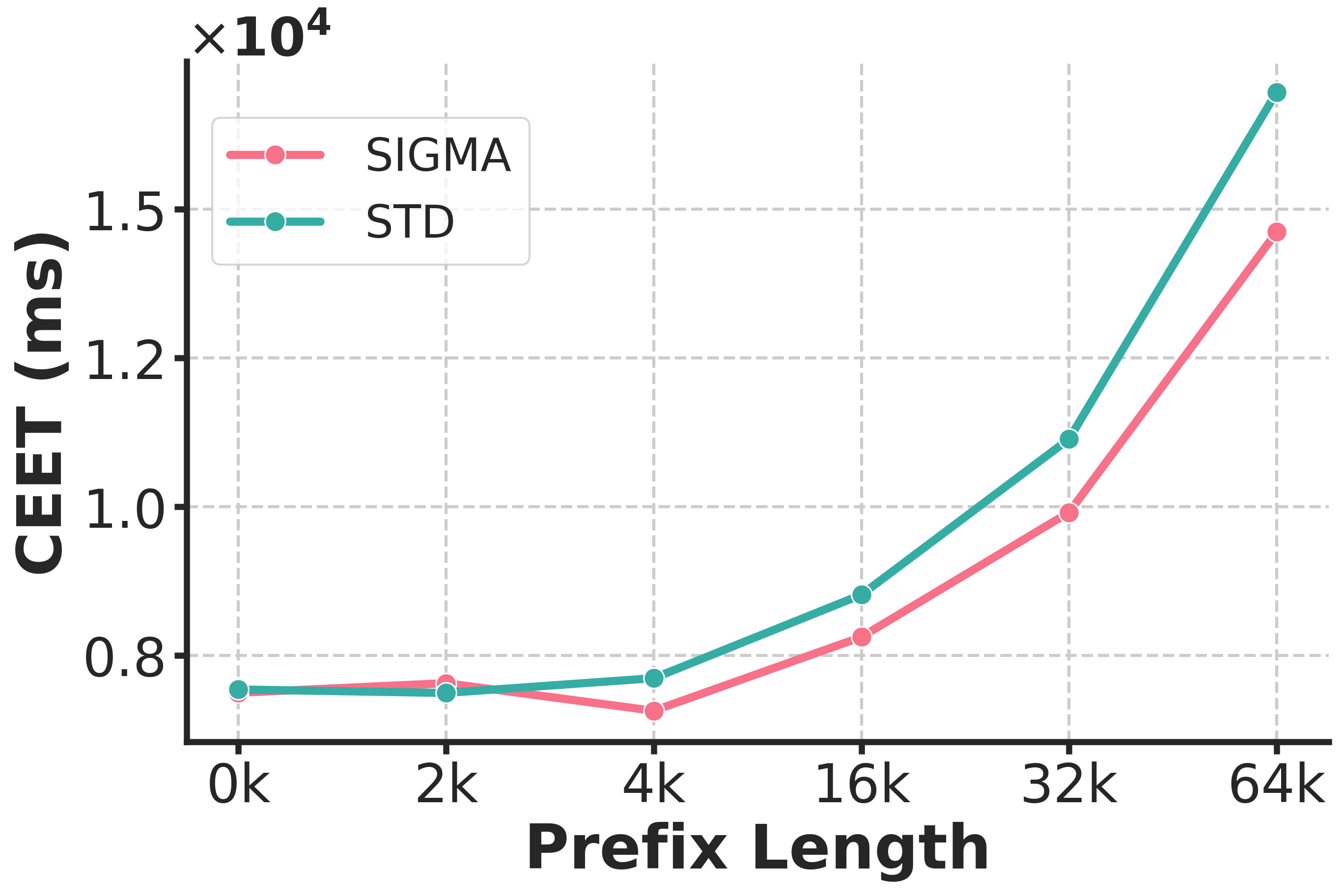}
        \caption{Output Length $= 4k$.}
        \label{fig:attn_cost_b}
    \end{subfigure}
    \hfill
    \begin{subfigure}[b]{0.32\textwidth}
        \includegraphics[width=\textwidth]{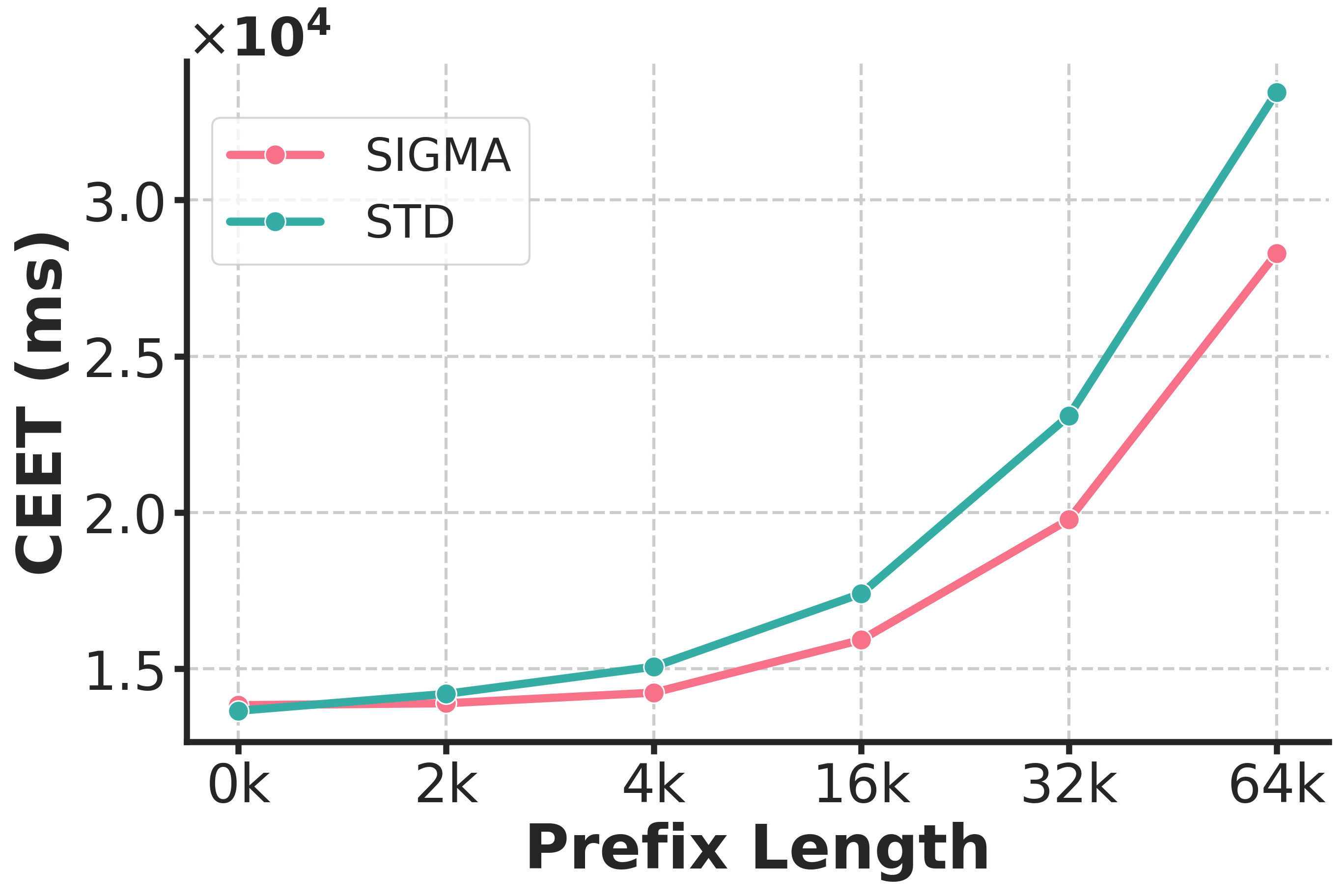}
        \caption{Output Length $= 8k$.}
        \label{fig:attn_cost_c}
    \end{subfigure}

    \centering
    \begin{subfigure}[b]{0.32\textwidth}
        \includegraphics[width=\textwidth]{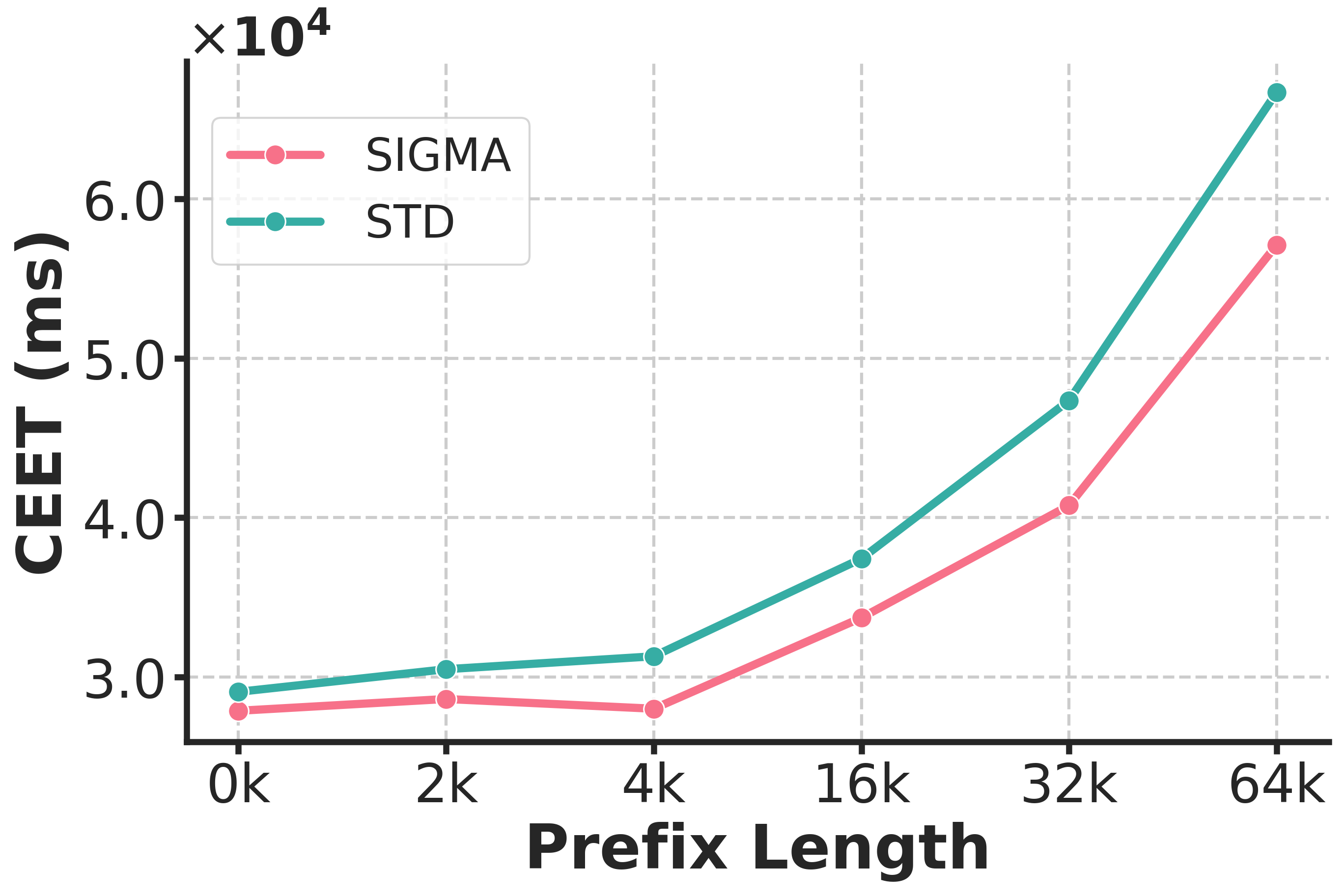}
        \caption{Output Length $= 16k$.}
        \label{fig:attn_cost_d}
    \end{subfigure}
    \hfill
    \begin{subfigure}[b]{0.32\textwidth}
        \includegraphics[width=\textwidth]{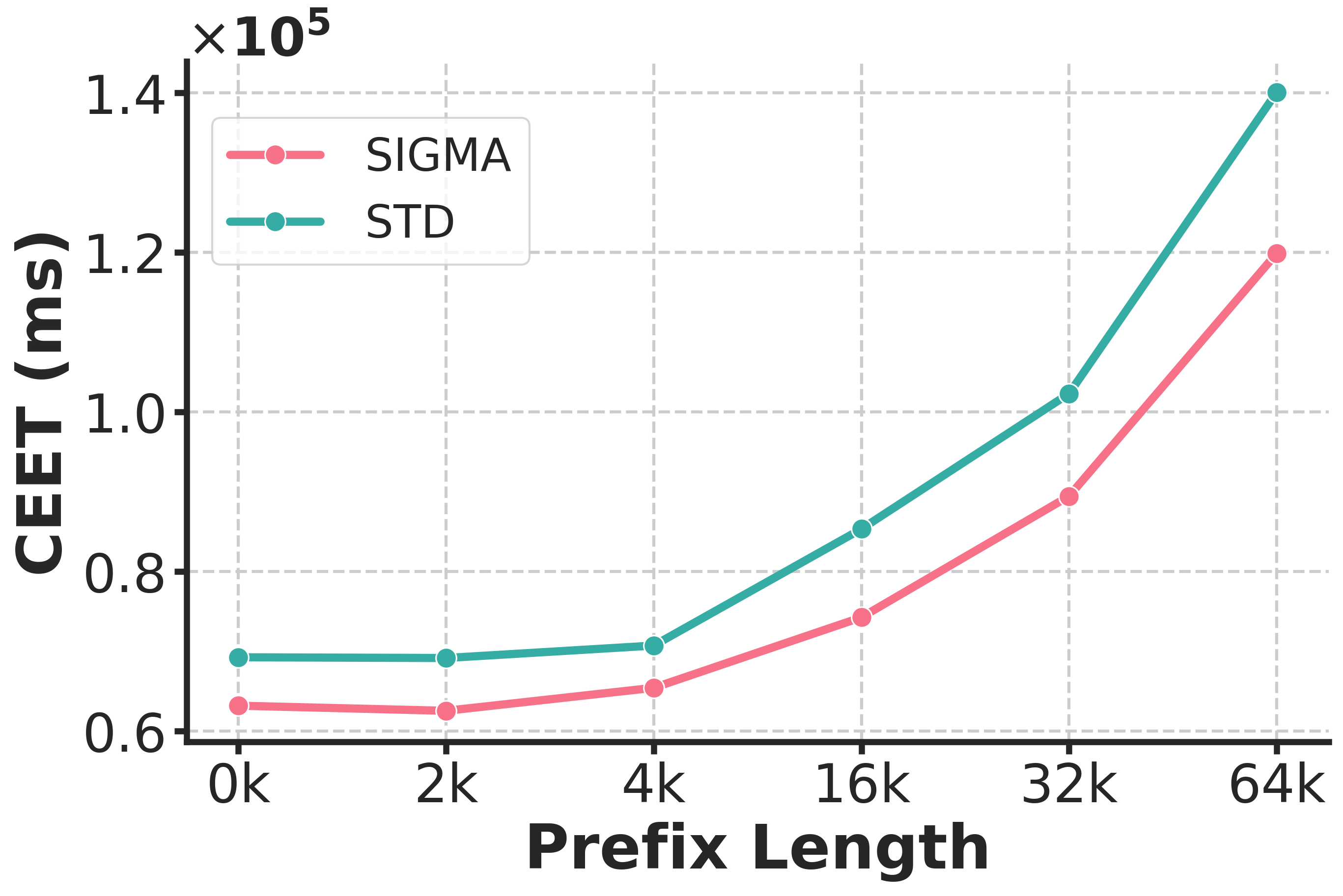}
        \caption{Output Length $= 32k$.}
        \label{fig:attn_cost_e}
    \end{subfigure}
    \hfill
    \begin{subfigure}[b]{0.32\textwidth}
        \includegraphics[width=\textwidth]{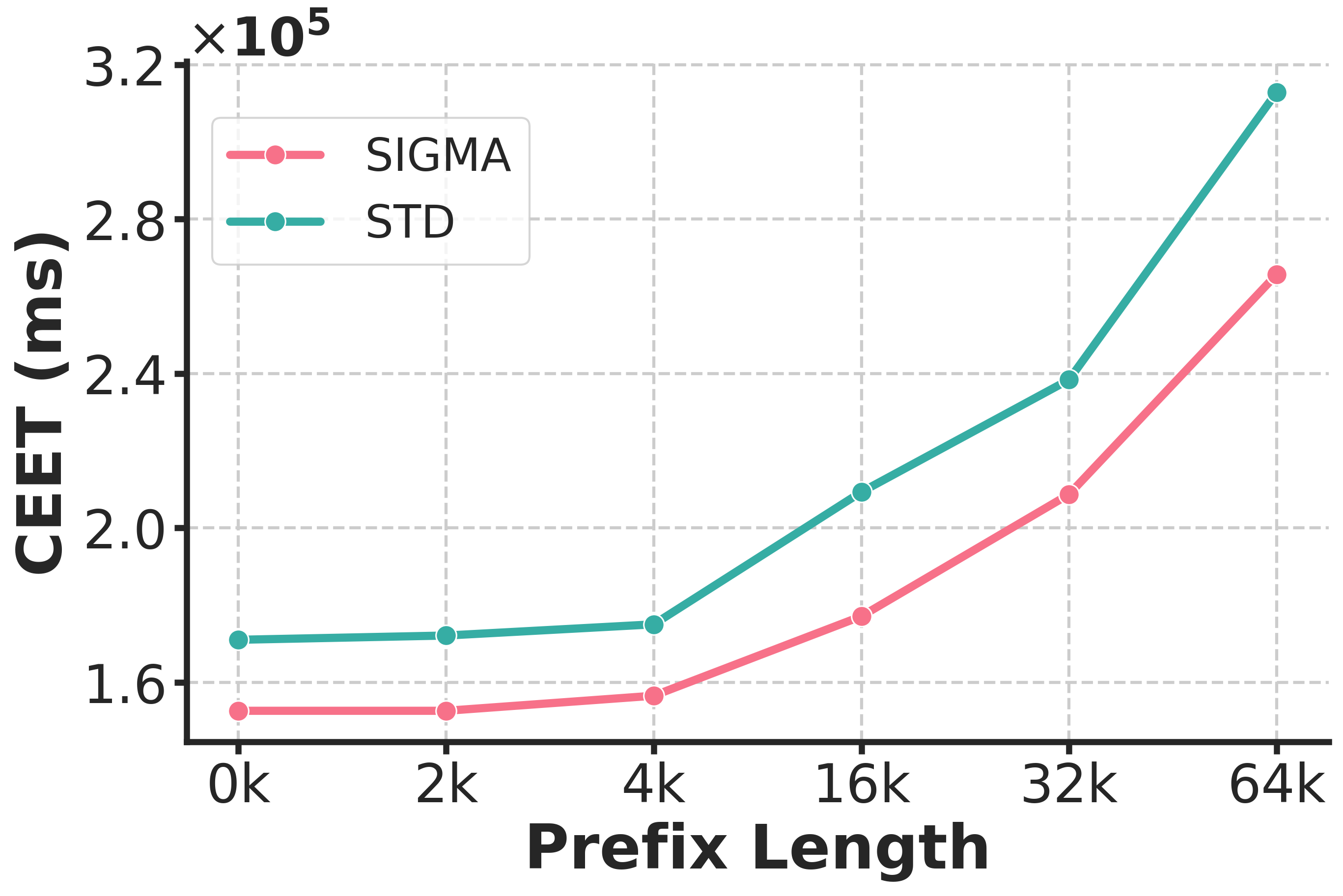}
        \caption{Output Length $= 64k$.}
        \label{fig:attn_cost_f}
    \end{subfigure}

    \centering
    \begin{subfigure}[b]{0.49\textwidth}
        \includegraphics[width=\textwidth]{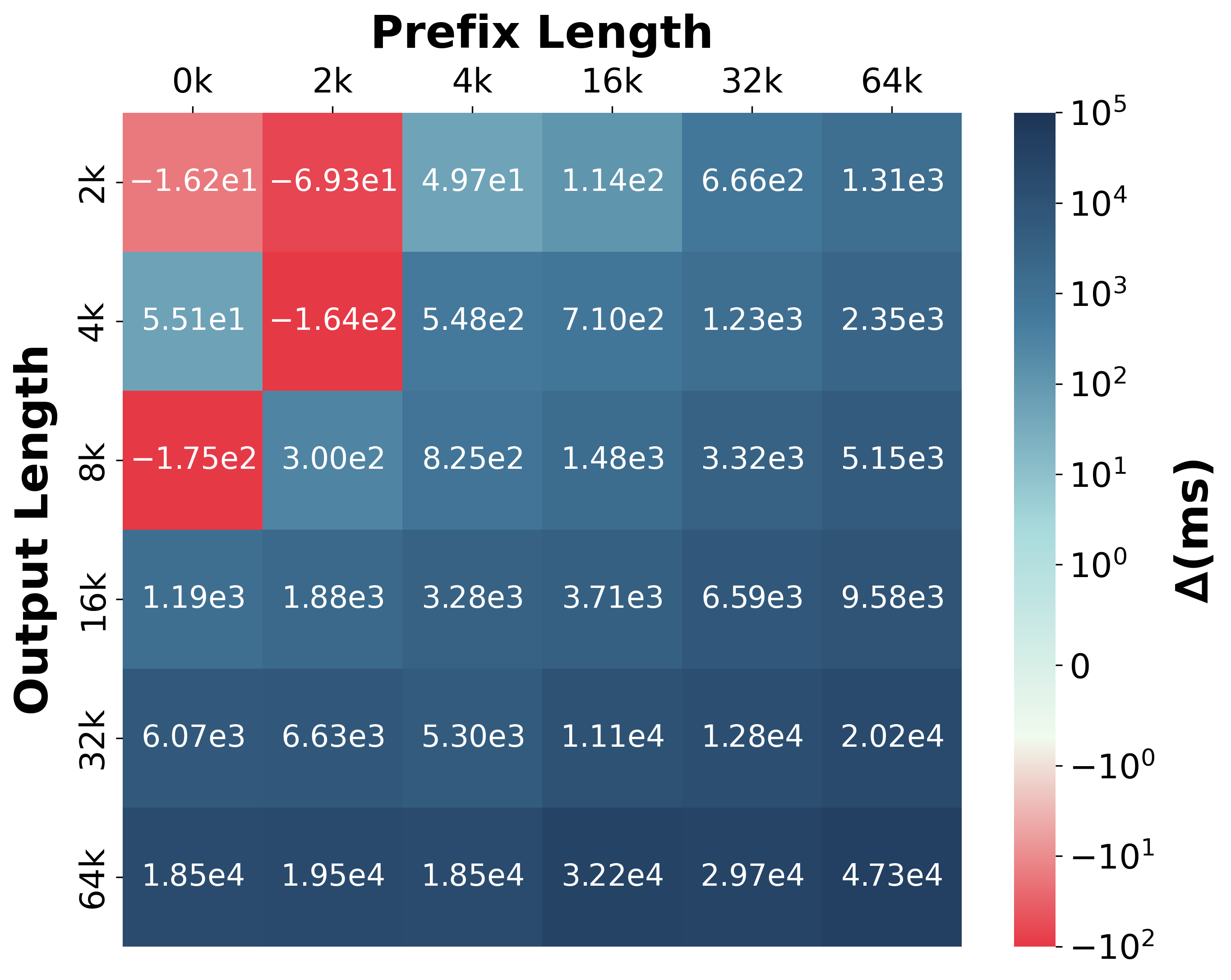}
        \caption{\textsc{Sigma}'s absolute CEET improvment vs \textsc{Std}.}
        \label{fig:attn_cost_g}
    \end{subfigure}
    \begin{subfigure}[b]{0.49\textwidth}
        \includegraphics[width=\textwidth]{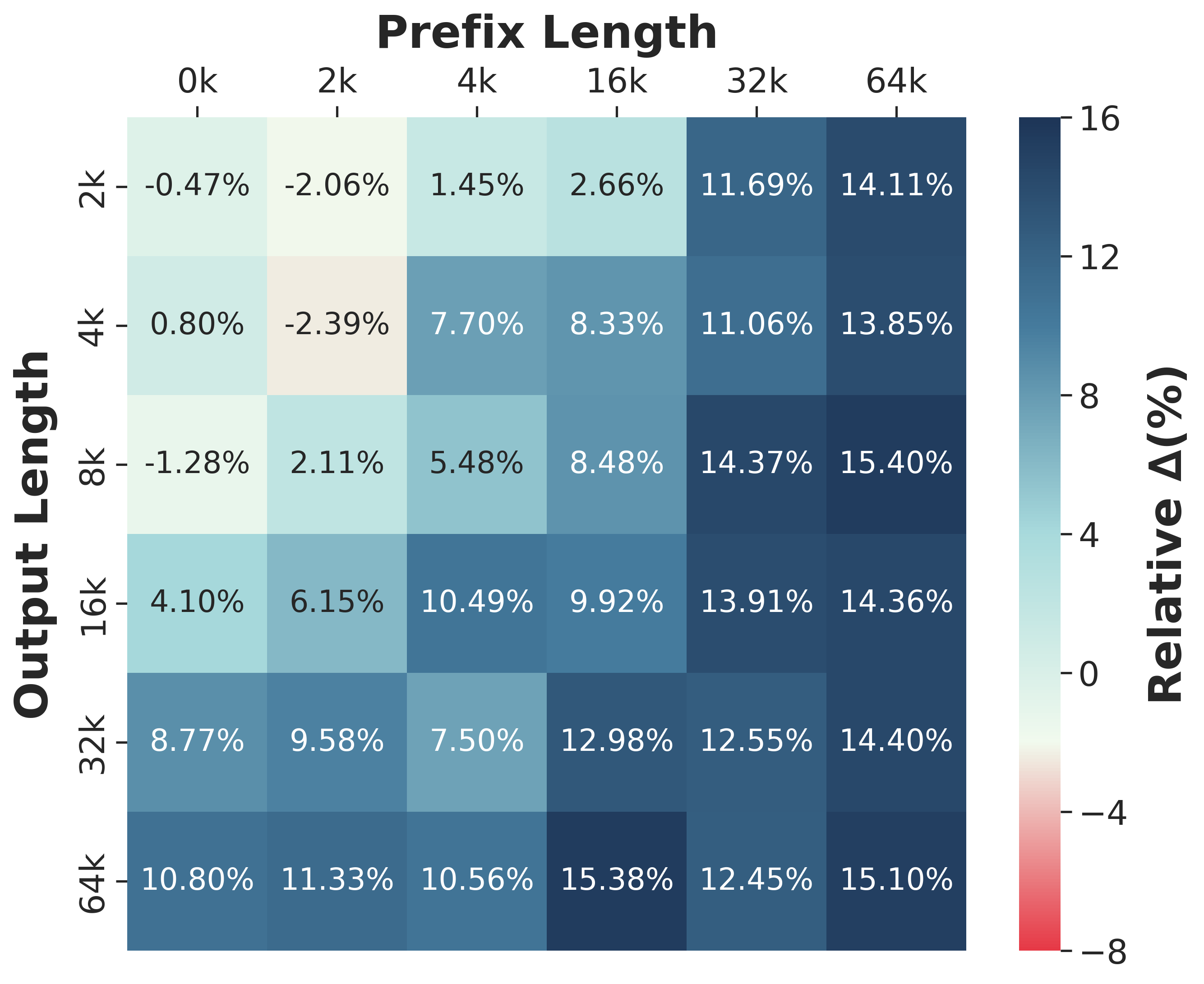}
        \caption{\textsc{Sigma}'s relative CEET improvment vs \textsc{Std}.}
        \label{fig:attn_cost_h}
    \end{subfigure}

    \caption{CEET comparison of attention computation between Standard model(\textsc{Std}) and \textsc{Sigma}. From (a) to (f), the output length increases progressively from 2k to 64k tokens.}
    \label{fig:attn_cost}

\end{figure*}
\paragraph{CEET Results - Attention Computation.} CEET results of the attention computation are demonstrated in \cref{fig:attn_cost}. We observe a steady increase in the relative improvement of Attention Computation. The ratio deviates more significantly from the theoretical value compared to the one of KV Cache. This observation aligns with our analysis in \cref{sec:efficiency_analysys:theoretical_analysis}, as Attention Computation involves a greater number of operations that are not directly influenced by the number of key heads, \textit{e.g.} \texttt{flash\_fwd\_splitkv\_combine\_kernel}. These operations will partially dilute the efficiency improvement we achieve. Besides, the relative improvement ratio here is also lower than the observed KET improvement shown in \cref{fig:kernel_cost}. The underlying reason is the same: since CEET measures the end-to-end time of attention computation, it accounts for additional overhead, such as context switching and other CPU operations, making the improvement less pronounced.

It is also important to note that when the prefix length increases and the output length is kept fixed, the relative improvement still increases. As shown in \cref{fig:attn_cost_h}, \textsc{Sigma} achieves up to a 15.4\% reduction in the CEET of attention computation and becomes increasingly efficient as the prefix or output lengths increase. This is attributed not only to the reduction in the number of key heads but also to the great capability of flash attention kernels to handle long sequences efficiently.

\section{Additional Performance Evaluations}

\subsection{Detailed Pre-training Settings}
\label{sec:detailed_pretraining_settings}
The pre-training data includes general domain data and some domain-specific property data, amounting to total of 6 trillion tokens.
For \textbf{general domain} data, we combine DCLM~\citep{li2024datacomp} and FineWeb-EDU~\citep{penedo2024fineweb} and then remove duplicates to obtain General Dataset $\mathrm{I}$. The number of tokens is approximately 4 trillion. After further quality filtering, we take the filtered data as General Dataset $\mathrm{II}$, with the number of tokens 1 trillion tokens. Finally, we select the data with higher scores and stricter rules as General Dataset $\mathrm{III}$, which contains approximately 200 billion tokens that participate in the annealing phase of \textsc{Sigma}'s pre-training.
For data in the \textbf{math domain}, we use proof-pile-2~\citep{azerbayev2023llemma} and combine it with 280 billion math-related data filtered from General Dataset $\mathrm{I}$ as the pre-training data for the math domain.
Regarding data in the \textbf{code domain}, we refer to the filtering method of StarcoderV2~\citep{li2023starcoder} and select the dataset related to the code domain of 500 billion tokens.
Moreover, we have approximately 1 trillion tokens of synthesized and rewritten pre-training data that have undergone quality screening and contain content from multiple domains, which participate in the later phase of \textsc{Sigma} pre-training.

We conducted pre-training for \textsc{Sigma}-1.5B utilizing 512$\times$A100-40G GPUs and for \textsc{Sigma}-10B using 256$\times$H100-80G GPUs. Initially, we combined datasets from general, mathematical, and coding domains with General Dataset I, adhering to a distribution ratio of General:Math:Code = 8:1:1. The training was carried out with a maximum learning rate of 1.5e-4, starting from a batch size of 4 million tokens, which was incrementally scaled up to 16 million tokens, culminating in a total training volume of 3.5 trillion tokens.

In the subsequent phase, we adjusted the data mix to favor mathematical and coding content, employing the quality-filtered General Dataset II. The mixing ratio for this phase was set to General: Math: Code = 4:3:3, with a total training volume of 1.0 trillion tokens.

For the third phase, we introduced a blend of synthesized and rewritten pre-training data alongside the General Dataset II at a ratio of 6: 4. This phase encompassed a total training volume of 1 trillion tokens, during which the learning rate was reduced to 20\% of its peak value, i.e., 3e-5.

Finally, in the annealing phase, we utilized General Dataset III, which was selected for its highest quality, along with meticulously chosen synthesized and rewritten pre-training data and the system domain data. The learning rate was gradually decreased to zero, concluding with a total training volume of 1 trillion tokens.

\subsection{Additional Results in System Domain}



To assess the quality of our carefully curated system domain data, we also continual pre-train several popular open-source LLMs with our 19.5B system domain pretrain data, including Mistral-7B \citep{jiang2023mistral} and Llama3-8B \citep{dubey2024llama}. We further fine-tune them using a preliminary version of our SFT dataset on the CMDGen NVIDIA subtask of \textsc{AIMicius}. The results are presented in \cref{tab:sigma_system_pre}. Both models exhibit notable performance improvements after fine-tuning, with absolute improvements of 1.5 and 6.4 on the Accuracy metric, respectively.
\begin{table*}[h]
\centering
\small
\caption{Performance of different models on CMDGen NVIDIA subtask in \textsc{AIMicius}. The postfix of ``\textbf{-S}'' indicates that the model has been SFTed using a preliminary version of our SFT dataset, while ``\textbf{-P}'' denotes that the model has been continually pre-trained on our system-domain pre-training dataset.}
\vspace{3pt}
\begin{tabular}{*{1}{l}*{6}{c}}
\toprule
\bf Model & 
\textbf{\begin{tabular}[c]{@{}c@{}} CMD \\ Score \end{tabular}} &
\textbf{\begin{tabular}[c]{@{}c@{}} Output \\ Score \end{tabular}} &
\textbf{\begin{tabular}[c]{@{}c@{}} Calibration \\ Score \end{tabular}} &
\textbf{\begin{tabular}[c]{@{}c@{}} Exact \\ Match \end{tabular}} &
\textbf{\begin{tabular}[c]{@{}c@{}} Success \\ Ratio \end{tabular}} &
\bf Accuracy \\
\midrule
\bf Mistral-7B-S & 80.6 & 58.7 & 62.0 & 24.9 & 19.0 & 30.7 \\
\bf Mistral-7B-P-S & 83.4 & 65.3 & 66.3 & 23.9 & 21.5 & 32.2 \\
\bf Llama3-8B-S & 86.4 & 69.1 & 64.4 & 42.0 & 32.7 & 50.7 \\
\bf Llama3-8B-P-S & 87.5 & 72.2 & 69.3 & 46.3 & 37.1 & 57.1 \\
\bottomrule
\end{tabular}
\label{tab:sigma_system_pre}
\end{table*}
\begin{table*}[h]
\centering
\small

\caption{Comparisons with baseline models on general, coding, and math problem-solving tasks. Differences with original reports in the baseline models are due to our unified re-evaluations for fair comparisons.}
\vspace{0.5em}
\label{tab:ps_result}
\begin{tabular}{l c c ccccccc}
\toprule
\multirow{2}{*}{\textbf{Model}} & \multirow{2}{*}{\textbf{Params}} & \multirow{2}{*}{\textbf{Avg.}} & \multicolumn{3}{c}{\textbf{General}} & \multicolumn{2}{c}{\textbf{Coding}} & \multicolumn{2}{c}{\textbf{Math}}\\ \cmidrule(lr){4-6}    \cmidrule(lr){7-8} \cmidrule(lr){9-10}
&&& {MMLU} & {MMLU-Pro} & {BBH} & {HumanEval} & {MBPP} & {MATH} & {GSM8K}\\
\midrule 
Pythia & 1.0B & 10.76 & 26.1 & 11.1 & 24.6 & 5.5 & 4.0 & 2.1 & 1.9 \\
TinyLlama & 1.1B & 11.13 & 26.7 & 11.3 & 25.4 & - & 10.8 & 1.9 & 1.8 \\
Bloom & 1.1B & 5.20 & 26.5 & 2.1 & 6.5 & - & - & 0.1 & 1.2 \\
OLMo & 1.2B & 10.53 & 26.2 & 10.8 & 25.8 & 5.5 & 0.3 & 2.3 & 2.8 \\
OPT & 1.3B & 8.95 & 24.7 & 10.8 & 22.8 & - & - & 1.7 & 2.7 \\
CerebrasGPT & 1.3B & 9.67 & 26.6 & 11.2 & 24.5 & 1.8 & 0.8 & 1.0 & 1.8\\
Phi1 & 1.3B & 19.81 & 24.9 & 11.0 & 22.9 & \textbf{48.2} & 27.2 & 2.0 & 2.5\\
Pythia & 1.4B & 10.73 & 25.7 & 10.9 & 24.8 & 4.9 & 4.9 & 2.1 & 1.7\\
DCLM & 1.4B & 17.92 & \textbf{47.7} & \underline{16.3} & 29.9 & 9.1 & 13.2 & 2.5 & 6.8\\
StableLM2 & 1.6B & 17.88 & 38.1 & 8.7 & 26.8 & 7.3 & 17.7 & 5.2 & \underline{21.3}\\
SmolLM & 1.7B & 17.60 & 29.9 & 11.7 & 28.9 & 1.2 & \underline{41.0} & 4.2 & 6.4\\
Gemma & 2.0B & \underline{26.58} & 41.2 & 14.7 & \textbf{36.0} & \underline{25.0} & \textbf{41.5} & \underline{10.9} & 16.8\\ \midrule
\textsc{Sigma} (Ours) & 1.5B & \textbf{27.10} & \underline{47.0} & \textbf{17.6} & \underline{32.7} & 21.3 & 30.7 & \textbf{12.7} & \textbf{27.8}\\
\bottomrule
\end{tabular}
\end{table*}

\subsection{Additional Results in General Domain}
\label{sec:appendix_problem_solving}
\noindent\textbf{Problem-Solving Tasks.}
We also evaluate on three general problem-solving benchmarks MMLU (5-shot)~\citep{hendrycks2020measuring}, MMLU-Pro (5-shot)~\citep{wang2024mmlu}, and BBH (3-shot)~\citep{suzgun2022challenging}, two code generation benchmarks HumanEval (0-shot)~\citep{chen2021evaluating} and  MBPP~\citep{austin2021program}, and two math problem datasets MATH (5-shot)~\citep{hendrycksmath2021} and GSM8K (5-shot)~\citep{cobbe2021training}.
The evaluation results on various general, coding and math problem-solving benchmarks are shown in \cref{tab:ps_result}. According to the results, \textsc{Sigma}-1.5B achieves an average of 27.1 on all tasks, which is comparable to strong baseline models such as Phi1 and Gemma-2B. Specifically, our model reaches top-2 performances on all three general benchmarks: MMLU, MMLU-Pro, and BBH, showing its capability to apply world knowledge to various problem-solving scenarios. In the code domain, \textsc{Sigma} achieves 21.3\% and 30.7\% pass rates on the HumanEval and MBPP benchmarks, outperforming Phi1-1.3B on MBPP, a model customized for code generation. However, in HumanEval, \textsc{Sigma} exhibits inferior performance relative to Gemma-2B and Phi1-1.3B, which may be attributed to the relatively lower proportion of code in our pre-training corpus. In the math domain, \textsc{Sigma} reaches 12.7\% and 27.8\% precision on the MATH and GSM8K benchmarks, surpassing competitive baseline models such as StableLM-1.6B and Gemma-2B by a large margin. Notably, \textsc{Sigma}-1.5B maintains similar commonsense reasoning and general problem-solving capabilities to DCLM-1.4B, but greatly advances on all coding and math benchmarks. 
Despite some overlap in the pre-training corpus adopted by \textsc{Sigma} and that by DCLM-1.4B, the results presented above further demonstrate the importance of mixing mathematical and coding data into the general corpus throughout the entire pre-training process, which contributes to a more balanced improvement of the model's general, reasoning, and programming capabilities.

\section{Future Work}

Despite its advancements, \textsc{Sigma} still presents significant opportunities for improvement. 
For instance, further optimization on the architecture of \textsc{Sigma} has not been fully explored. Key areas for investigation include the trade-off between Augmented Q and Feed-Forward Network (FFN) parameters, varied key-value (KV) heads compression across layers, and appropriate hyper-parameters for scale up.
Additionally, the number of tasks currently included in \textsc{AIMicius} is still limited. It is essential to evaluate the model's ability on a wider spectrum of system domain challenges. 
Besides, we discover that \textsc{Sigma}’s performance is largely constrained by the quality of the synthetic data used for pre-training. 
We believe that model self-evolution and lifelong learning could be possible avenues to overcome this limitation.
In future work, we plan to address these issues, 
unlock the vast potential of \textsc{Sigma} within the system domain.

\end{document}
